**Master's Thesis**
MSC-016

Mechanik und Meerestechnik

# Learning Model Predictive Control for Competitive Autonomous Racing

by
Lukas Brunke

Supervisors (TUHH): Prof. Dr.-Ing. Robert Seifried
Prof. Dr.-Ing. Herbert Werner
Supervisors (UC Berkeley): Prof. Dr.-Ing. Francesco Borrelli
Ugo Rosolia, M. Sc.

Hamburg University of Technology
Institute of Mechanics and Ocean Engineering
Prof. Dr.-Ing. R. Seifried

# Contents

# Chapter 1

# Introduction

Autonomous driving has been a main focus of research efforts over the last decade. There already exist a number of successful techniques for self-driving vehicles in highway and urban driving scenarios, refer to [GaoEtAl12, RosoliaBruyneAlleyne17, PadenEtAl16]. However, the interaction of self-driving vehicles with other cars, either autonomous or human-controlled, remains a major challenge. This is mainly due to the fact that other cars' intentions are unknown.

While determining an optimal, safe, and robust strategy for successfully driving on highways and urban settings is demanding, executing the calculated strategy can be achieved by using simple control methods. In contrast, consider the racing problem, where a vehicle is driven around a predefined race track at high velocities to minimize lap time. The vehicle needs to be controlled at its limit of handling, where the dynamics are highly nonlinear. The challenge of controlling a vehicle at its handling limits while in a dynamic environment with other vehicles whose intentions are unknown yields the multi-agent racing problem.

The problem of car racing can be formulated as an iterative control problem, where each iteration of the controller is equal to completing one lap on the race track. Recently, [RosoliaBorrelli17b] introduced the novel control framework called learning model predictive control (LMPC). LMPC combines model predictive control (MPC) with iterative learning control (ILC). LMPC creates a terminal safe set from past iterations and converges to a locally optimal solution. Since LMPC has already been successfully applied to single-agent racing in [RosoliaCarvalhoBorrelli17], the goal of this thesis is to design a learning model predictive controller that allows multiple agents to race competitively on a predefined race track in real-time. The proposed controller assumes that each agent is able to perfectly predict other agent's trajectories.

This thesis addresses two major shortcomings in the single-agent formulation. Previously, the agent determines a locally optimal trajectory but does not explore the state space, which may be necessary for overtaking maneuvers. Additionally, obstacle avoidance for LMPC has been achieved in [RicciutiRosoliaGonzales18] by using a non-convex terminal set, which increases the complexity for determining a solution to the optimization problem. The proposed algorithm for multi-agent racing explores the state space by executing the LMPC for multiple different initializations, which yields a richer terminal safe set. Furthermore, a new method for selecting states in the terminal set is developed, which keeps the convexity for the terminal safe set and allows for taking suboptimal states.

The thesis has the following structure: Chapter 2 introduces the kinematic and dynamic bicycle models. These models are used in the simulator and controller for modeling the behavior of a car. Chapter 3 presents the MPC-framework and an overview of LMPC. Chapter 4 focuses on the application of LMPC to the problem of autonomous race driving in the single- and multi-agent case. Chapter 5 validates the proposed controller in simulations and on the Berkeley Autonomous Race Car (BARC) platform. Finally, chapter 6 summarizes and evaluates the results of the validation and provides an outlook on future work.

# Chapter 2

# Vehicle Dynamics

This chapter introduces the different vehicle models used in this thesis and points out the necessary assumptions, advantages and drawbacks of each model. First, the simple kinematic bicycle model is derived in both the Cartesian coordinate system and the Frenet reference frame. Then, the equations of motions for the more accurate dynamic bicycle model in the Frenet coordinate system are determined. These equations describe the state evolution in the LMPC for the application to racing.

## 2.1 Kinematic Bicycle Model

The kinematic bicycle model requires little computational effort and no parameter estimation. It provides a good approximation of the motion of a vehicle with Ackermann steering at low velocities. The necessary quantities for the kinematic bicycle model are indicated in Figure 2.1. The model simplifies the two front and rear tires of the real vehicle by considering only one tire for the front and rear. The front and rear tires are apart from the vehicle's center of gravity at a distance of $l_{\mathrm{f}}$ and $l_{\mathrm{r}}$, respectively. The front and rear tires can be steered to the steering angles $\delta_{\mathrm{f}}$ and $\delta_{\mathrm{r}}$, respectively. However, considering the employed hardware during the experiments, it is sufficient to consider front steering only. It is therefore assumed that for the rear steering angle $\delta_{\mathrm{r}} = 0$. The orientation, or heading angle of the vehicle, is given by $\psi$, while the slip angle, which is the angle between the velocity vector $\boldsymbol{v}$ and the longitudinal vehicle axis, is given by $\beta$.

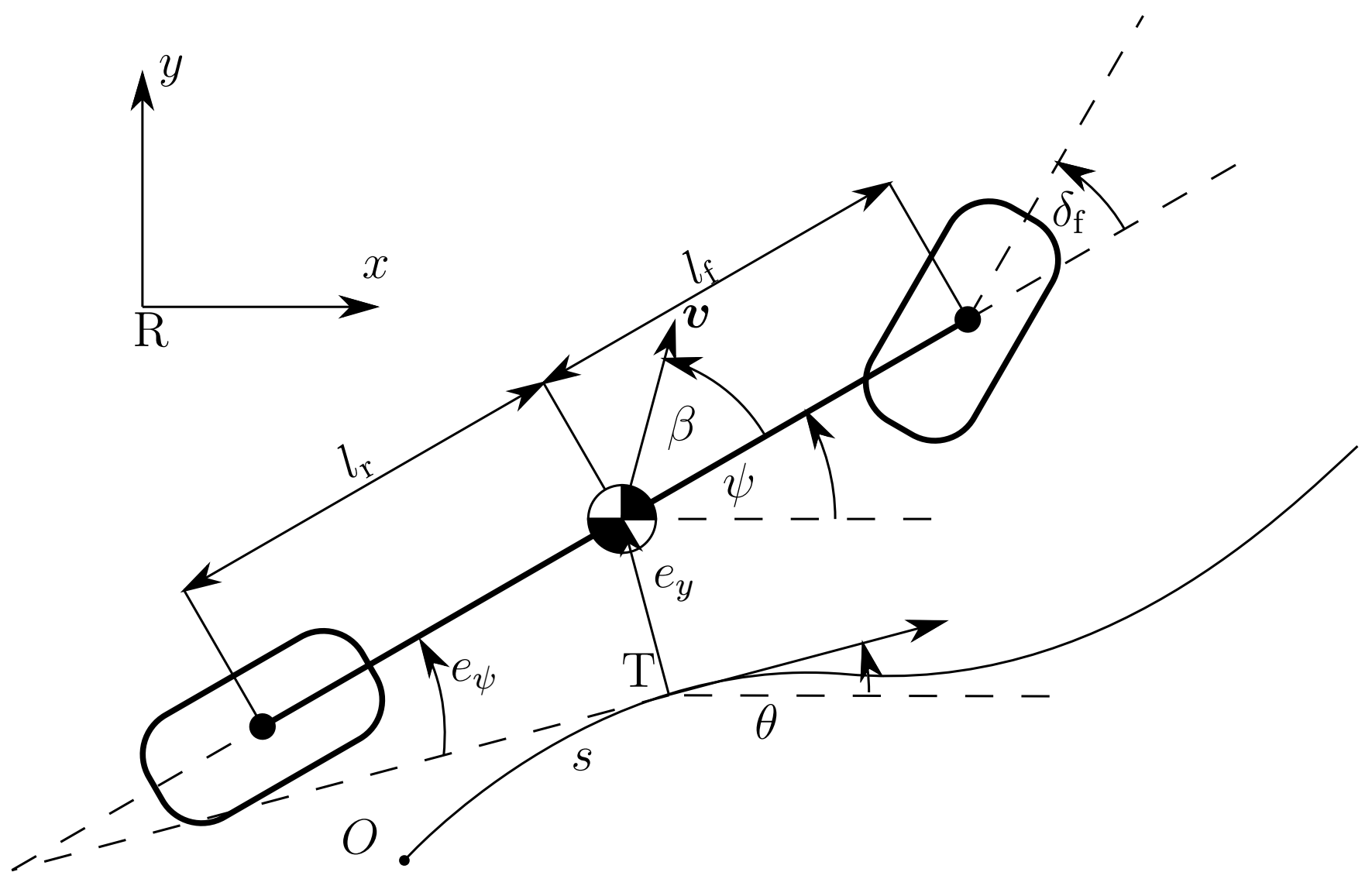

Figure 2.1: Kinematic bicycle model with the relevant quantities in the Cartesian coordinate system and the Frenet reference frame.

### 2.1.1 Cartesian Coordinate System

The state of the kinematic bicycle model is fully described by

$$\boldsymbol{x}_t = [x, y, \psi, v]^{\mathsf{T}} . \tag{2.1}$$

The equations of motion of the kinematic bicycle model in the $xy$-plane are, e.g., derived in [Rajamani12] and given as

$$\dot{x} = v \cdot \cos(\psi + \beta), \tag{2.2}$$
$$\dot{y} = v \cdot \sin(\psi + \beta), \tag{2.3}$$
$$\dot{\psi} = \frac{v}{l_{\mathrm{r}}} \cdot \sin(\beta), \tag{2.4}$$
$$\dot{v} = a, \tag{2.5}$$
$$\beta = \arctan\left(\frac{l_{\mathrm{r}}}{l_{\mathrm{f}} + l_{\mathrm{r}}} \tan(\delta_{\mathrm{f}})\right), \tag{2.6}$$

with the acceleration $a$ and the front steering angle $\delta_{\mathrm{f}}$ as inputs.

This model does not consider the vehicle's mass, nor does it consider the forces acting on the tires. Therefore, this model is only valid under the assumption that the slip angles of the front and rear tires is zero. It is assumed that the angle of velocity vector at each tire is the same as that tire's steering angle. Consequently, the lateral forces that might result from a tire slip or slippery road conditions are neglected. For that reason, the model mismatch increases at high velocities and the kinematic bicycle model can only accurately model vehicle dynamics at low velocities.

### 2.1.2 Frenet Coordinate System

Transforming the kinematic bicycle model from the Cartesian coordinate system to a moving Frenet reference frame, allows a race track's state constraints to be defined as simple box constraints. The vehicle state can equivalently be expressed with respect to a curve, which can be the center line of a race track. The Frenet coordinate system defines the position using the signed curvilinear abscissa $s$, which is the length of the path from the origin of the curve to the point on the curve closest to the center of gravity, and the signed distance from the curve $e_y$, which is the lateral distance from the closest point on the curve to the vehicle. The Frenet reference frame, which is displayed in Figure 2.1, is therefore considered a moving reference frame since the frame moves with a tangent and normal vector to the curve. The Frenet system is rotated about the tangent angle $\theta$ compared to the Cartesian coordinate system.

The motion inertial coordinate system R, indicated in Figure 2.1, can be transformed to the Frenet reference frame T by the relationship for absolute velocities in the moving reference frame as described in [MicaelliSamson94]. Consider the absolute position of the center of gravity of the vehicle in R with

$$\boldsymbol{r}_{\mathrm{R}} = [x, y, 0]^{\mathsf{T}} \,. \tag{2.7}$$

Differentiation yields the absolute velocity of the vehicle in R

$$\boldsymbol{v}_{\mathrm{R}} = \dot{\boldsymbol{r}}_{\mathrm{R}} = [\dot{x}, \dot{y}, 0]^{\mathsf{T}} \,. \tag{2.8}$$

The absolute velocity can be expressed in the Frenet reference frame T by applying the transformation matrix $\boldsymbol{C}_{\mathrm{TR}}$, such that

$$\boldsymbol{v}_{\mathrm{T}} = \boldsymbol{C}_{\mathrm{TR}} \boldsymbol{v}_{\mathrm{R}}, \tag{2.9}$$

where the transformation matrix is the rotation matrix around the axis orthogonal to the $(x,\, y)$-plane by angle $\theta$, and is given by

$$\boldsymbol{C}_{\mathrm{TR}} = \begin{bmatrix} \cos(\theta) & \sin(\theta) & 0 \\ -\sin(\theta) & \cos(\theta) & 0 \\ 0 & 0 & 1 \end{bmatrix} . \tag{2.10}$$

This transforms the velocity from R to T. Furthermore, the absolute velocity in T can also be determined by

$$\boldsymbol{v}_{\mathrm{T}} = \boldsymbol{v}_{0'\mathrm{T}} + \boldsymbol{\omega}_{\mathrm{T}} \times \boldsymbol{r}'_{\mathrm{T}} + \boldsymbol{v}'_{\mathrm{T}}, \tag{2.11}$$

where $\boldsymbol{v}_{0'\mathrm{T}} = [\dot{s}, 0, 0]^\mathsf{T}$ is the velocity of the origin of T, $\boldsymbol{r}'_\mathrm{T} = [0, e_y, 0]^\mathsf{T}$ is the relative position in T, and $\boldsymbol{v}'_\mathrm{T} = [0, \dot{e}_y, 0]^\mathsf{T}$ is the relative velocity in T. The rotation velocity vector of T with respect to R is given by $\boldsymbol{\omega}_\mathrm{T} = \left[0, 0, \dot{\theta}\right]^\mathsf{T} = [0, 0, \kappa(s)\dot{s}]^\mathsf{T}$. The curvature of the centerline of the race track, $\kappa(s)$, is defined as

$$\kappa(s) = \frac{1}{r(s)} \tag{2.12}$$

with the radius of the curve of the centerline $r(s)$. Substituting Equation 2.9 into Equation 2.11 and rearranging the equation yields

$$\begin{bmatrix} \dot{s} \\ \dot{e_y} \\ 0 \end{bmatrix} = \begin{bmatrix} \cos(\theta) & \sin(\theta) & 0 \\ -\sin(\theta) & \cos(\theta) & 0 \\ 0 & 0 & 1 \end{bmatrix} \begin{bmatrix} \dot{x} \\ \dot{y} \\ 0 \end{bmatrix} - \begin{bmatrix} 0 \\ 0 \\ \kappa(s)\dot{s} \end{bmatrix} \times \begin{bmatrix} 0 \\ e_y \\ 0 \end{bmatrix}. \tag{2.13}$$

Solving for $\dot{s}$ and $\dot{e}_y$ results in

$$\dot{s} = \frac{\dot{x}\cos(\theta) + \dot{y}\sin(\theta)}{1 - \kappa(s)\mathrm{e}_y}, \tag{2.14}$$

$$\dot{e}_y = -\dot{x}\sin(\theta) + \dot{y}\cos(\theta). \tag{2.15}$$

Plugging Equation 2.2 and Equation 2.3 into Equation 2.14 and Equation 2.15 yields

$$\dot{s} = \frac{v\cos(\psi + \beta - \theta)}{1 - \kappa(s)\mathrm{e}_y}, \tag{2.16}$$

$$\dot{e}_y = v\sin(\psi + \beta - \theta). \tag{2.17}$$

The orientation error between the heading angle $\psi$ and the tangent angle $\theta(s)$ is defined as

$$e_\psi = \psi - \theta(s). \tag{2.18}$$

Differentiation yields

$$\dot{e_\psi} = \dot{\psi} - \dot{\theta}(s) = \dot{\psi} - \kappa(s)\dot{s}, \tag{2.19}$$

$$= v\frac{\cos(\beta)}{l_\mathrm{f} + l_\mathrm{r}}\tan(\delta_\mathrm{f}) - \kappa(s)\frac{v\cos(e_\psi + \beta)}{1 - \kappa(s)\mathrm{e}_y}, \tag{2.20}$$

$$= \frac{v}{l_\mathrm{r}}\sin(\beta) - \kappa(s)\dot{s}. \tag{2.21}$$

Using the above definitions, it is possible to write the kinematic bicycle model in the Frenet reference system [Rajamani12]

$$\dot{s} = v\frac{\cos(e_\psi + \beta)}{1 - e_y\kappa(s)}, \tag{2.22}$$

$$\dot{e_y} = v\sin(e_\psi + \beta), \tag{2.23}$$

$$\dot{e_\psi} = \frac{v}{l_\mathrm{f}}\sin(\beta) - \kappa(s)v\frac{\cos(e_\psi + \beta)}{1 - e_y\kappa(s)}, \tag{2.24}$$

$$\dot{v} = a, \tag{2.25}$$

$$\beta = \arctan\left(\frac{l_\mathrm{r}}{l_\mathrm{f} + l_\mathrm{r}}\tan(\delta_\mathrm{f})\right). \tag{2.26}$$

Finally, the above equations of the kinematic bicycle model have to be discretized for the MPC controller. The discretization is achieved by the forward Euler method with time step $T$:

$$s_{k+1} = s_k + T\left(v_k\frac{\cos(e_{\psi,k} + \beta_k)}{1 - e_{y,k}\kappa(s)}\right), \tag{2.27}$$

$$e_{y,k+1} = e_{y,k} + T\left(v_k\sin(e_{\psi,k} + \beta_k)\right), \tag{2.28}$$

$$e_{\psi,k+1} = e_{\psi,k} + T\left(\frac{v_k}{l_f}\sin(\beta_k) - \kappa(s)v_k\frac{\cos(e_{\psi,k} + \beta_k)}{1 - e_{y,k}\kappa(s)}\right), \tag{2.29}$$

$$v_{k+1} = v_k + T\cdot a_k, \tag{2.30}$$

$$\beta_k = \arctan\left(\frac{l_\mathrm{r}}{l_\mathrm{f} + l_\mathrm{r}}\tan(\delta_{\mathrm{f},k})\right). \tag{2.31}$$

For the remainder of the thesis, Equation 2.27 through Equation 2.31 are referred to as

$$\boldsymbol{x}_{k+1} = f_\mathrm{kin}(\boldsymbol{x}_k, \boldsymbol{u}_k), \tag{2.32}$$

where

$$\boldsymbol{x}_k = [s_k, e_{y,k}, e_{\psi,k}, v_k]^\mathrm{T} \tag{2.33}$$

is the vehicle state and

$$\boldsymbol{u}_k = [a_k, \delta_{\mathrm{f},k}]^\mathrm{T} \tag{2.34}$$

is the input vector.

## 2.2 Dynamic Bicycle Model

As discussed in section 2.1, the kinematic bicycle model does not sufficiently model the vehicle dynamics at high velocities. Therefore, a dynamic bicycle model is introduced in this section, which takes into account the forces acting on the tires. The dynamics of this model are defined in a body fixed frame. The states are the longitudinal and lateral velocity $v_x$ and $v_y$, respectively, and the yaw rate $r = \dot{\psi}$. The dynamic bicycle model states and the acting forces are shown in Figure 2.2. The Newton-Euler equations are derived in [KongEtAl15] and are given as

$$\dot{v}_x = a + \dot{\psi} v_y, \tag{2.35}$$

$$\dot{v}_y = \frac{1}{m}\left(F_{y,\mathrm{f}} \cos(\delta_\mathrm{f}) + F_{y,\mathrm{r}}\right) - \dot{\psi} v_x, \tag{2.36}$$

$$\ddot{\psi} = \frac{1}{I_\mathrm{z}}\left(l_\mathrm{f} F_{y,\mathrm{f}} - l_\mathrm{r} F_{y,\mathrm{r}}\right). \tag{2.37}$$

where $F_{y,\mathrm{f}}$ and $F_{y,\mathrm{r}}$ are the front and rear tire forces, respectively, $m$ is the mass of the vehicle, and $I_\mathrm{z}$ is the moment of inertia of the vehicle.

The tire forces $F_{y,\mathrm{f}}$ and $F_{y,\mathrm{r}}$ are related to the lateral slipping angles $\alpha_\mathrm{f}$ and $\alpha_\mathrm{r}$ of the front and the rear tire, respectively. Commonly, the tire forces are approximated by the Pacejka function from [BakkerNyborgPacejka87], often referred to as the magic formula

$$f_\mathrm{P}(\alpha) = D \sin(C \arctan(B\alpha)), \tag{2.38}$$

with the following Pacejka coefficients: the stiffness factor $B$, the shape factor $C$ and the peak factor $D$. Table 2.1 shows Pacejka coefficients for dry and snowy road coefficients, which are displayed in Figure 2.3. The figure illustrates the saturation of the tires at a certain slip angle. Beyond this slip angle the tire forces decrease and the vehicle starts drifting. Furthermore, the displayed coefficients also demonstrate expected behavior since the lateral tire forces of a vehicle on a snowy road start to decrease at a smaller slip angle and also decrease at a higher rate.

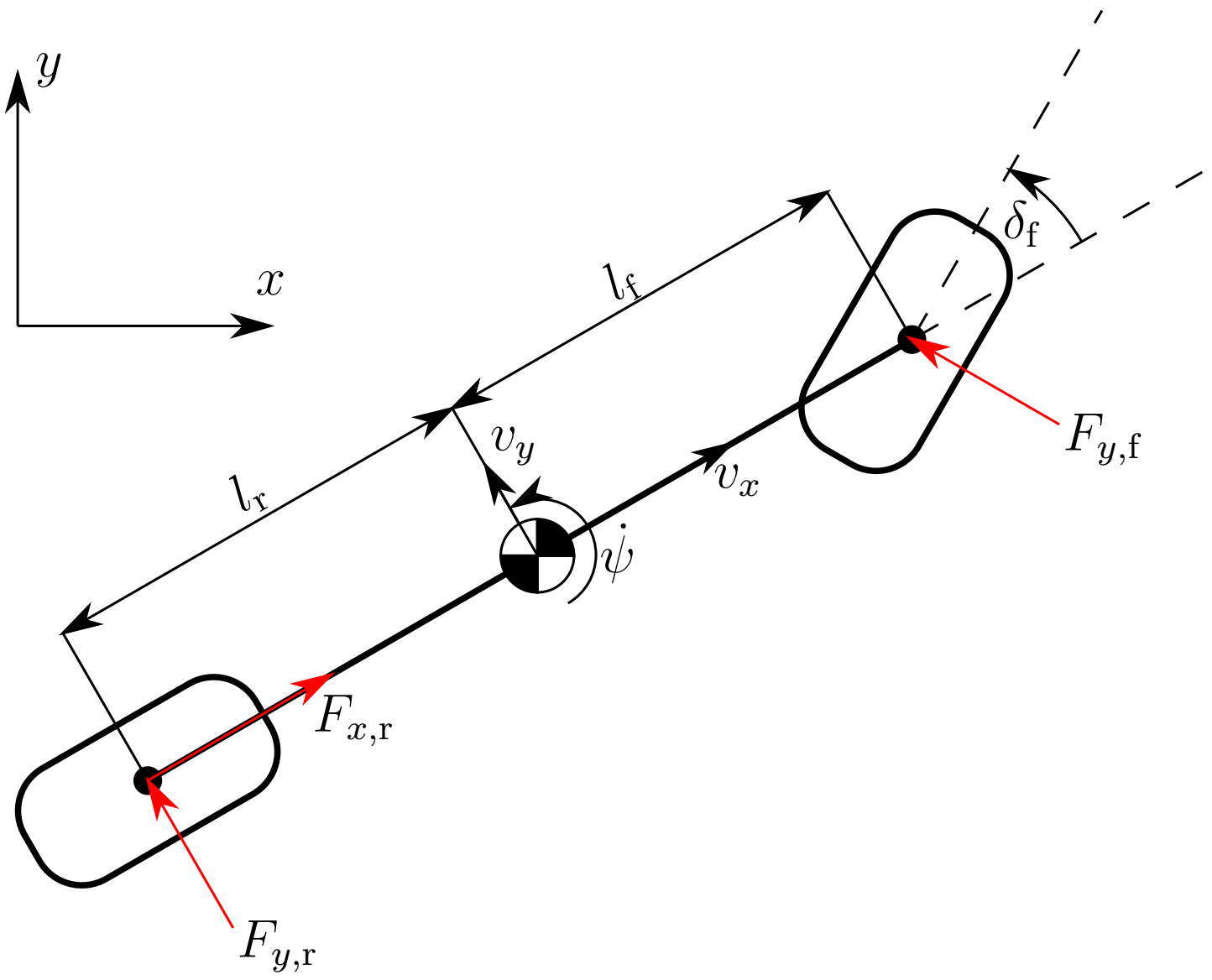

Figure 2.2: Dynamic bicycle model with the acting forces.

Table 2.1: Typical Pacejka coefficients for different road conditions.

| Road condition | Coefficent | | |
|---|---|---|---|
| | $B$ | $C$ | $D$ |
| Dry | 10.00 | 1.90 | 1.00 |
| Snowy | 12.00 | 2.30 | 0.82 |

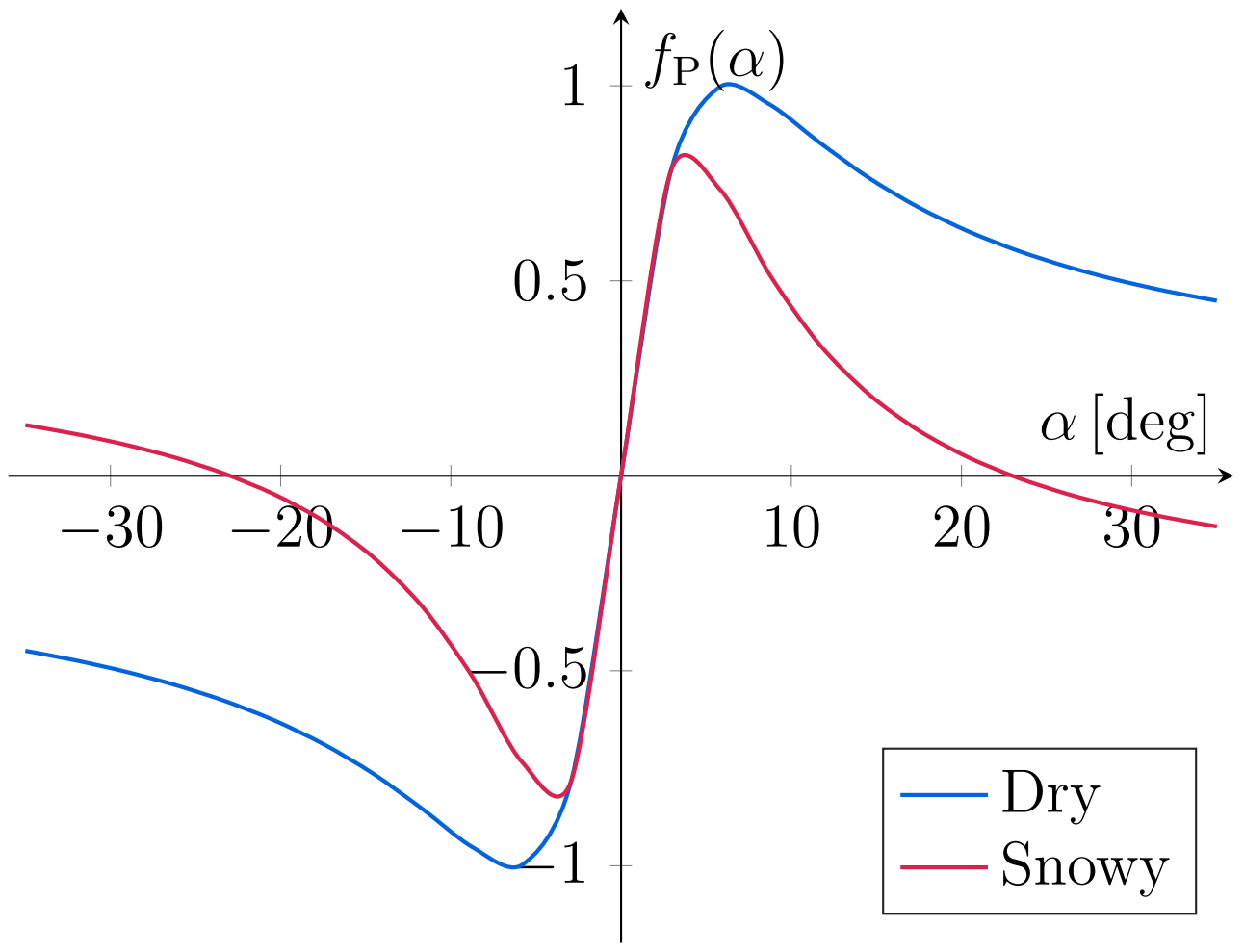

Figure 2.3: Pacejka functions for different road conditions.

The lateral tire forces $F_{y,\mathrm{f}}$ and $F_{y,\mathrm{r}}$ can then be expressed using the Pacejka function as

$$F_{y,\mathrm{f}} = -\frac{1}{2} \cdot m \cdot g \cdot \mu \cdot f_{\mathrm{P}}(\alpha_{\mathrm{f}}), \tag{2.39}$$

$$F_{y,\mathrm{r}} = -\frac{1}{2} \cdot m \cdot g \cdot \mu \cdot f_{\mathrm{P}}(\alpha_{\mathrm{r}}), \tag{2.40}$$

where $g$ is the gravitational constant and $\mu$ is the road-tire friction, such that the slip angles follow from the geometric relationships

$$\alpha_\mathrm{f} = \arctan\left(\frac{v_y + l_\mathrm{f}\dot{\psi}}{|v_x|}\right) - \delta_\mathrm{f}, \tag{2.41}$$

$$\alpha_\mathrm{r} = \arctan\left(\frac{v_y - l_\mathrm{r}\dot{\psi}}{|v_x|}\right). \tag{2.42}$$

The dynamic bicycle model is able to model the behavior of a vehicle at higher speeds and during different road conditions with more accuracy than the kinematic bicycle model. However, this is only the case if the Pacejka coefficients have been carefully selected or measured.

From Figure 2.2 $\dot{x}$ and $\dot{y}$ for the dynamic bicycle model can be determined as

$$\dot{x} = v_x \cos(\psi) - v_y \sin(\psi), \tag{2.43}$$

$$\dot{y} = v_x \sin(\psi) + v_y \cos(\psi). \tag{2.44}$$

The addition theorems for sine and cosine yield

$$\sin(\theta)\cos(\psi) - \cos(\theta)\sin(\psi) = \sin(\theta - \psi) = \sin(-e_\psi) = -\sin(e_\psi), \tag{2.45}$$

$$\cos(\theta)\cos(\psi) + \sin(\theta)\sin(\psi) = \cos(\theta - \psi) = \cos(-e_\psi) = \cos(e_\psi). \tag{2.46}$$

Then substituting $\dot{x}$ and $\dot{y}$ in Equation 2.14 and Equation 2.15 with Equation 2.43 and Equation 2.44, respectively, and using Equation 2.45 and Equation 2.46 results in

$$\dot{s} = \frac{v_x \cos(e_\psi) - v_y \sin(e_\psi)}{1 - e_y \kappa(s)}, \tag{2.47}$$

$$\dot{e_y} = v_x \sin(e_\psi) + v_y \cos(e_\psi). \tag{2.48}$$

The orientation error $e_\psi$ for the dynamic bicycle model is received by using Equation 2.47 and inserting it into Equation 2.19:

$$\dot{e}_\psi = \dot{\psi} - \kappa(s)\dot{s}, \tag{2.49}$$

$$= \dot{\psi} - \kappa(s)\frac{v_x \cos(e_\psi) - v_y \sin(e_\psi)}{1 - e_y \kappa(s)}. \tag{2.50}$$

Using all of the above yields the dynamic bicycle model in the Frenet reference frame

$$\dot{s} = \frac{v_x \cos(e_\psi) - v_y \sin(e_\psi)}{1 - e_y \kappa(s)}, \tag{2.51}$$

$$\dot{e}_y = v_x \sin(e_\psi) + v_y \cos(e_\psi), \tag{2.52}$$

$$\dot{e}_\psi = \dot{\psi} - \kappa(s) \frac{v_x \cos(e_\psi) - v_y \sin(e_\psi)}{1 - e_y \kappa(s)}, \tag{2.53}$$

$$\ddot{\psi} = \frac{1}{I_\mathrm{z}} \left( l_\mathrm{f} F_\mathrm{f} - l_\mathrm{r} F_\mathrm{r} \right), \tag{2.54}$$

$$\dot{v}_x = a + \dot{\psi} v_y, \tag{2.55}$$

$$\dot{v}_y = \frac{1}{m} \left( F_\mathrm{f} \cos(\delta_\mathrm{f}) + F_\mathrm{r} \right) - \dot{\psi} v_x. \tag{2.56}$$

Similarly as before, the continuous system is discretized by the forward Euler method with time step $T$

$$s_{k+1} = s_k + T \cdot \left( \frac{v_{x,k} \cos(e_{\psi,k}) - v_{y,k} \sin(e_{\psi,k})}{1 - e_{y,k} \kappa(s)} \right), \tag{2.57}$$

$$e_{y,k+1} = e_{y,k} + T \cdot \left( v_{x,k} \sin(e_{\psi,k}) + v_{y,k} \cos(e_{\psi,k}) \right), \tag{2.58}$$

$$e_{\psi,k+1} = e_{\psi,k} + T \cdot \left( r_k - \kappa(s) \frac{v_{x,k} \cos(e_{\psi,k}) - v_{y,k} \sin(e_{\psi,k})}{1 - e_{y,k} \kappa(s)} \right), \tag{2.59}$$

$$r_{k+1} = r_k + T \cdot \left( \frac{1}{I_\mathrm{z}} \left( l_\mathrm{f} F_\mathrm{f} - l_\mathrm{r} F_\mathrm{r} \right) \right), \tag{2.60}$$

$$v_{x,k+1} = v_{x,k} + T \cdot \left( a_k + r_k v_{y,k} \right), \tag{2.61}$$

$$v_{y,k+1} = v_{y,k} + T \cdot \left( \frac{1}{m} \left( F_\mathrm{f} \cos(\delta_{\mathrm{f},k}) + F_\mathrm{r} \right) - r_k v_{x,k} \right), \tag{2.62}$$

where $r$ is the change of the heading angle or yaw rate.

Equation 2.57 through Equation 2.62 is the discrete dynamic bicycle model parameterized by the Pacejka coefficients $B$, $C$, and $D$. For the remainder of the thesis they are referred to as

$$\boldsymbol{x}_{k+1} = f_\mathrm{dyn}(\boldsymbol{x}_k, \boldsymbol{u}_k; B, C, D), \tag{2.63}$$

where

$$\boldsymbol{x}_k = \left[ s_k, e_{y,k}, e_{\psi,k}, r_k, v_{x,k}, v_{y,k} \right]^\mathsf{T} \tag{2.64}$$

is the vehicle state and

$$\boldsymbol{u}_k = \left[ a_k, \delta_{\mathrm{f},k} \right]^\mathsf{T} \tag{2.65}$$

is the input vector.

# Chapter 3

# Learning Model Predictive Control

## 3.1 Model Predictive Control Theory

A model predictive controller (MPC), or a receding horizon controller (RHC), is based on the idea of using an optimization procedure to obtain the optimal control action by using the system's predicted evolution. The following is based on [BorrelliBemporadMorari17] and serves as a short introduction of the most important fundamentals and preliminaries for the model predictive controller. For a more detailed description and examples please refer to the book by [BorrelliBemporadMorari17].

### 3.1.1 Notation

In this subsection some of the mathematical notation, that is used in the following chapters, is introduced. In general, the evolution of the discrete nonlinear system $f(\cdot,\cdot)$ from time step $t$ to time step $t+1$ is written as

$$\boldsymbol{x}(t+1) = f(\boldsymbol{x}(t), \boldsymbol{u}(t)), \tag{3.1}$$

where $\boldsymbol{x}$ is the state vector and $\boldsymbol{u}$ is the input vector. Subsequently, this will be written as

$$\boldsymbol{x}_{t+1} = f(\boldsymbol{x}_t, \boldsymbol{u}_t). \tag{3.2}$$

Furthermore, in the context of MPC it is important to differentiate among the measured states of the system and the predicted states of the controller. Therefore, the measured state at time $t$ is $\boldsymbol{x}_t$ and the predicted state at time $t+k$, predicted at time $t$, is $\boldsymbol{x}_{t+k|t}$, which is obtained by propagating from the state $\boldsymbol{x}_{t|t} = \boldsymbol{x}_t$ using the system model $f(\cdot,\cdot)$. Similarly, $\boldsymbol{u}_{t+k|t}$ is the input $\boldsymbol{u}$ at time $t+k$, computed at time $t$.

### 3.1.2 Preliminiaries

This section introduces necessary concepts and different optimization problems and properties, which MPC is based on and are also needed for the understanding of MPC.

#### Convexity

The property of convexity is of great significance for optimization problems, since convexity enables globally optimal solutions. Futhermore, there exist solvers, which are able to solve optimization problems efficiently and fast, as long as they are convex. A set $S \in \mathbb{R}^n$ is convex if

$$\lambda z_1 + (1-\lambda) z_2 \in S \ \forall z_1, z_2 \in S, \lambda \in [0,1]. \tag{3.3}$$

In other words, for a linear combination $\tilde{z}$ of two points $z_1 \in S$ and $z_2 \in S$, where $\tilde{z} \in S$ and $\lambda \in [0,1]$, the set $S$ is convex. An example of a convex and non-convex set are displayed in Figure 3.1 and Figure 3.2, respectively. While all linear combinations of $z_1$ and $z_2$ are in the set $S_{\text{conv}}$, this is not the case for $S_{\text{non}}$, which is why the set is non-convex. Moreover, a function $f \colon S \to \mathbb{R}$ is convex if

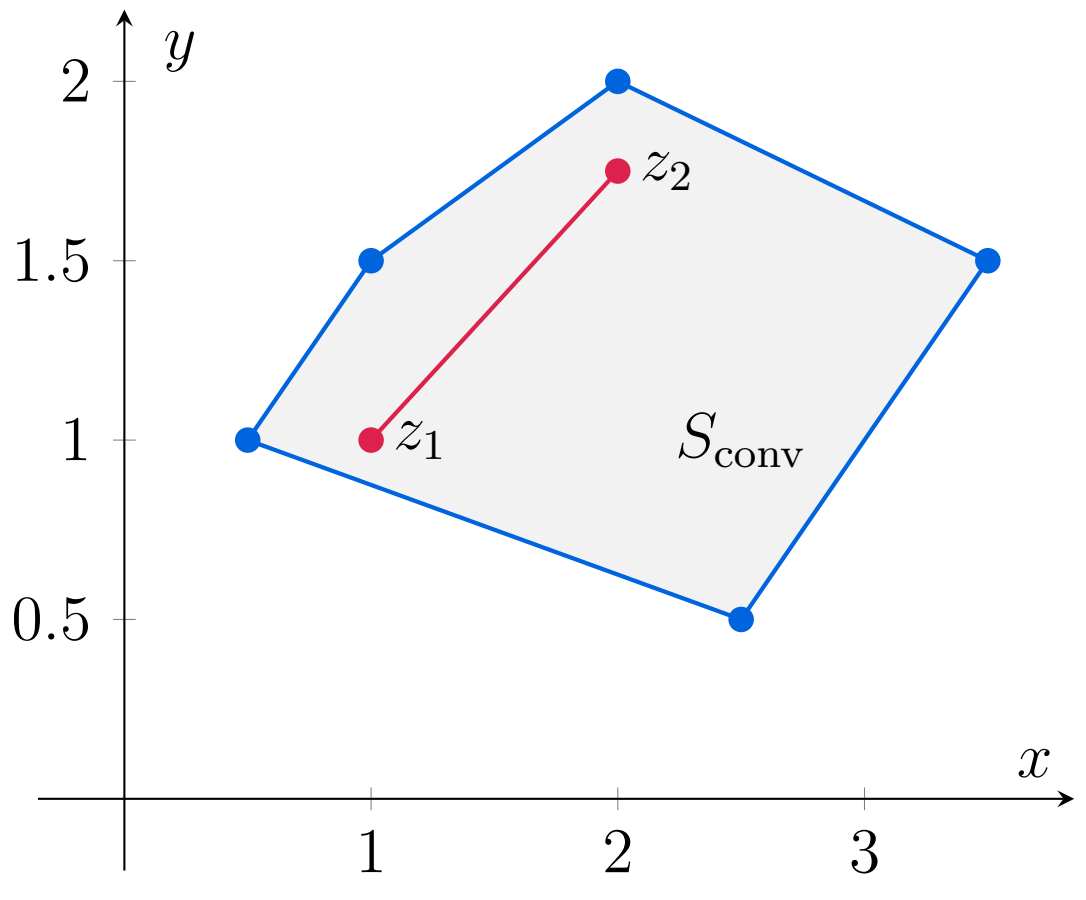

Figure 3.1: Convex set

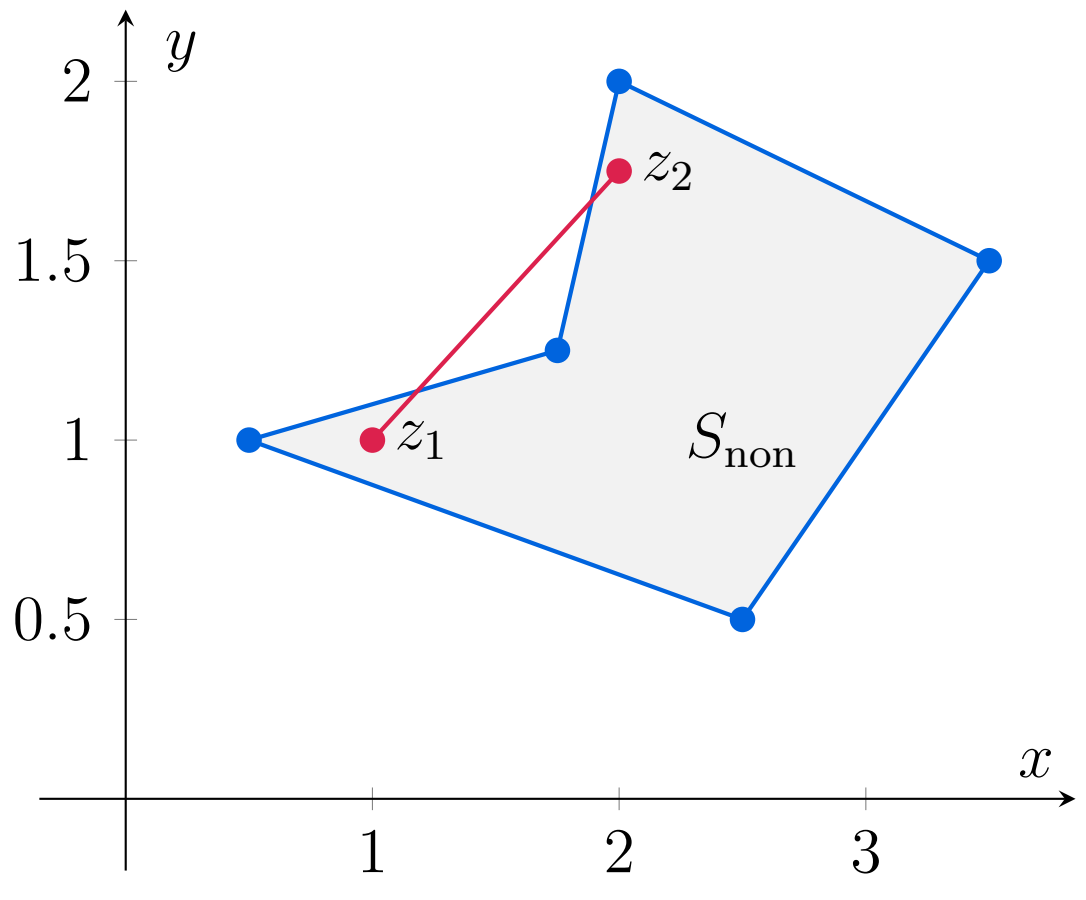

Figure 3.2: Non-convex set

$S$ is a convex set and

$$f(\lambda z_1 + (1-\lambda) z_2) \leq \lambda f(z_1) + (1-\lambda) f(z_2) \ \forall z_1, z_2 \in S, \lambda \in [0,1], \tag{3.4}$$

illustrated in Figure 3.3.

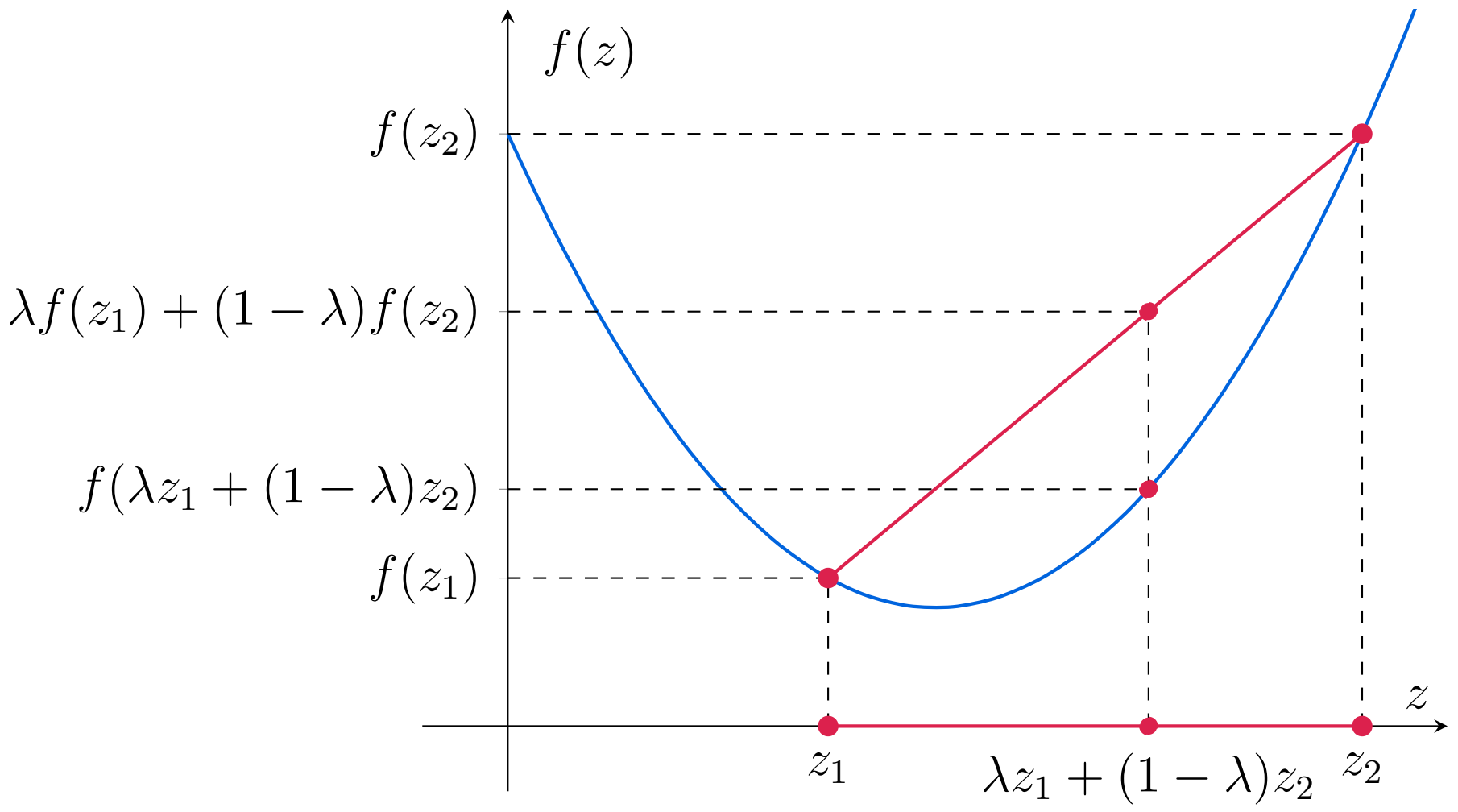

Figure 3.3: Example of a convex function.

### Optimization Problems

A continuous optimization problem in its standard form is given by

$$f^* = \min_z \; f(z) \tag{3.5}$$

$$\text{subj. to } g_i(z) \leq 0, \;\; \forall i \in \{1, ..., m\}, \tag{3.6}$$

$$h_j(z) = 0, \;\; \forall j \in \{1, ..., p\}, \tag{3.7}$$

$$z \in Z, \tag{3.8}$$

where

- $f(z) : \mathbb{R}^n \rightarrow \mathbb{R}$ is the objective or cost function, which is to be minimized over $z \in \mathbb{R}^n$,
- $g_i(z) \leq 0$ are called inequality constraints and $g_i(z)$ are real-valued functions defined over $\mathbb{R}^n$,
- $h_j(z) = 0$ are called equality constraints and $h_j(z)$ are real-valued functions defined over $\mathbb{R}^n$,
- the problem domain $Z$ is a subset of the $\mathbb{R}^n$,
- $f^*$ is the optimal value for the optimization problem.

The optimal value or the least possible cost $f^*$ is given by the optimal solution or optimizer $z^*$, which is defined as

$$f(z) \geq f(z^*) = f^*, \;\; \forall z \in Z, \text{with } z^* \in Z. \tag{3.9}$$

A point $\bar{z} \in \mathbb{R}^n$ is feasible for the above optimization problem if

(i) $\bar{z} \in Z$

(ii) $\bar{z}$ satisfies all inequality and equality constraints, such that $g_i(\bar{z}) \leq 0,\ \forall i \in \{1, ..., m\}$ and $h_j(\bar{z}) = 0,\ \forall j \in \{1, ..., p\}$.

If there is no point $\bar{z}$, that satisfies the constraints, the problem is infeasible. Furthermore, the optimization problem is referred to as unconstrained if $m = p = 0$.

An optimization problem is convex, if the cost function $f$ is convex on $Z$ and the set $S$ of feasible vectors is convex. Convex optimization problems are of significant importance, because for all local optimizers applies that they are also global optimal solutions. This is the reason why convexity plays a major role in the solution of continuous optimization problems. However, the convexity of an optimization problem does not allow any claims on the existence of the solution to the optimization problem. A linear program (LP) is a convex optimization problem. The cost and constraints of the linear program are affine and the linear program in the standard form is given as

$$\min_{z} c^{\mathsf{T}} z \tag{3.10}$$

$$\text{subj. to } Gz \leq w, \tag{3.11}$$

where $z \in \mathbb{R}^n$, $c \in \mathbb{R}^n$, $G \in \mathbb{R}^{m \times n}$, $w \in \mathbb{R}^m$. If the LP is feasible and the solution is bounded, then the optimal solution always lies on the boundary of the feasible constraints. Another important class of optimization problems are the quadratic programs (QP). A given continuous optimization problem is a quadratic program, if the constraint functions are affine and the cost function is a convex quadratic function. The standard form of the QP is given as

$$\min_{z} \frac{1}{2} z^{\mathsf{T}} H z + q^{\mathsf{T}} z \tag{3.12}$$

$$\text{subj. to } Gz \leq w, \tag{3.13}$$

where $z \in \mathbb{R}^n$, $H = H^{\mathsf{T}} \succ 0 \in \mathbb{R}^{n \times n}$, $q \in \mathbb{R}^n$, $G \in \mathbb{R}^{m \times n}$, $w \in \mathbb{R}^m$. Assuming that there exists a feasible solution, the optimal solution either lies on the boundary or inside of the polyhedron constructed by the constraints. However, the optimization problem in chapter 4 for the autonomous racing agent is a nonlinear program. In a nonlinear program at least one of the objective function, inequality constraints or equality constraints is a nonlinear function. If there exists a feasible solution, then global optimality is not guaranteed and computed solutions might only be locally optimal solutions. In general, non-convex optimization problems, like the nonlinear programs, are solved by iteratively solving convex sub-problems.

### 3.1.3 Model Predictive Control

Let the following discrete time system be given:

$$\boldsymbol{x}_{t+1} = f(\boldsymbol{x}_t, \boldsymbol{u}_t) \tag{3.14}$$

with time $t$, the system state $\boldsymbol{x} \in \mathbb{R}^n$ and the control input $\boldsymbol{u} \in \mathbb{R}^m$, subject to the constraints

$$\boldsymbol{x}_t \in \mathcal{X},\ \ \boldsymbol{u}_t \in \mathcal{U}, \forall t \in \mathbb{Z}_{0+}, \tag{3.15}$$

where $\mathcal{X}$ and $\mathcal{U}$ are sets, that are assumed to include the origin. Additionally, a cost function $J^*_{0\rightarrow\infty}(\boldsymbol{x}_0)$ that evaluates the controller's performance at all time steps, is defined as

$$J^*_{0\rightarrow\infty}(\boldsymbol{x}_0) = \min_{\boldsymbol{u}_0,\boldsymbol{u}_1,\dots} \sum_{k=0}^{\infty} q(\boldsymbol{x}_k, \boldsymbol{u}_k), \tag{3.16}$$

where $h(\cdot,\cdot)$ is the stage cost, which quantifies the cost for the system being in state $\boldsymbol{x}$ and applying input $\boldsymbol{u}$. Minimizing the above cost functions for the optimal control inputs $\boldsymbol{u}_0^*, \boldsymbol{u}_1^*, \dots$ and satisfying the state and inputs constraints and the system dynamics, yields the infinite time horizon control problem

$$J^*_{0\rightarrow\infty}(\boldsymbol{x}_t) = \min_{\boldsymbol{u}_0,\boldsymbol{u}_1,\dots} \sum_{k=0}^{\infty} q(\boldsymbol{x}_k, \boldsymbol{u}_k) \tag{3.17}$$

$$\text{subj. to } \boldsymbol{x}_{k+1} = f(\boldsymbol{x}_k, \boldsymbol{u}_k),\ \forall k \geq 0 \tag{3.18}$$

$$\boldsymbol{x}_0 = \boldsymbol{x}(0) \tag{3.19}$$

$$\boldsymbol{x}_k \in \mathcal{X},\ \forall k \geq 0 \tag{3.20}$$

$$\boldsymbol{u}_k \in \mathcal{U},\ \forall k \geq 0, \tag{3.21}$$

which is only solvable under certain conditions. Therefore, the problem is reformulated as a constrained finite time optimal control (CFTOC) problem with horizon $N$, which is repeatedly solved. Using the system model in the controller, determine an optimal open-loop input sequence according to the specified objective function. Yet, only the first input $\boldsymbol{u}_{t|t}$ is applied to the system. Then the CFTOC is solved again for the next time step $t+1$ over a shifted horizon. The resulting controller is referred to as Receding Horizon Controller (RHC) or Model Predictive Controller (MPC).

The CFTOC problem is then given as

$$J^*_{t\to t+N}(\boldsymbol{x}_t) = \min_{U_{t\to t+N|t}} p(\boldsymbol{x}_{t+N|t}) + \sum_{k=t}^{t+N-1} q(\boldsymbol{x}_{k|t}, \boldsymbol{u}_{k|t}) \tag{3.22}$$

$$\text{subj. to } \boldsymbol{x}_{k+1|t} = f(\boldsymbol{x}_{k|t}, \boldsymbol{u}_{k|t}),\ \forall k \in \{t, ..., t+N-1\} \tag{3.23}$$

$$\boldsymbol{x}_{t|t} = \boldsymbol{x}_t \tag{3.24}$$

$$\boldsymbol{x}_{k|t} \in \mathcal{X},\ \forall k \in \{t, ..., t+N\} \tag{3.25}$$

$$\boldsymbol{u}_{k|t} \in \mathcal{U},\ \forall k \in \{t, ..., t+N-1\} \tag{3.26}$$

$$\boldsymbol{x}_{N|t} \in \mathcal{X}_f, \tag{3.27}$$

with the optimization variable $U_{t\to t+N|t} = \left\{u_{t|t}, ..., u_{t+N-1|t}\right\}$, the terminal cost $p(\cdot)$ and the terminal set $\mathcal{X}_f$.

Stability and feasibility are not necessarily ensured by the RHC, when solving over a finite horizon repeatedly. Therefore, the goal of the controller design is to yield a closed-loop behavior, that is similar to the one of the infinite horizon controller. The terminal cost is used to approximate the initial cost function from the infinte time horizon control problem for the steps from $N \to \infty$. The terminal set $\mathcal{X}_f$ is chosen, such that the system mimics the infinite time horizon case. The conditions on the choice of the terminal cost $p(\cdot)$ and the terminal constraint set $\mathcal{X}_f$, for which the closed-loop system is stable and recursively feasible, are given in [BorrelliBemporadMorari17].

## 3.2 Learning Model Predictive Control Formulation

In the past control systems, that autonomously perform a repetitive task, have been studied extensively. The execution of one task is often referred to as one "iteration". The area of research, that focuses on improving closed-loop tracking performance using previous iterations, is called Iterative Learning Control (ILC). Previously ILC has been limited to reference-tracking and fixed time processes. This section introduces a reference-free learning model predictive controller, which is able to learn from previous iterations and overcomes the shortcomings of ILC. The presented controller enables convergence to a locally optimal trajectory for nonlinear systems or even globally optimal trajectory for linear systems with variable time duration using a terminal safe set and a terminal cost function. The learning model predictive controller (LMPC) has been introduced in [RosoliaBorrelli17a] and extended in [RosoliaBorrelli17b]. The following serves as a summary of the two sources and short overview of the LMPC.

Consider the same discrete time system as in Equation 3.14

$$\boldsymbol{x}_{t+1} = f(\boldsymbol{x}_t, \boldsymbol{u}_t), \tag{3.28}$$

subject to the constraints

$$\boldsymbol{x}_t \in \mathcal{X},\ \ \boldsymbol{u}_t \in \mathcal{U}, \forall t \in \mathbb{Z}_{0+}. \tag{3.29}$$

At iteration $j$ the matrices

$$\boldsymbol{U}^j = \left[\boldsymbol{u}_0^j, \boldsymbol{u}_1^j, ..., \boldsymbol{u}_t^j, ...\right], \tag{3.30}$$

$$\boldsymbol{X}^j = \left[\boldsymbol{x}_0^j, \boldsymbol{x}_1^j, ..., \boldsymbol{x}_t^j, ...\right] \tag{3.31}$$

collect the inputs and the corresponding states during iteration $j$. It is assumed, that each iteration $j$ starts with the system being in the same initial state,

$$\boldsymbol{x}_0^j = \boldsymbol{x}_S,\ \forall j \geq 0. \tag{3.32}$$

The goal of the learning model predictive controller is to solve the following infinite horizon optimal control problem:

$$J^*_{0\to\infty}(\boldsymbol{x}_S) = \min_{\boldsymbol{u}_0, \boldsymbol{u}_1, \dots} \sum_{k=0}^{\infty} h(\boldsymbol{x}_k, \boldsymbol{u}_k) \tag{3.33}$$

$$\text{subj. to } \boldsymbol{x}_{k+1} = f(\boldsymbol{x}_k, \boldsymbol{u}_k),\ \forall k \geq 0, \tag{3.34}$$

$$\boldsymbol{x}_0 = \boldsymbol{x}_S, \tag{3.35}$$

$$\boldsymbol{x}_k \in \mathcal{X},\ \forall k \geq 0, \tag{3.36}$$

$$\boldsymbol{u}_k \in \mathcal{U},\ \forall k \geq 0. \tag{3.37}$$

Here, the stage cost $h(\cdot, \cdot)$ is continuous and satisfies

$$h(\boldsymbol{x}_f, 0) = 0\ \ \wedge h(\boldsymbol{x}_t^j, \boldsymbol{u}_t^j) \succ 0,\ \forall \boldsymbol{x}_t^j \in \mathbb{R}^n \setminus \{\boldsymbol{x}_f\}, \boldsymbol{u}_t^j \in \mathbb{R}^m \setminus \{0\}, \tag{3.38}$$

with the final state $\boldsymbol{x}_f$ being a feasible equilibrium for the unforced discrete time system,

$$f(\boldsymbol{x}_f, 0) = \boldsymbol{x}_f. \tag{3.39}$$

For the LMPC one iteration is the successful execution of the task, which is indicated by the state $\boldsymbol{x} \in \mathcal{X}_F$, where $\mathcal{X}_F$ is the set of goal states. Then the LMPC formulation follows from [RosoliaBorrelli17a], where the superscript $j$ indicates the iteration:

$$J^{j,*}_{t\to t+N}(\boldsymbol{x}_t^j) = \min_{U_{t\to t+N|t}} Q^{j-1}(\boldsymbol{x}_{t+N|t}) + \sum_{k=t}^{t+N-1} h(\boldsymbol{x}_{k|t}, \boldsymbol{u}_{k|t}) \tag{3.40}$$

$$\text{subj. to } \boldsymbol{x}_{k+1|t} = f(\boldsymbol{x}_{k|t}, \boldsymbol{u}_{k|t}),\ \forall k \in \{t, ..., t+N-1\}, \tag{3.41}$$

$$\boldsymbol{x}_{t|t} = \boldsymbol{x}_t^j, \tag{3.42}$$

$$\boldsymbol{x}_{k|t} \in \mathcal{X},\ \forall k \in \{t, ..., t+N\}, \tag{3.43}$$

$$\boldsymbol{u}_{k|t} \in \mathcal{U},\ \forall k \in \{t, ..., t+N-1\}, \tag{3.44}$$

$$\boldsymbol{x}_{N|t} \in \mathcal{SS}^{j-1}, \tag{3.45}$$

with the sampled safe set $\mathcal{SS}^j$ and the iteration cost $Q^j(\cdot)$, which are presented in the following. Assuming they are picked in the described way, it is guaranteed, that the cost is non-increasing at each iteration, that the state and input constraints are satisfied at iteration $j$ if they were satisfied at iteration $j-1$ and that the closed-loop equilibrium is asymptotically stable. For the complete proofs please refer to [RosoliaBorrelli17a] and [RosoliaBorrelli17b].

*Assumption* 1. At iteration $j = 1$ it is assumed, that $\mathcal{SS}^{j-1} = \mathcal{SS}^0$ is a non-empty set and that the trajectory $\boldsymbol{X}^0 \in \mathcal{SS}^0$ is feasible and convergent to $\boldsymbol{x}_f$.

### Sampled Safe Set

Previous successful trials form the sampled safe set at iteration $j$ :

$$\mathcal{SS}^j = \left\{ \bigcup_{i \in M^j} \bigcup_{t=0}^{\infty} x_t^i \right\}, \tag{3.46}$$

where $M^j$ is the set of indices, which successfully completed the $k$-th iteration:

$$M^j = \left\{ k \in [0, j] : \lim_{t \to \infty} x_t^k = x_F \right\}. \tag{3.47}$$

The set $\mathcal{SS}^j$ is a control invariant set, since:

$$\forall x \in \mathcal{SS}^j, \exists u \in \mathcal{U} : f(x, u) \in \mathcal{SS}^j. \tag{3.48}$$

### Iteration Cost

At time $t$ of the $j$-th iteration, the iteration cost is defined as

$$J_{t \to \infty}^j(x_t^j) = \sum_{k=t}^{t_j} h(x_k^j, u_k^j), \tag{3.49}$$

given by [RosoliaBorrelli17a], where $h(\cdot, \cdot)$ is the stage cost. The function $Q^j(\cdot)$ is then:

$$Q^j(x) = \begin{cases} \min_{(i,t) \in F^j(x)} J_{t \to \infty}^j(x_t^j), & \text{if } x \in \mathcal{SS}^j \\ \infty, & \text{else} \end{cases}, \tag{3.50}$$

where $F^j(x)$ is defined as:

$$F^j(x) = \left\{ (i, t) : i \in [0, \infty), t \geq 0 \text{ with } x = x_t^i,\ x_t^i \in \mathcal{SS}^j \right\}. \tag{3.51}$$

Intuitively, the function $Q^j(\cdot)$ assigns the minimum cost-to-go along the trajectories in $\mathcal{SS}^j$ to every point in the sampled safe set $\mathcal{SS}^j$.

# Chapter 4

# Learning Model Predictive Control for Autonomous Racing

## 4.1 Single-Agent Racing

The goal of the single-agent racing scenario is to have a single vehicle finish multiple laps on a given race track in minimum time, while meeting track boundary constraints. This is accomplished by a controller using LMPC. The LMPC for the racing scenario uses the previously introduced bicycle models to propagate the vehicle states.

This part of the chapter describes the adaptations to the nominal LMPC to the autonomos racing scenario in the single-agent case. These adaptations allow for the problem to be tractable and solved in real-time.

### 4.1.1 Safe Set Initialization

As previously stated, LMPC always requires a safe set in order to run. The safe set for the racing scenario is initialized by a path-following MPC controller, which follows the track at $e_{y,\text{ref}}$ = const. with a low reference velocity $v_{\text{ref}}$ = const. However, any other type of controller could be used to create the initial safe set. For example, another possibility to accomplish the path-following task could be a simple PID-controller.

The path-following controller creates feasible trajectories on the defined race track. As long as a low reference velocity is used, the assumptions of the presented kinematic bicycle model still hold. Therefore, the discretized kinematic bicycle model $f_{\text{kin}}(\boldsymbol{x}, \boldsymbol{u})$ can be used in the constraints of the MPC to describe the system's model. The complete MPC formulation of the path-following controller can then be stated as

$$J_t^*(\boldsymbol{x}_t, \boldsymbol{u}_{t\to t+N-1}) = \min_{U_{t\to t+N|t}} \sum_{k=t}^{t+N} \Big(w_v(v_k - v_{ref})^2 + w_{e,y}(e_{y,k} - e_{y,\text{ref}})^2\Big) + \sum_{k=t}^{t+N-1} \Big(w_a a_k^2 + w_{\delta,f}\delta_f^2\Big) + \sum_{k=t+1}^{t+N-1} \Big(w_{\Delta\delta,f}(\delta_{f,k} - \delta_{f,k-1})^2\Big) \tag{4.1}$$

$$\text{subj. to } \boldsymbol{x}_{k+1|t} = f_{\text{kin}}(\boldsymbol{x}_{k|t}, \boldsymbol{u}_{k|t}),\ \forall k \in \{t, ..., t+N-1\} \tag{4.2}$$

$$\boldsymbol{x}_{t|t} = \boldsymbol{x}_t \tag{4.3}$$

$$a_{\min} \le a_k \le a_{\max},\ \forall k \in \{t, ..., t+N-1\} \tag{4.4}$$

$$\delta_{f,\min} \le \delta_{f,k} \le \delta_{f,\max},\ \forall k \in \{t, ..., t+N-1\} \tag{4.5}$$

$$\boldsymbol{x}_{k|t} \in \mathcal{X},\ \forall k \in \{t, ..., t+N\} \tag{4.6}$$

with the system inputs acceleration $a$ and steering angle $\delta_f$ and the state vector $\boldsymbol{x}_t = [s_t, e_{y,t}, e_{\psi,t}, v_t]^\mathsf{T}$, the optimization variable $U_{t\to t+N|t} = \left\{u_{t|t}, ..., u_{t+N-1|t}\right\}$ and the weights $w_v$, $w_{e,y}$, $w_a$, $w_{\delta,f}$, and $w_{\Delta\delta,f}$, which are tuning parameters. The above constraints Equation 4.4 and Equation 4.5 are the input constraints on $a$ and $\delta_\text{f}$, respectively, which model the true behavior of the actuators. The initial constraint is given by Equation 4.3 and the state constraints are satisfied through Equation 4.2. The first term of the defined cost function Equation 4.1 penalizes deviation from the given references $v_\text{ref}$ and $e_{y,\text{f}}$. The second term lets the MPC favors small inputs for the acceleration and steering angle and the third term penalizes fast changing steering inputs.

### 4.1.2 Stage Cost

The stage cost for the racing problem is defined by a constant function, when the vehicle is located between the start and the finish line, and zero cost, when the car reached the final state $\mathcal{X}_f$:

$$h(\boldsymbol{x}_t, \boldsymbol{u}_t) = \begin{cases} 1, & \text{if } x \notin \mathcal{X}_f \\ 0, & \text{if } x \in \mathcal{X}_f \end{cases}, \tag{4.7}$$

where the final state is defined as

$$\mathcal{X}_f = \left\{\boldsymbol{x} \in \mathbb{R}^6 : e_1^\mathsf{T}\boldsymbol{x} = s > s_f\right\} \tag{4.8}$$

with $s_f$ being the length of the track.

### 4.1.3 State-Varying Safe Set

The sampled safe set grows unboundedly with every additional iteration. Therefore, limiting the number of states in the safe set decreases the computational complexity. This is the reason why a space-varying sampled safe set $\mathcal{VS}^j(x) \subset \mathcal{SS}$ is defined in [RicciutiRosoliaGonzales18] as

$$\mathcal{VS}^j(x) = \bigcup_{i=j-N_l}^{j} \bigcup_{t=\gamma^i(x)}^{\gamma^i(x)+N_p} x_t^i \tag{4.9}$$

where $N_l$ is the number of considered previous iterations, $N_p$ is the number of considered previous states per iteration, and $\gamma^i(x)$ is the time index $t$ of the $i$-th iteration, for which $x_{d+t}^i$ is closest to the current state $x$:

$$\gamma^i(x) = \arg\min_{t\geq 0} ||x_{d+t}^i - x||_2^2 \tag{4.10}$$

where $d$ is the offset from the current state. The parameters $N_l$, $N_p$ and $d$ are tuning parameters, which are chosen in a way, such that the safety and performance improvement guarantees still hold.

A contribution of this work is the introduction of a state-varying safe set $\mathcal{VS}^j(\boldsymbol{x})$. However, before introducing an enhanced sampled safe set, the set $\mathcal{T}^j(n)$ needs to be defined. This set indicates the $n$ iterations, which need the least number of time steps to reach the terminal set $\mathcal{X}_f$. Let

$$\mathcal{T}^j = \bigcup_{i=0}^{j} \bar{t}_i \tag{4.11}$$

be the set of time steps needed to complete each iteration up to iteration $j$ with

$$\bar{t}_j = \arg\min_t \{t+1 : x_{t+1}^j \in \mathcal{X}_F\} \tag{4.12}$$

being the time at which the transition between the the $j$-th and $j+1$-th iteration occurs. Then the number of time steps needed for the iteration, for which $n-1$ other iterations needed fewer time steps, is given by

$$\bar{t}^{j,n} = \min\left\{t : \left|\left\{l \in \mathcal{T}^j : l \leq t\right\}\right| = n\right\}, \tag{4.13}$$

where $|\{l \in \mathcal{T}^j : l \leq t\}|$ is the cardinality of the set $\{l \in \mathcal{T}^j : l \leq t\}$. For example, if $n = 2$, then $\bar{t}^{j,2}$ would give the number of time steps for the iteration, which needed the second least steps. Finally, $\mathcal{T}^j(n)$ is defined as:

$$\mathcal{T}^j(n) = \left\{i \in \mathbb{N} : i < j \wedge \bar{t}^{j,n} \geq \bar{t}^{i,n}\right\}. \tag{4.14}$$

Intuitively, $\mathcal{T}^j(n)$ collects the indices to the iterations, with the $n$-least number of time steps needed to reach the terminal set.

Now, instead of selecting only states from the previous $N_l$ iterations, it might be necessary e.g. for obstacle avoidance, to also consider other iterations. Therefore, $\mathcal{VS}^j(\boldsymbol{x})$ is extended to the state-varying safe set

$$\mathcal{VS}^j(\boldsymbol{x}) = \left[ \bigcup_{i \in \mathcal{T}^j(N_l)} \bigcup_{t=\gamma^i(x)}^{\gamma^i(x)+N_p} \boldsymbol{x}_t^i \right] \cap \mathcal{S}(\boldsymbol{x}) \cap \mathcal{V}(\boldsymbol{x}) \tag{4.15}$$

with the above definition for $\mathcal{T}^j(n)$,

$$\mathcal{S}(\boldsymbol{x}) = \left\{ \boldsymbol{x} \in \mathcal{X} : ||\boldsymbol{e}_1^\mathsf{T} \boldsymbol{x}^i_{\gamma^i(x)} - \boldsymbol{e}_1^\mathsf{T} \boldsymbol{x}||_2^2 \leq N \Delta t \boldsymbol{e}_5^T \boldsymbol{x} \right\} \tag{4.16}$$

and

$$\mathcal{V}(\boldsymbol{x}) = \left\{ \boldsymbol{x} \in \mathcal{X} : ||\boldsymbol{e}_5^\mathsf{T} \boldsymbol{x}^i_{\gamma^i(x)+N} - \boldsymbol{e}_5^\mathsf{T} \boldsymbol{x}||_2^2 \leq N \Delta t a_{\max} \right\}. \tag{4.17}$$

The state-varying safe set $\mathcal{VS}^j(x)$ always contains the $N_l$ fastest reachable laps. Reaching these states is enforced by the sets $\mathcal{S}(\boldsymbol{x})$ and $\mathcal{V}(\boldsymbol{x})$. They restrict the safe set to only include states, which can be reached by the end of the horizon.

### 4.1.4 Safe Set Relaxation

Since the sampled safe set consists of discrete states, the LMPC-problem becomes a Nonlinear Mixed Integer Program and cannot be guaranteed to be solved in real-time. Therefore, the sampled safe set is relaxed to its convex hull, which results in a reduction of the computation time. Furthermore, for each convex combination of the elements in $\mathcal{SS}^j$ there exists a control sequence that steers the system to the final state $\boldsymbol{x}_f$, because $\mathcal{X}$ and $\mathcal{U}$ are convex. Therefore, the convex safe set

$$\mathcal{VS}^j(x) = \mathrm{Conv}(\mathcal{SS}^j) = \left\{ \sum_{i=1}^{|\mathcal{SS}^j|} \lambda_i z_i : \lambda_i \geq 0, \sum_{i=1}^{|\mathcal{SS}^j|} \lambda_i = 1, z_i \in \mathcal{SS}^j \right\} \tag{4.18}$$

is a control invariant set, where $|\mathcal{SS}^j|$ is the cardinality of the $\mathcal{SS}^j$.

### 4.1.5 Modified Teminal Cost Approximation

Relaxing the safe set to its convex hull also requires adapting the terminal cost accordingly. For this reason we introduce the barycentric function

$$P^j(x) = \begin{cases} p^{j,*}, & \text{if } x \in \mathcal{CS}^j \\ +\infty, & \text{if } x \notin \mathcal{CS}^j \end{cases}, \tag{4.19}$$

where

$$p^{j,*}(x) = \min_{\lambda_t \geq 0, \forall t \in [0,\infty)} \sum_{k=0}^{j} \sum_{t=0}^{\infty} \lambda_t^k J_{t\to\infty}^k(x_t^k) \tag{4.20}$$

$$\text{subj. to} \sum_{k=0}^{j} \sum_{t=0}^{\infty} \lambda_t^k = 1, \tag{4.21}$$

$$\sum_{k=0}^{j} \sum_{t=0}^{\infty} \lambda_t^k x_t^k = x. \tag{4.22}$$

Note, that in real applications each iteration $j$ has a finite time duration $t_j$. Consequently, the summation up to infinity has be replaced by a summation up to $t_j$. However, the previous formulation from [RosoliaBorrelli17b] does not demonstrate desirable behavior close to the final states $\boldsymbol{x}_f \in \mathcal{X}_f$, which leads to a suboptimal racing behavior, where the vehicle decelerates close to the finish line.

The iteration cost for the racing scenario is therefore modified to

$$\hat{J}_{t\to\infty}^k(x_t^k) = J_{t\to\infty}^k(x_t^k) + J_{0\to\infty}^k(x_0^k) - \min_{i\ \in[0,j]} J_{0\to\infty}^i(x_0^i). \tag{4.23}$$

This formulation takes into account the difference in the amount of time steps needed to finish each iteration and can provide the correct cost even close to the finish line. The terminal cost for the racing application is then given by

$$\hat{P}^j(x) = \begin{cases} \hat{p}^{j,*}, & \text{if } x \in \mathcal{CS}^j \\ +\infty, & \text{if } x \notin \mathcal{CS}^j \end{cases} \tag{4.24}$$

with

$$\hat{p}^{j,*}(x) = \min_{\lambda_t \geq 0, \forall t \in [0,\infty)} \sum_{k=0}^{j} \sum_{t=0}^{t_j} \lambda_t^k \hat{J}_{t\to\infty}^k(x_t^k) \tag{4.25}$$

$$\text{subj. to} \sum_{k=0}^{j} \sum_{t=0}^{t_j} \lambda_t^k = 1 \tag{4.26}$$

$$\sum_{k=0}^{j} \sum_{t=0}^{t_j} \lambda_t^k x_t^k = x, \tag{4.27}$$

with $\hat{J}$ being the modified iteration cost.

### 4.1.6 Repetitive LMPC

In [BrunnerEtAl17] the application of LMPC to the autonomous racing problem, which was initially proposed by [RosoliaCarvalhoBorrelli17], was extended into handling repetitive tasks. In [RosoliaCarvalhoBorrelli17] the initial conditions are unchanged at each iteration (i.e. $x_S = x_0^j,\ \forall j \geq 0$). Generally, in a repetitive setting the initial conditions of the $j$-th iteration are a function of the final state of the previous iteration $j-1$. The new presented formulation in [BrunnerEtAl17] enables the design of a controller which learns from data improving its performance with respect to the infinite horizon optimal control problem, where $\boldsymbol{x}_0$ is an optimization variable. In the specific autonomous racing application, given the track length $s_f$, yields

$$\boldsymbol{x}_0^{j+1} = \boldsymbol{x}_{t_j+1}^{j} - \boldsymbol{x}_{\text{shift}},\ \forall j \geq 0 \tag{4.28}$$

with $\boldsymbol{x}_{\text{shift}} = [s_f,\ 0,\ 0,\ 0,\ 0,\ 0]^\mathsf{T}$.

### 4.1.7 Single-Agent LMPC

The following nonlinear program is the LMPC adapted to the single-agent autonomous racing problem:

$$\begin{aligned} J_{t\to t+N}^{j,*}(\boldsymbol{x}_t^j) = \min_{U_{t\to t+N|t}} \ & \hat{P}^{j-1}(\boldsymbol{x}_{t+N|t}) + \sum_{k=t}^{t+N-1} h(\boldsymbol{x}_{k|t}, \boldsymbol{u}_{k|t}) + \\ & \sum_{k=t}^{t+N-1} \left( \boldsymbol{w}_u^\mathsf{T} \begin{bmatrix} a_k^2 \\ \delta_{\text{f},k}^2 \end{bmatrix} \right) + \sum_{k=t+1}^{t+N-1} \left( \boldsymbol{w}_{\Delta u}^\mathsf{T} \begin{bmatrix} (a_k - a_{k-1})^2 \\ (\delta_{\text{f},k} - \delta_{\text{f},k-1})^2 \end{bmatrix} \right) \\ & \sum_{k=t}^{t+N} \left( (\boldsymbol{w}_x \odot \boldsymbol{x}_{k|t})^\mathsf{T} \boldsymbol{x}_{k|t} \right) + \\ & \sum_{k=t+1}^{t+N} \left( (\boldsymbol{w}_{\Delta x} \odot (\boldsymbol{x}_{k|t} - \boldsymbol{x}_{k-1|t}))^\mathsf{T} (\boldsymbol{x}_{k|t} - \boldsymbol{x}_{k-1|t}) \right) \end{aligned} \tag{4.29}$$

$$\text{subj. to } \boldsymbol{x}_{k+1|t} = f_{\text{dyn}}(\boldsymbol{x}_{k|t}, \boldsymbol{u}_{k|t}; B, C, D),\ \forall k \in \{t, ..., t+N-1\} \tag{4.30}$$

$$\boldsymbol{x}_{t|t} = \boldsymbol{x}_t^j \tag{4.31}$$

$$\boldsymbol{x}_{k|t} \in \mathcal{X},\ \forall k \in \{t, ..., t+N\} \tag{4.32}$$

$$-\frac{w_{\text{track}}}{2} \leq e_2^T \boldsymbol{x}_{k|t} \leq \frac{w_{\text{track}}}{2},\ \forall k \in \{t, ..., t+N\} \tag{4.33}$$

$$\boldsymbol{u}_{k|t} \in \mathcal{U},\ \forall k \in \{t, ..., t+N-1\} \tag{4.34}$$

$$a_{\min} \leq a_k \leq a_{\max},\ \forall k \in \{t, ..., t+N-1\} \tag{4.35}$$

$$\delta_{\text{f,min}} \leq \delta_{\text{f},k} \leq \delta_{\text{f,max}},\ \forall k \in \{t, ..., t+N-1\} \tag{4.36}$$

$$\boldsymbol{x}_{N|t} \in \mathcal{VS}^{j-1}(\boldsymbol{x}_t^j), \tag{4.37}$$

where $\odot$ is the element-wise matrix product and $\boldsymbol{w}_u$, $\boldsymbol{w}_{\Delta u}$, $\boldsymbol{w}_x$ and $\boldsymbol{w}_{\Delta x}$ are the weights on the control inputs, the derivative of the control inputs, the states and the derivative of the states, respectively. Equation 4.33 is the box constraint, which ensures that the agent stays on the race track.

## 4.2 Multi-Agent Racing

In the multi-agent racing setting, multiple agents independently race on a given race track, where the objective is to cross the finish line after a certain amount of iterations or laps before all other opponents. Instead of specifying the objective function as reaching the finish line in the first place, the controller pursues three main objectives: successfully completing each iteration in a minimum amount of time steps, avoiding crashes with adversarial agents and to maximize the signed distance towards the other opponents. The number of agents considered here is limited to two agents. Nevertheless, the proposed algorithm can be extended to a greater number of agents with little effort. The developed algorithm assumes, that each agent is always able to perfectly predict the adversarial agent's predicted trajectory from the previous time step. All of the following adaptations are contributions of this thesis, except for subsection 4.2.4, which is mostly based on the developments in [RicciutiRosoliaGonzales18].

### 4.2.1 Game Theoretic Formulation

The multi-agent race is formulated as a non-cooperative two-player game based on game theoretic concepts as described in [WilliamsEtAl17]. The game is given by the set of possible control inputs for both agents $\mathfrak{U} = (U_1, U_2)$ and the set of objective functions $\mathfrak{J} = (J_1, J_2)$. Because of the interdependencies among the competing agents, each objective function also depends on the control inputs for the opponents. A common approach to solving this type of problems is the Nash Equilibrium [Nash51]. Consider the set of strategies $(U_1^*, U_2^*)$. The set of strategies is in a Nash Equilibrium, if

$$U_i^* = \arg\min_{U_i} J_i(U_i, U_{-i}), \ \forall i \in \{1, 2\}, \tag{4.38}$$

where $U_{-i} = \{U_1, U_2\} \backslash \{U_i\}$. This states that each agent's strategy is optimal, given that the strategies of the adversarial agents are fixed. However, determining the Nash Equilibrium is challenging and can potentially violate the real-time constraints. Therefore, the best response dynamics are applied for the presented problem. The best response is the strategy, which minimizes the agent's cost given the opponents strategies are fixed. In [SpicaEtAl18] and [WilliamsEtAl17] the best response dynamics for all agents are iteratively determined and can potentially result in a Nash Equilibrium. However, considering the computational complexity for the single-agent LMPC, the multi-agent LMPC does not allow for iteratively determining the best response dynamics. Consequently, only one iteration for the best response dynamics is executed, while using a fixed guess for the opponents prediction. For two agents $b$ and $r$ this yields

$$\boldsymbol{u}^{*,b}_{t+i|t} = \arg\min_{\boldsymbol{u}^b} J(\boldsymbol{x}^b, \boldsymbol{u}^b, \boldsymbol{x}^{*,r}, \boldsymbol{u}^{*,r}), \ \forall i \in \{0, \dots N\} \tag{4.39}$$

$$\boldsymbol{u}^{*,r}_{t+i|t} = \arg\min_{\boldsymbol{u}^r} J(\boldsymbol{x}^r, \boldsymbol{u}^r, \boldsymbol{x}^{*,b}, \boldsymbol{u}^{*,b}), \ \forall i \in \{0, \dots N\}. \tag{4.40}$$

The fixed guess is the predicted prediction for the adversarial agent. In the implementation the perfect information of the prediction is achieved by each agent communicating their predictions. Since both agents simultaneously optimize their trajectories, they have to rely on the predicted trajectory from the previous time step. Under the assumption, that the prediction only changes marginally for every iteration, this will still provide a valid prediction of the opponent. Therefore, the optimization problem for both agents is adapted to:

$$\boldsymbol{u}^{b}_{t+i|t} = \arg\min_{\boldsymbol{u}^b} J(\boldsymbol{x}^b, \boldsymbol{u}^b, \hat{\boldsymbol{x}}^{r}, \hat{\boldsymbol{u}}^{r}), \ \forall i \in \{0, \dots N\} \tag{4.41}$$

$$\boldsymbol{u}^{r}_{t+i|t} = \arg\min_{\boldsymbol{u}^r} J(\boldsymbol{x}^r, \boldsymbol{u}^r, \hat{\boldsymbol{x}}^{r}, \hat{\boldsymbol{u}}^{r}), \ \forall i \in \{0, \dots N\}, \tag{4.42}$$

where $\hat{\boldsymbol{x}}$ and $\hat{\boldsymbol{u}}$ are the communicated state and inputs of each adversarial agent.

### 4.2.2 Trajectory Propagation

The predicted trajectory of the previous time step $t-1$ is known by the adversarial agent $r$ at time step $t$. Applying the past prediction to the current time step $t$ in agent $b$'s LMPC formulation, requires the trajectory's propagation in the following way:

$$\hat{\boldsymbol{x}}^{r,j}_{t+i|t} = \boldsymbol{x}^{*,r,j}_{t+1+i|t-1}, \ \forall i \in \{0, \dots N-1\}. \tag{4.43}$$

Since the final terminal state of a prediction is constrained to satisfy in the terminal state constraint, the last propagated state should also lie in the convex safe set. This is achieved by shifting $\lambda$, such that

$$\hat{\boldsymbol{x}}^{r,j}_{t+N|t} = \sum_{i=0}^{j-1} \sum_{k=0}^{\infty} \lambda^{r,i}_{k+1} \boldsymbol{x}^{r,i}_{k}. \tag{4.44}$$

The propagated trajectory $\hat{\boldsymbol{x}}^{r,j}_{t+i|t}$, $\forall i \in \{0, \dots N\}$ is then used for obstacle avoidance and the distance term in the cost function of the LMPC. Note, that while using the opponents prediction, this approach depends on the accuracy of the model used in the opponent's controller. In case there is a great model mismatch, the obstacle avoidance will not be accurate and can result in suboptimal performance.

### 4.2.3 Safe Set Initialization

The original LMPC formulation does not explore suboptimal states. Nevertheless, suboptimal states might have to be considered for overtaking maneuvers. Assuming that both agents converge to a similar closed-loop trajectory, the adversarial agents are occupying the same positional states on the race track. The original formulation might still yield a result if the horizon $N$ is chosen to be long enough, but that also increases the computation time. A simple exploration strategy can be achieved by starting multiple different LMPCs based on different path-following initializations, where the objective of each path-following MPC is to not necessarily follow the center line of the track but arbitrary constant values for $e_{y,\text{ref}}$. Here, three different initializations are used starting from the center, inside and outside of the track. The cost for the references for the lateral distance have therefore to be replaced by

$$J_{\text{center}} = \sum_{k=t}^{t+N} \left( w_v (v_k - v_{\text{ref}})^2 + w_{e,y} (e_{y,k} - e_{y,\text{ref,center}})^2 \right), \tag{4.45}$$

$$J_{\text{inner}} = \sum_{k=t}^{t+N} \left( w_v (v_k - v_{\text{ref}})^2 + w_{e,y} (e_{y,k} - e_{y,\text{ref,inner}})^2 \right), \tag{4.46}$$

$$J_{\text{outer}} = \sum_{k=t}^{t+N} \left( w_v (v_k - v_{\text{ref}})^2 + w_{e,y} (e_{y,k} - e_{y,\text{ref,outer}})^2 \right), \tag{4.47}$$

respectively.

### 4.2.4 Obstacle Avoidance

Adversarial agents are modeled as dynamic obstacles in the multi-agent LMPC formulation. This subsection introduces the approach to obstacle avoidance as presented in [RicciutiRosoliaGonzales18]. One of the challenges related to dealing with obstacles originates from the non-convexity of the problem, which is addressed in subsection 4.2.5. A possibility for incorporating obstacle avoidance is the addition of the state constraint

$$\left( \frac{\boldsymbol{e}_1^\top \boldsymbol{x}_{k+t|t} - s_k^{\text{obs}}}{r_s} \right)^2 + \left( \frac{\boldsymbol{e}_2^\top \boldsymbol{x}_{k+t|t} - e_{y,k}^{\text{obs}}}{r_{e_y}} \right)^2 \geq 1, \ \forall k \in \{0, ..., N\} \tag{4.48}$$

to the existing optimization problem, where $\boldsymbol{s}^{\text{obs}}$ and $\boldsymbol{e}_y^{\text{obs}}$ are the curvilinear abscissa and lateral distance of the obstacle over the horizon and where $r_s$ and $r_{e_y}$ are the radii of the ellipse in $s$ and $e_y$ direction, respectively. This assumes that the error in the heading angle is always small, because a potential rotation of the ellipse constraint is not considered. This inaccuracy in the modeling of the constraint can be taken into account by simply increasing the radii of the ellipse.

Because of the imposed challenges for numerically solving this type of non-convex problem, the computation tends to be not real-time feasible. Instead, the above state constraint is reformulated as part of the objective function using a logarithmic barrier function. The additional cost is then given by

$$J_{\text{obs}}(\boldsymbol{x}_t, s_k^{\text{obs}}, e_{y,k}^{\text{obs}}) = -w_{\text{obs}} \log \left( w_{\text{safe}} \left( \left( \frac{\boldsymbol{e}_1^\top \boldsymbol{x}_{k+t|t} - s_k^{\text{obs}}}{r_s} \right)^2 + \left( \frac{\boldsymbol{e}_2^\top \boldsymbol{x}_{k+t|t} - e_{y,k}^{\text{obs}}}{r_{e_y}} \right)^2 - 1 \right) \right), \forall k \in \{0, ..., N\}, \tag{4.49}$$

where $w_{\text{obs}}$ and $w_{\text{safe}}$ are the weights on the barrier function, which determine the magnitude and steepness of the logarithmic function.

### 4.2.5 Modified Safe Set for Overtaking Maneuvers

The selection of states for the terminal state constraint is also based on the strategy the agent is pursuing. In [LinigerLygeros17] only the trailing vehicle is considering the opponent as an obstacle. Here, the obstacle is always considered, but the weights in the cost function change according to the relative position to the adversary. If no adversarial agent $r$ is in front of agent $b$, only the states from the set $\mathcal{VS}^j(x)$ get added. Additionally, the same terminal states are selected, if agent $b$ is behind an adversarial agent $r$, but $v_t^b < v_t^r$. In any other case, the agent should try an overtaking maneuver and might therefore need to select different states. The modified sampled safe set $\mathcal{VS}^j(x^b, \hat{x}^r)$ then also depends on the prediction of the adversarial agent. Since the fastest iterations potentially do not allow an overtaking maneuver, different states are selected for overtaking. This requires additional consideration of the $e_y$-state of the previous states.

In past implementations, to assure obstacle avoidance and therefore safety, previous trajectories that collide with the last prediction step of the adversary are removed from the safe set [RicciutiRosoliaGonzales18]. Using the ellipse function, it is determined for each state $\boldsymbol{x}_t^i \in \mathcal{VS}^{b,j}(x)$, where $i$ is the iteration in which the state was recorded, if it collides with the adversarial agent:

$$\left( \frac{\boldsymbol{e}_1^\top \boldsymbol{x}_t^i - s_N^{\text{obs}}}{r_s} \right)^2 + \left( \frac{\boldsymbol{e}_2^\top \boldsymbol{x}_t^i - e_{y,N}^{\text{obs}}}{r_{e_y}} \right)^2 \leq 1, \tag{4.50}$$

In case of a collision, the states:

$$\left[ x_{t-\tau}^i, \ldots, x_{\gamma^i(x)+N_p}^i \right] \tag{4.51}$$

are removed from $\mathcal{VS}^{b,j}(x)$, where $\tau \in \{0, \ldots, t - \gamma^i(x)\}$ is a tuning parameter.

For the multi-agent racing, a different approach has been chosen for modifying the selected states based on the obstacle's position in the final prediction step. First, the side on which the agent $b$ is considering for overtaking has to be determined to keep the problem convex. The following defines multiple inequalities, which identify the left or the right side for the overtaking maneuver. These inequalities take into account the amount of space between the adversarial agent and the track boundary. In case sufficient amount of space is unoccupied, the overtaking maneuver can be considered. The agent should also prefer executing an overtaking maneuver on the right side, if the agent is already on the right of the opponent. Agent $b$ should then execute and overtaking maneuver on the left side, if

$$e^b_{y,t|t} - \hat{e}^r_{y,t+N|t} \leq 0 \ \vee \ \hat{e}^r_{y,t+N|t} - \frac{w}{2} \leq \alpha \cdot w^b, \alpha \in \left[1, \frac{w}{w^b}\right] \tag{4.52}$$

and an overtaking maneuver on the right for every other case, with the width $w^b$ of agent $b$. Consider the sets $S = [0, s_{\max}]$, $E_y(\hat{x}^r) = [e_{y,\min}(\hat{x}^r), e_{y,\max}(\hat{x}^r)]$, $E_\psi = [0, 2\pi]$, $V = [0, v_{\max}]$, then the set $\mathcal{E}_y(\hat{x}^r) = S \times E_y(\hat{x}^r) \times E_\psi \times V$ describes all possible states with the above restriction on $e_y$. The selected terminal states for an overtaking maneuver are therefore:

$$\mathcal{VS}^j(x^b, \hat{x}^r) = \mathcal{E}_y(\hat{x}^r) \cap \mathcal{VS}^j(x^b) \tag{4.53}$$

The interval $E_y$ is calculated in which the last prediction $e^b_{y,t+N|t}$ should lie. For overtaking on the left, which is displayed in Figure 4.1, if $e^b_{y,t|t} - \hat{e}^r_{y,t+N|t} \leq 0$ the interval is:

$$e^{b,k}_{y,t+N|t} \in [e_{y,\min}, e_{y,\max}] \tag{4.54}$$

$$e_{y,\max} = \min\left(\hat{e}^r_{y,t+N|t},\ e^b_{y,t|t} + \alpha \cdot \frac{w^b}{2}\right) \tag{4.55}$$

$$e_{y,\min} = e_{y,\max} - \alpha \cdot \frac{w^b}{2} \tag{4.56}$$

Else the agent $b$ is still right ($e^b_y \geq e^r_y$) of the adversarial agent $r$ and the following interval is chosen:

$$e^{b,k}_{y,t+N|t} \in \left[\max\left(\hat{e}^r_{y,t+N|t} - \alpha \cdot w^b,\ -\frac{w}{2}\right),\ \hat{e}^r_{y,t+N|t}\right] \tag{4.57}$$

### 4.2.6 Competitive Racing

Encouraging competitive behavior is achieved by adding a cost for the distance between the agents. In [WilliamsEtAl17] there is a constant positive cost $J^+_{\text{dist}}$ added and a negative constant cost $J^{-}_{\text{dist}} = -J^{+}_{\text{dist}}$, if the agent is trailing or leading, respectively. A different approach is presented in [LinigerLygeros17], where only a cost is added for staying ahead of the other agent at the final prediction step of the horizon. For the multi-agent LMPC this cost is chosen to be linearly depending on the signed distance between the two agents. A great distance between the agents, while the agent is also leading, then results in a small cost for the distance. The distance cost, defined over all steps of the prediction horizon, is given by

$$J_{\text{d}}(\boldsymbol{x}_t, \boldsymbol{s}^{\text{obs}}) = -w_{\text{d}} \sum_{k=t}^{t+N} \boldsymbol{e}_1^{\top} \boldsymbol{x}_{k|t} - s^{\text{obs}}_{k|t}, \; w_{\text{d}} > 0. \tag{4.58}$$

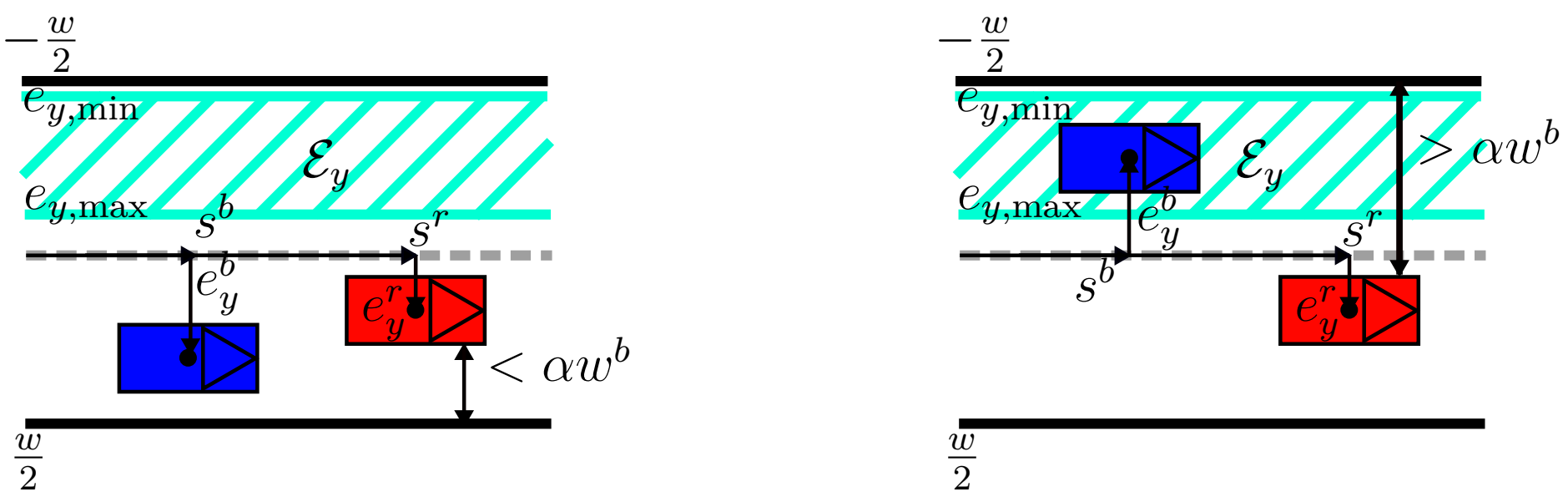

Figure 4.1: The two different scenarios for overtaking on the left.

### 4.2.7 Racing Algorithm

**Algorithmus 1** SafeSetInitialisation

**for** $agent_i$ in $[agent_1, agent_2]$ **do**
  **for** $init$ in $[inner, center, outer]$ **do**
    $t \leftarrow 0$
    **while** $\vec{x}_t \notin \mathcal{X}_F$ **do**
      $\mathcal{VS}_i^{0,t}(\vec{x}) \leftarrow PathFollowingMPC(\vec{x}_t, init)$
      $t \leftarrow t + 1$
    **end while**
    **for** $j \leftarrow 1, numLapsInit$ **do**
      $t \leftarrow 0$
      **while** $\vec{x}_t \notin \mathcal{X}_F$ **do**

$\mathcal{VS}_i^{j,t}(\vec{x}) \leftarrow LMPC(\vec{x}_t, \mathcal{VS}_i^{j-1})$
$t \leftarrow t+1$
**end while**
**end for**
**end for**
**end for**

**Algorithmus 2** Multi-Agent Racing

**for** $agent_i$ in $[agent_1, agent_2]$ **do in parallel**
$k \leftarrow 1 + 3 \cdot numLapsInit$
$t \leftarrow 0$
**while** $\vec{x}_t \notin \mathcal{X}_F$ **do**
$\mathcal{VS}_i^{k,t}(\vec{x}) \leftarrow PathFollowingMPC(\vec{x}_t, init)$
$t \leftarrow t+1$
**end while**
**for** $l \leftarrow k+1, k+numLapsRace$ **do**
$t \leftarrow 0$
**while** $\vec{x}_t \notin \mathcal{X}_F$ **do**
$\mathcal{VS}_i^{l,t}(\vec{x}) \leftarrow ObstacleLMPC(\vec{x}_t, \mathcal{VS}_i^{l-1})$
$t \leftarrow t+1$
**end while**
**end for**
**end for**

# Chapter 5

# Validation

This chapter describes the steps taken to validate the presented control method in a multi-agent racing scenario. In the beginning of the chapter the required software tools and necessary adaptations to the controller are introduced. Subsequently, the adapted controller is implemented and experimentally validated in both, a simulator and on the BARC platform.

## 5.1 Implementation

The goal of this section is to outline the implementation of the proposed multi-agent LMPC on a physical system. The implementation includes, the introduction of suitable software packages, the definition of the race tracks, a method to accurately determine the dynamics of the vehicles and necessary adaptations and simplifications of the algorithm.

### 5.1.1 Software

Different programming languages and software libraries are used for the implementation. The following introduces the Julia language and Julia for Mathematical Optimization (JuMP), which are used for solving nonlinear optimization problems efficiently, the nonlinear solver interior point optimizer (IPOPT) and the Robot Operating System (ROS), which enables communication and scheduling among multiple computers and programs.

### Julia and JuMP

Julia is a free and open-source programming language used for numerical computing and scientific research and is introduced in [BezansonEtAl17]. The Julia language itself is a high-level, high-performance dynamic programming language. The design of the Julia language and its sophisticated just-in-time compiler enable Julia to nearly reach the performance of C, while having a syntax comparable to Python. Furthermore, the Julia PyCall package provides any Python function to the user, which guarantees great accessibility.

JuMP is a modeling language for mathematical optimization in Julia [DunningHuchetteLubin17]. It supports various solvers, e.g. IPOPT, for different problem classes, including linear-programming, mixed-integer programming and nonlinear programming. Using JuMP is advantageous, since its syntax allows for easy translation of mathematical expressions to programming code. Consider the following QP:

$$J^*(\boldsymbol{x}) = \min_{\boldsymbol{x}} 3x_1^2 + x_2^2 + 2x_1x_2 + x_1 + 6x_2 + 2 \tag{5.1}$$

$$\text{subj. to } 2x_1 + 3x_2 \geq 4 \tag{5.2}$$

$$x_i \geq 0, \forall i \in \{1, 2\} \tag{5.3}$$

The following lines of code implement the above optimization problem in Julia with JuMP:

```
using JuMP  # Include Julia for Mathematical Optimization
using IPOPT  # Include the IPOPT solver

# Create the optimization model
m = Model(solver = IPOPTSolver())
# Create variable x, constrained to be always positive
@variable(m, x[1 : 2] >= 0)

# Define the cost function
@objective(m, Min, 3*x[1]^2 + x[2]^2 + 2*x[1]*x[2] + x[1]
                  + 6*x[2] + 2)
@constraint(m, 2*x[1] + 3*x[2] >= 4)  # Add constraint

status = solve(m)  # Solve the optimization problem

# Print the cost and optimal values
println("Objective value: ", getobjectivevalue(m))
println("x_1 = ", getvalue(x[1]))
println("x_2 = ", getvalue(x[2]))
```

All the above advantages make Julia suitable for the autonomous racing application and, therefore, the LMPC is implemented in the Julia language.

### IPOPT

The interior point optimizer (IPOPT) software library is used to solve the nonlinear optimization problem as stated in the LMPC for autonomous racing. IPOPT is able to find, depending on the type of problem, globally or locally optimal solutions to mathematical optimization problems in the standard form, introduced in section 3.1.2. IPOPT applies the interior point method to solve an optimization problem. Interior point methods iteratively use a natural logarithmic barrier function to reformulate the inequality constraints of the optimization problem as part of the objective function. The details on IPOPT and how to solve the sequence of formulated barrier problems can be found in [WächterBiegler06].

### Robot Operating System

The Robot Operating System (ROS) has initially been developed by [QuigleyEtAl09] and has been extended through the open-source community. ROS is middleware for operating robotic and control systems. It provides the scheduling of different threads, as well as the package management. Moreover, it can handle the message-passing among processes even distributed on different hardware systems and written in different programming languages. This allows for a modular software design and encapsulation. ROS is, among other programming languages, available in C++ and Python, but can also be used in the Julia language through the RobotOS package.

An instance of a program in ROS is called a node. Each node has the ability to communicate with other running nodes over a defined interface. These interfaces are called topics in ROS. Each topic has a defined message type, which specifies the format of the data communicated among the nodes. Each node is then able to publish messages to different topics, as long as the format is correct. In a similar way nodes subscribe to topics and receive the sent messages from other nodes. Since the topics define the message types, it is possible to have nodes written in different programming languages communicate. Additionally, ROS allows for execution on multiple devices in parallel, where one device is responsible for setting up the network and communication, which is called the ROS master. The other devices are called ROS slaves.

### 5.1.2 Race Track

The race track defines the constraints on the positional states of the vehicle. First, the type of considered race tracks are presented. Furthermore, the two different tracks for the controller validation are given. Then, the transformation from the Cartesian coordinate system to the Frenet reference system is presented. The transformation is necessary, because the estimated vehicle state is expressed in Cartesian coordinates, while the controller expects the position in the Frenet system.

#### Race Track Defintion

The race tracks are defined using a piecewise constant curvature function $\kappa(s)$. Therefore, the track is constructed by multiple connected circular arcs with different lengths and radii. This has the advantage, that the transformation from cartesian coordiantes to the Frenet reference system can be performed using an analytic solution. Previous implementations, e.g. [BrunnerEtAl17], used a linear regression to fit a polynomial at every time step, which can be inaccurate or more computationally expensive. Furthermore, this will have an advantage for the implementation in the real system, which is discussed in section 5.1.4. To define the piecewise constant curvature function let there be $n+1$ points, which define the positions where the curvature changes:

$$s_0, s_1, s_2, ..., s_n \tag{5.4}$$

with

$$0 = s_0 < s_1 < s_2 < ... < s_n = s_f. \tag{5.5}$$

Furthermore, there are $n$ curvature values $\kappa_j, j \in \{1, 2, ..., n\}$, for each interval. Then the piecewise constant curvature function is

$$\kappa(s) = \kappa_j, \forall s \in [s_j, s_{j+1}), j \in \{0, 1, ..., n-1\}. \tag{5.6}$$

#### Tracks

There are two tracks available for the experiments. The first track is a simple oval track with the length $s_f = 16.00\,\mathrm{m}$ and the width $w_{\text{Track}} = 1.20\,\mathrm{m}$. The track in the Cartesian coordinate system and its curvature profile are displayed in Figure 5.1 and Figure 5.2. The track only contains left turns, because the vehicle is defined to run around the track counter clock wise.

The second track is a more complicated track with an L-shape. The track has the length $s_f = 19.60\,\mathrm{m}$ and the width $w_{\text{Track}} = 1.00\,\mathrm{m}$. The track in the cartesian coordinate system and its curvature profile are displayed in Figure 5.3 and

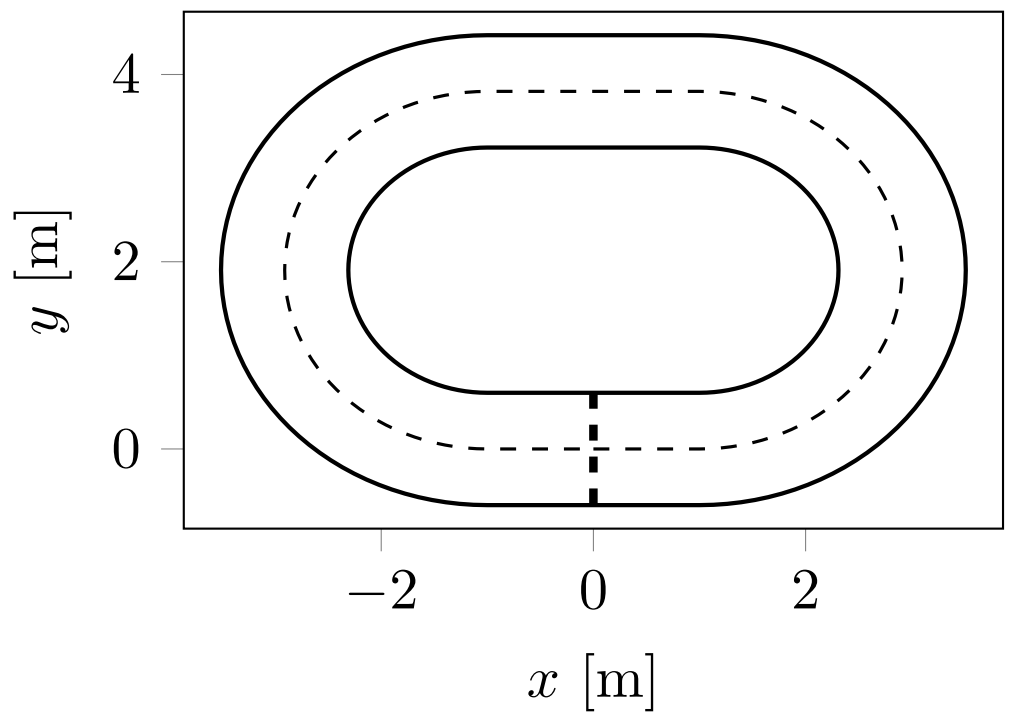

Figure 5.1: Oval race track.

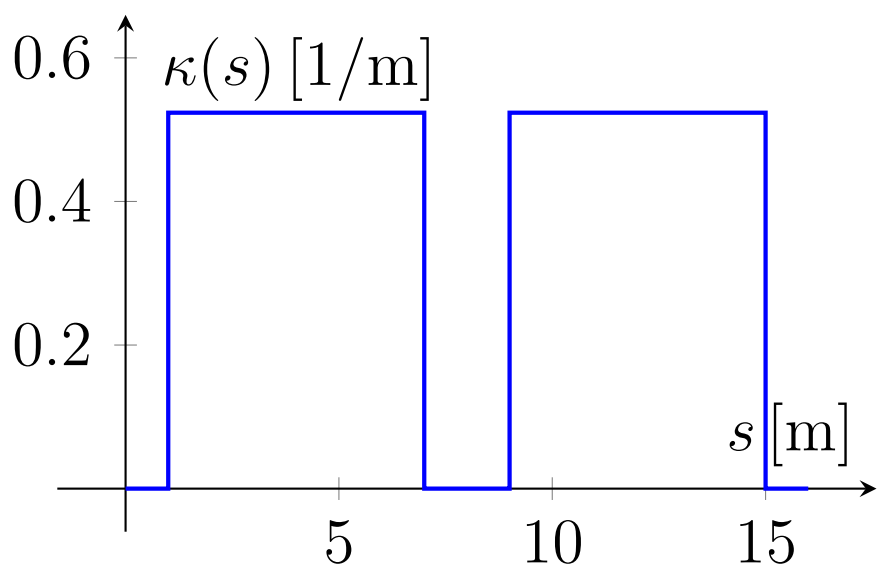

Figure 5.2: Curvature of the oval race track.

Figure 5.4. The track exhibits left and right turns and longer straight stretches. Therefore, this track requires more steering action but should also allow for higher velocities, which makes the handling of the vehicle more demanding.

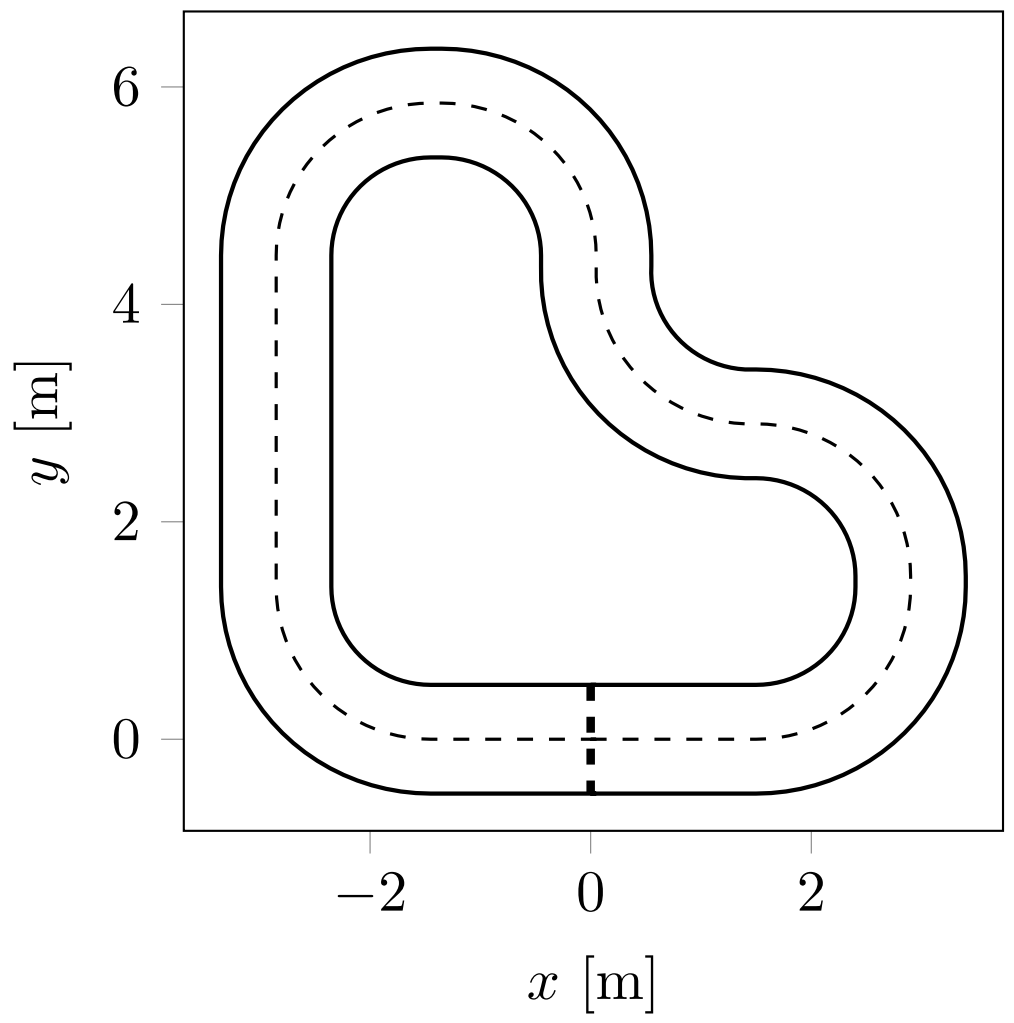

Figure 5.3: L-shape race track.

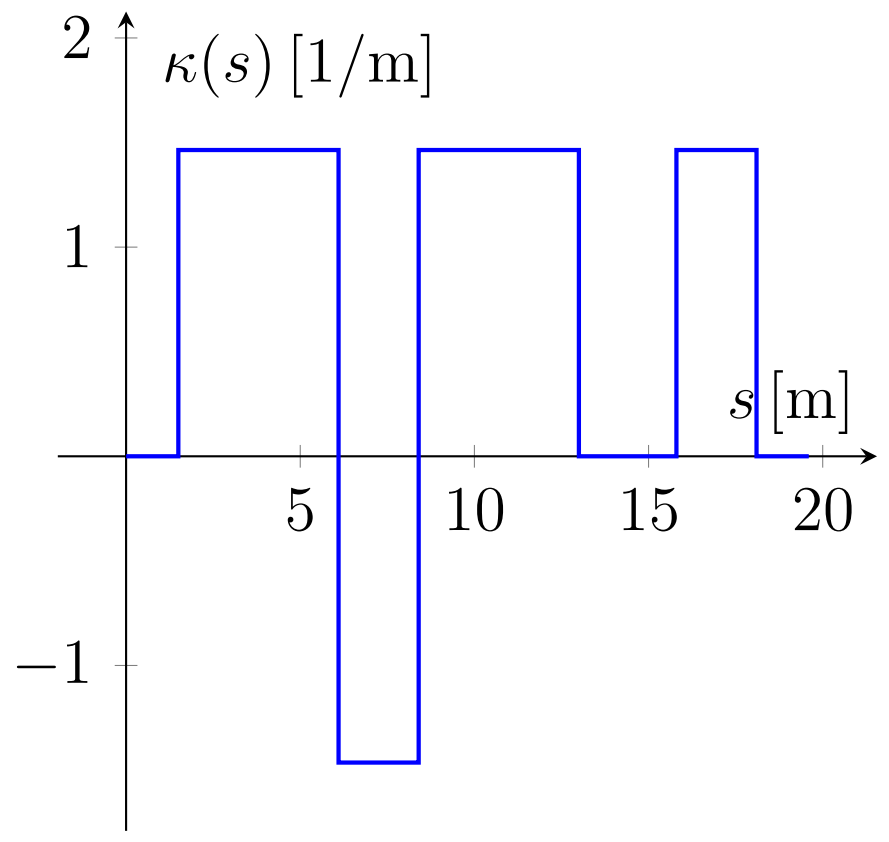

Figure 5.4: Curvature of the L-shape race track.

### Transformations

The advantage of the presented definition of the track is, that analytic tools are able to transform the position coordinates from the cartesian coordinate system to the Frenet reference frame. In the implementation, the track is discretized with $\Delta s$, which is the distance between two consecutive points on the track. During the creation of the track, a mapping is constructed, which maps every discrete point of the track center line $s_k$ in the Frenet reference frame to a pair of values $(x_k, y_k)$, which describe the exact same position in the carteseian coordinate system. For transforming a pair of values $(x, y)$ to the equivalent pair $(s, e_y)$, the two closest discrete points on the track center line need to be determined. Using their according $s_k$-values in the Frenet coordinate system, the interval $s \in [s_j, s_{j+1}]$ can be calculated. Because of the way the track is defined, the curvature in the interval $[s_j, s_{j+1}]$ is constant and $r(s_j) = r(s) = r(s_{j+1})$. The origin $T$ of the arc can be calculated using the two points $\boldsymbol{x}_j$ and $\boldsymbol{x}_{j+1}$ and the curvature $\kappa(s_j)$. Subsequently, the application of the law of cosines yields the angular offset $\Delta\theta$ of the state $(s, e_y)$ in relation to $(s_j, e_{y,j})$. The offset is then given by:

$$\Delta\theta = \arccos\left(\frac{|r(s_j)|^2 + ||T - \boldsymbol{x}||_2^2 - ||\boldsymbol{x}_j - \boldsymbol{x}||_2^2}{2 \cdot |r(s_j)| \cdot ||T - \boldsymbol{x}||_2}\right), \tag{5.7}$$

where $||\boldsymbol{x}_j - \boldsymbol{x}||_2$ is the distance from $(x_j, y_j)$ to $(x, y)$ and

$$||T - \boldsymbol{x}||_2 = r(s) + e_y. \tag{5.8}$$

The relationship is shown in Figure 5.5. The current curvilinear abscissa can then be expressed as

$$s = s_j + \Delta\theta \cdot r(s_j) \tag{5.9}$$

and Equation 5.8 yields $e_y$.

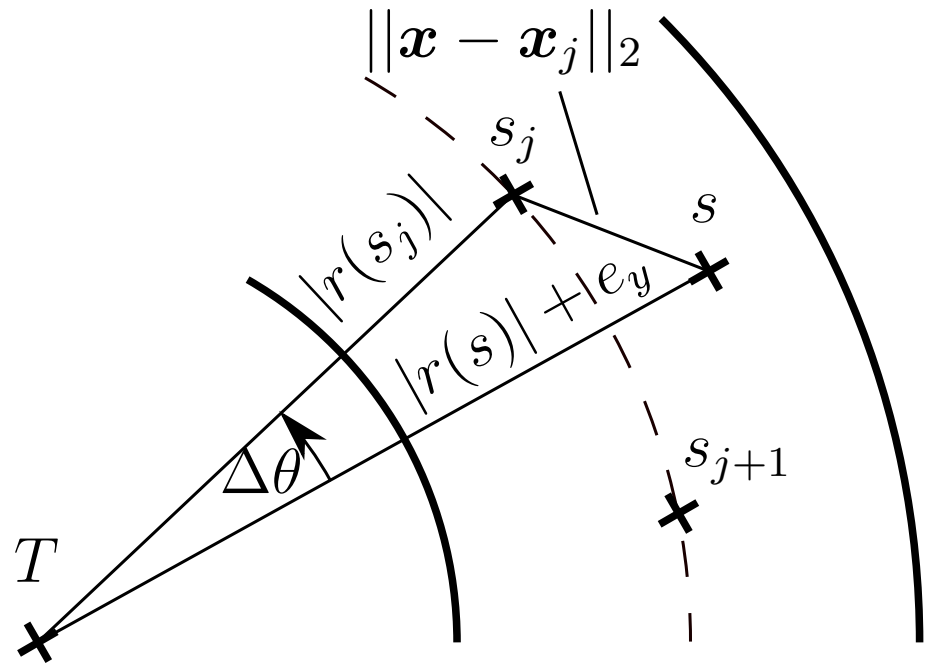

Figure 5.5: Applied law of cosines for the transformation.

### 5.1.3 System Identification

Accurately modeling the vehicle dynamics is crucial in order to safely control the vehicle, especially at higher velocities. Identifying or measuring the Pacejka coefficients sufficiently is difficult and has to be adapted for different surfaces and wheels. As proposed in [RosoliaCarvalhoBorrelli17], a linear regression model is used to identify the dynamics of the vehicle accurately at each time step online. The linear regression model uses nonlinear features based on the derived dynamic bicycle and the interdependencies of each state.

From Equation 2.35 to Equation 2.37 it follows that

$$\begin{aligned} \dot{v}_x &= f\left(a, \dot{\psi} v_y, v_x\right) &(5.10)\\ \dot{v}_y &= f\left(\frac{v_y}{v_x}, \frac{\dot{\psi}}{v_x}, \delta_{\mathrm{f}}, \dot{\psi} v_x\right) &(5.11)\\ \ddot{\psi} &= f\left(\frac{v_y}{v_x}, \frac{\dot{\psi}}{v_x}, \delta_{\mathrm{f}}\right), &(5.12) \end{aligned}$$

where the dependency on $v_x$ for $\dot{v}_x$ has been added to Equation 5.10 to model a potential motor drag. Let $\boldsymbol{\theta}_{v_x}$, $\boldsymbol{\theta}_{v_y}$ and $\boldsymbol{\theta}_{\dot{\psi}}$ be the parameters determined by the linear regression. Then the dynamics of the system can be written as a set of linear equations in discretized form

$$\begin{aligned} v_{x,k+1} - v_{x,k} &= \theta_{v_x,1} a_k + \theta_{v_x,2} r_k v_{y,k} + \theta_{v_x,3} v_{x,k}, &(5.13)\\ v_{y,k+1} - v_{y,k} &= \theta_{v_y,1} \frac{v_{y,k}}{v_{x,k}} + \theta_{v_y,2} \frac{r_k}{v_{x,k}} + \theta_{v_y,3} \delta_{\mathrm{f},k} + \theta_{v_y,4} r_k v_{x,k}, &(5.14)\\ r_{k+1} - r_k &= \theta_{\dot{\psi},1} \frac{v_{y,k}}{v_{x,k}} + \theta_{\dot{\psi},2} \frac{r_k}{v_{x,k}} + \theta_{\dot{\psi},3} \delta_{\mathrm{f},k}. &(5.15) \end{aligned}$$

Using multiple samples $n$ form previously recorded states, each equation can be rewritten in matrix-vector notation

$$\boldsymbol{y} = \boldsymbol{X}\boldsymbol{\theta}, \tag{5.16}$$

where for example the identification of $v_x$ yields

$$\boldsymbol{y}_{v_x} = \begin{bmatrix} v_{x,2} - v_{x,1} \\ v_{x,3} - v_{x,2} \\ \vdots \\ v_{x,n} - v_{x,n-1} \end{bmatrix}, \ \boldsymbol{X}_{v_x} = \begin{bmatrix} a_1 & r_1 v_{y,1} & v_{x,1} \\ a_2 & r_2 v_{y,2} & v_{x,2} \\ \vdots & \vdots & \vdots \\ a_{n-1} & r_{n-1} v_{y,n-1} & v_{x,n-1} \end{bmatrix}, \ \boldsymbol{\theta}_{v_x} = \begin{bmatrix} \theta_{v_x,1} \\ \theta_{v_x,2} \\ \theta_{v_x,3} \end{bmatrix}. \tag{5.17}$$

The matrices and vectors for $v_y$ and $r$ can be written accordingly.

In order to determine the optimal coefficents $\boldsymbol{\theta}^*$, the following optimization problem has to be solved

$$\boldsymbol{\theta}^* = \arg\min_{\boldsymbol{\theta}} ||\boldsymbol{X}\boldsymbol{\theta} - \boldsymbol{y}||_2. \tag{5.18}$$

Various methods for solving this type of problem exist. Here, the normal equation method is chosen, which solves the optimization problem analytically with

$$\boldsymbol{\theta}^* = \left(\boldsymbol{X}^\mathsf{T}\boldsymbol{X}\right)^{-1}\boldsymbol{X}^\mathsf{T}\boldsymbol{y}. \tag{5.19}$$

The most expensive computation in Equation 5.19 is the calculation of the inverse. Since either $\boldsymbol{X} \in \mathbb{R}^{n\times 3}$ or $\boldsymbol{X} \in \mathbb{R}^{n\times 4}$, then $\boldsymbol{X}^\mathsf{T}\boldsymbol{X} \in \mathbb{R}^{3\times 3}$ or $\boldsymbol{X}^\mathsf{T}\boldsymbol{X} \in \mathbb{R}^{4\times 4}$. In both cases the inversion of the matrices is fast and only adds a neglectable amount of additional computation. Also note, that the inversion is independent of the number of samples $n$ and therefore even big number of samples could be used without increasing the computational cost of the inversion step.

Now, the discretized dynamic bicycle model can be rewritten using the identified dynamics as

$$s_{k+1} = s_k + T \cdot \left(\frac{v_{x,k}\cos(e_{\psi,k}) - v_{y,k}\sin(e_{\psi,k})}{1 - e_{y,k}\kappa(s)}\right), \tag{5.20}$$

$$e_{y,k+1} = e_{y,k} + T \cdot \left(v_{x,k}\sin(e_{\psi,k}) + v_{y,k}\cos(e_{\psi,k})\right), \tag{5.21}$$

$$e_{\psi,k+1} = e_{\psi,k} + T \cdot \left(r_k - \kappa(s)\frac{v_{x,k}\cos(e_{\psi,k}) - v_{y,k}\sin(e_{\psi,k})}{1 - e_{y,k}\kappa(s)}\right), \tag{5.22}$$

$$r_{k+1} = r_k + T \cdot \left(\theta_{\dot{\psi},1}\frac{v_{y,k}}{v_{x,k}} + \theta_{\dot{\psi},2}\frac{r_k}{v_{x,k}} + \theta_{\dot{\psi},3}\delta_{\mathrm{f},k}\right), \tag{5.23}$$

$$v_{x,k+1} = v_{x,k} + T \cdot \left(\theta_{v_x,1}a_k + \theta_{v_x,2}r_k v_{y,k} + \theta_{v_x,3}v_{x,k}\right), \tag{5.24}$$

$$v_{y,k+1} = v_{y,k} + T \cdot \left(\theta_{v_y,1}\frac{v_{y,k}}{v_{x,k}} + \theta_{v_y,2}\frac{r_k}{v_{x,k}} + \theta_{v_y,3}\delta_{\mathrm{f},k} + \theta_{v_y,4}r_k v_{x,k}\right). \tag{5.25}$$

For the remainder of the thesis, Equation 5.20 through Equation 5.25, which is the dynamic bicycle model, parameterized by the linear regression model with $\boldsymbol{\theta}_{v_x}$, $\boldsymbol{\theta}_{v_y}$ and $\boldsymbol{\theta}_{\dot{\psi}}$, are referred to as

$$\boldsymbol{x}_{k+1} = f_{\mathrm{dyn}}(\boldsymbol{x}_k, \boldsymbol{u}_k; \boldsymbol{\theta}_{v_x,k}, \boldsymbol{\theta}_{v_y,k}, \boldsymbol{\theta}_{\dot{\psi},k}). \tag{5.26}$$

Furthermore, the selection of states, which are used for the system identification, are of great significance. The best approximation of the vehicle's current dynamics $\boldsymbol{x}^j_{\mathrm{dyn},t} = \left(\dot{v}_{x,t}, \dot{v}_{y,t}, \ddot{\psi}, t\right)^T$ are obtained by only selecting previous dynamic states, which are close to the current dynamics. However, the definition of close is non-trivial in this context. A plausible approach is to select the dynamic states, which minimize the 2-norm of the difference of the previous dynamic state and the current dynamic state. Unfortunately, the dynamic states show different ranges, such that a normalization has to be performed. Since the normalization is ambiguous, the selection of states in this implementation is based on previous states in the safe set and the recently visited ones in the current iteration. The set of dynamic states used for the linear regression is then given by

$$\mathcal{SI}^j(\boldsymbol{x}_{\text{dyn}}, \boldsymbol{u}) = \left[ \bigcup_{i \in \mathcal{T}^j(N_{ID})} \bigcup_{t=\max(\gamma^i(x)-N_{\text{before}},0)}^{\min(\gamma^i(x)+N_{\text{after}},\bar{t}^i)} \begin{bmatrix} \boldsymbol{x}^i_{\text{dyn},t} \\ \boldsymbol{u}^i_t \end{bmatrix} \right] \cup \left[ \bigcup_{t=\max(\gamma^j(x)-1-N_{\text{before}},0)}^{\gamma^j(x)-1} \begin{bmatrix} \boldsymbol{x}^j_{\text{dyn},t} \\ \boldsymbol{u}^j_t \end{bmatrix} \right] \tag{5.27}$$

with $N_{ID} \leq N_l$, $N_{\text{before}} \in \mathbb{N}$ and $N_{\text{after}} \in \mathbb{N}$. Note, that the values for $N_{\text{before}}$ and $N_{\text{after}}$ should be selected depending on the number of time steps in each iteration in order to achieve a valid approximation of the vehicle dynamics.

### 5.1.4 Adaptations

Some adaptations and minor changes to the original problem have to be made, for implementing the presented learning model predictive controller, such that real-time constraints are satisfied.

#### Curvature Propagation

In the previously stated optimization problem for the LMPC for autonomous racing, it is assumed that it is possible to optimize over the curvature $\kappa(s)$. This is assumption is necessary, because $s$ is an optimization variable. However, IPOPT does not provide functionality to implement the constant piecewise function $\kappa(s)$, which defines the race track. This is the reason why previous implementations approximated the curvature as a polynomial. However, depending on the true $\kappa(s)$ this can be inaccurate. To overcome this issue, the curvature of the used race tracks has been defined as constant piecewise functions. This has the advantage that the previous prediction can be used to approximate the curvature of the current prediction, assuming that the previous and current prediction will be similar. As long as

$$\kappa(\boldsymbol{e}_1^\mathsf{T} \boldsymbol{x}_{k|t}) = \kappa(\boldsymbol{e}_1^\mathsf{T} \boldsymbol{x}_{k|t+1}), k \in \{0, ..., N-1\} \tag{5.28}$$

holds, that is, the values for $s$ stay in the same interval, the curvature approximation is accurate. However, this is not the case for values of $s$ which are close to a change in the curvature. In order to reduce this error, the values from the previous prediction are propagated by one time step. That way the difference in the approximated $s$ values and the actual predicted $s$ values is smaller. Therefore, at time step $t+1$ of an iteration the approximated curvature is given by

$$\hat{\kappa}_{t+1,k} = \kappa(\boldsymbol{e}_1^\mathsf{T} \boldsymbol{x}_{k+1|t}), k \in \{0, ..., N-1\}. \tag{5.29}$$

### Slack Variables

State constraints follow from restrictions on the real system. However, it cannot be excluded, that the state of the real system never leaves the feasible region. This can lead to potential complications in the controller implementation. Therefore, the state constraints are typically "softened" by reformulating them as soft constraints, such that

$$x \leq x_{\max} \tag{5.30}$$

is approximated as

$$x \leq x_{\max} + \epsilon, \epsilon \geq 0, \tag{5.31}$$

with the slack variable $\epsilon$. Furthermore, in order to not change the behavior of the system, when the original state constraint is not violated, a penalty term $l(\epsilon)$ is added to the objective function. The slack variable then becomes an optimization variable in the reformulated problem:

$$\min_{x} f(x) \quad \Leftrightarrow \quad \min_{x,\epsilon} f(x) + l(\epsilon) \tag{5.32}$$

$$\text{subj. to } g(x) \leq 0 \qquad \text{subj. to } g(x) \leq \epsilon \tag{5.33}$$

$$\epsilon \geq 0. \tag{5.34}$$

Moreover the choice of $l(\epsilon)$ has to satisfy the property $l(0) = 0$ and is chosen in a way such that it provides the solution to the original problem, if it is feasible. [BorrelliBemporadMorari17] suggest two different realizations

$$l(\epsilon) = u\epsilon, u > u^* \tag{5.35}$$

and

$$l(\epsilon) = u\epsilon + v\epsilon^2, u > u^*, v > 0, \tag{5.36}$$

where $u^*$ is an optimal dual variable of the problem

$$\min_{x} \; f(x) + u^* g(x). \tag{5.37}$$

The second choice for $l(\epsilon)$ is preferred, because of the smoothness of the function. The state constraints, namely the terminal state constraint and the track boundary constraint on $e_y$, have been implemented using soft constraints. Therefore, a deviation of the vehicle from the track and a deviation of the last prediction step from the terminal set is possible. The amount of deviation is defined by the choice of the weights $u$ and $v$.

### Trajectory Propagation

The use of a softened terminal state constraint allows for a feasible real-time implementation. As a result, the original terminal constraint can be violated and the final state of the prediction can deviate by a margin greater than $m \in \mathbb{R}$. Since the trajectory is propagated for the adversarial agent to yield the prediction of the opponent's behavior with perfect information, the propagated trajectory needs to be as close to the adversarial's predicted trajectory in the next time step as possible. Therefore, in case the deviation is greater than $d$, the final sate of the adversarial's trajectory is then given by the final prediction of the previous time step. The final prediction step for the adversary is then implemented as

$$\tilde{\boldsymbol{x}}^{r,j}_{t+N|t} = \begin{cases} \hat{\boldsymbol{x}}^{r,j}_{t+N|t}, & \text{if } ||\hat{\boldsymbol{x}}^{r,j}_{t+N|t} - \boldsymbol{x}^{*,r,j}_{t-1+N|t-1}||^2_2 \leq m, m \geq 0 \\ \boldsymbol{x}^{*,r,j}_{t-1+N|t-1}, & \text{else.} \end{cases} \tag{5.38}$$

### Continuous Lap Transition

The presented controller repeatedly tries to solve the same problem for each iteration with the distinction, that each iteration starts from a different initial state [BrunnerEtAl17]. In a racing scenario the agent is expected to smoothly switch iterations (or laps) once $\boldsymbol{x} \in \mathcal{X}_f$. The extension of the safe set with states before the start and the finish line of the track yields the continuous lap transition and also allows for a simple implementation of the state selection for the system identification. The extension is achieved by appending states from the previous and following iteration. The collection of states for iteration $j$ are then extended to

$$\hat{\boldsymbol{X}}^j = \left[\boldsymbol{x}^j_{-N_{\text{before}}}, ..., \boldsymbol{x}^j_{-1}, \boldsymbol{x}^j_0, ..., \boldsymbol{x}^j_m, \boldsymbol{x}^j_{m+1}, ..., \boldsymbol{x}^j_{m+N_{\text{trans}}}\right], \tag{5.39}$$

where

$$\left[\boldsymbol{x}^j_{-N_{\text{before}}}, ..., \boldsymbol{x}^j_{-1}\right] \tag{5.40}$$

are the last $N_{\text{before}}$ states from $\boldsymbol{X}^{j-1}$ and

$$\left[\boldsymbol{x}^j_{m+1}, ..., \boldsymbol{x}^j_{m+N_{\text{trans}}}\right] \tag{5.41}$$

are the first $N_{\text{trans}}$ states from $\boldsymbol{X}^{j+1}$, with $N_{\text{trans}} \geq \max(N, N_{\text{after}})$.

## 5.2 Simulation

The validation through simulation allows to test the algorithm under optimal conditions, e.g. no noise, no latency and the advantage, that the ground truth is available. The simulators and controllers for each vehicle are implemented on a laptop from MSI with 16 GB of memory and 8 CPUs, which have a frequency of 2.8 GHz. The MSI laptop is running Ubuntu 16.04. The communication among the different programs or nodes is managed by ROS. The simulators implement dynamic bicycle models using the parameters from Table 5.1, which are comparable to the parameters for the BARC. Note, that each agent uses the same parameters in the controller and simulator, except that their masses are specified differently. That way agent 2 should be able to achieve a faster lap time than agent 1. The parameters $N_{\text{PF}}$, $N_{\text{LMPC}}$ and $N_{\text{Race}}$ are the number of iterations or rather laps executed for the path-following MPC, the single-agent LMPC and the multi-agent LMPC, respectively. Furthermore, there are two values for $w_{\text{obs}}$ and $w_{\text{safe}}$, which are applied depending on the current constellation on the race track. If the distance to the other agent ahead is greater than $\frac{s_{\text{f}}}{2}$, then the first value of each of $w_{\text{obs}}$ and $w_{\text{safe}}$ is chosen, which motivates are more risky behavior. In the other case the behavior is more risk averse by choosing the second value. Furthermore, $\mathbf{0}_p$ and $\mathbf{1}_p$ are column vectors of zeros and ones, respectively, with length $p$.

Table 5.1: Parameters for the multi-agent simulation.

| | | Vehicle Parameters | | | |
|---|---|---|---|---|---|
| Parameter | Value | Parameter | Value | Parameter | Value |
| $m_1$ | 1.75 kg | $l_{\text{f}}$ | 0.125 m | $I_{\text{z}}$ | 0.03 kgm$^2$ |
| $m_2$ | 1.98 kg | $l_{\text{r}}$ | 0.125 m | $w$ | 0.1 m |
| | | **Simulation Parameters** | | | |
| Parameter | Value | Parameter | Value | Parameter | Value |
| $g$ | 9.81 m/s$^2$ | $T_{\text{Sim}}$ | 0.01 s | $C$ | 1.6 |
| $\mu$ | 0.85 | $B$ | 6.0 | $D$ | 1.0 |
| | | **Controller Parameters** | | | |
| Parameter | Value | Parameter | Value | Parameter | Value |
| $T$ | 0.1 s | $N_{\text{trans}}$ | 15 | $\boldsymbol{w}_u$ | $[0,0]^\mathsf{T}$ |
| $N$ | 10 | $N_{\text{ID}}$ | 2 | $\boldsymbol{w}_{\Delta u}$ | $[10,0.1]^\mathsf{T}$ |
| $N_{\text{PF}}$ | 5 | $d$ | 5 | $\boldsymbol{w}_x$ | $\mathbf{0}_6$ |
| $N_{\text{LMPC}}$ | 30 | $\alpha$ | 4 | $\boldsymbol{w}_{\Delta x}$ | $\mathbf{1}_6$ |
| $N_{\text{Race}}$ | 30 | $a_{\text{min}}$ | -1.3 m/s$^2$ | $w_d$ | -0.5 |
| $N_l$ | 4 | $a_{\text{max}}$ | 3.0 m/s$^2$ | $w_{\text{obs}}$ | $\{0.1, 0.5\}$ |
| $N_p$ | 20 | $\delta_{\text{f,min}}$ | -0.4 rad | $w_{\text{safe}}$ | $\{1.0, 0.5\}$ |
| $N_{\text{before}}$ | 15 | $\delta_{\text{f,max}}$ | 0.4 rad | | |
| $N_{\text{after}}$ | 15 | $v_{ref}$ | 1.2 m/s | | |

The simulators run at 100 Hz and publish new state information at 16.7 Hz. The controller node runs at a frequency of 10 Hz, which yields $T = 0.1\,\text{s}$, and implements the adapted LMPC for multi-agent systems. The setup of the simulation is shown in Figure 5.6, where the nodes are the ellipses and the topics are the rect-angles. The topics *pos_info* and *ecu* (Electronic Control Unit) provide the state information and the steering commands, respectively. The *agent_1/prediction* and *agent_2/prediction* are the topics to which each of the agent's prediction is published to and subsequently subscribed to from the adversarial agent. That way each opponent has perfect information of the prediction in the previous time step. All of the nodes are locally running on the MSI laptop in parallel.

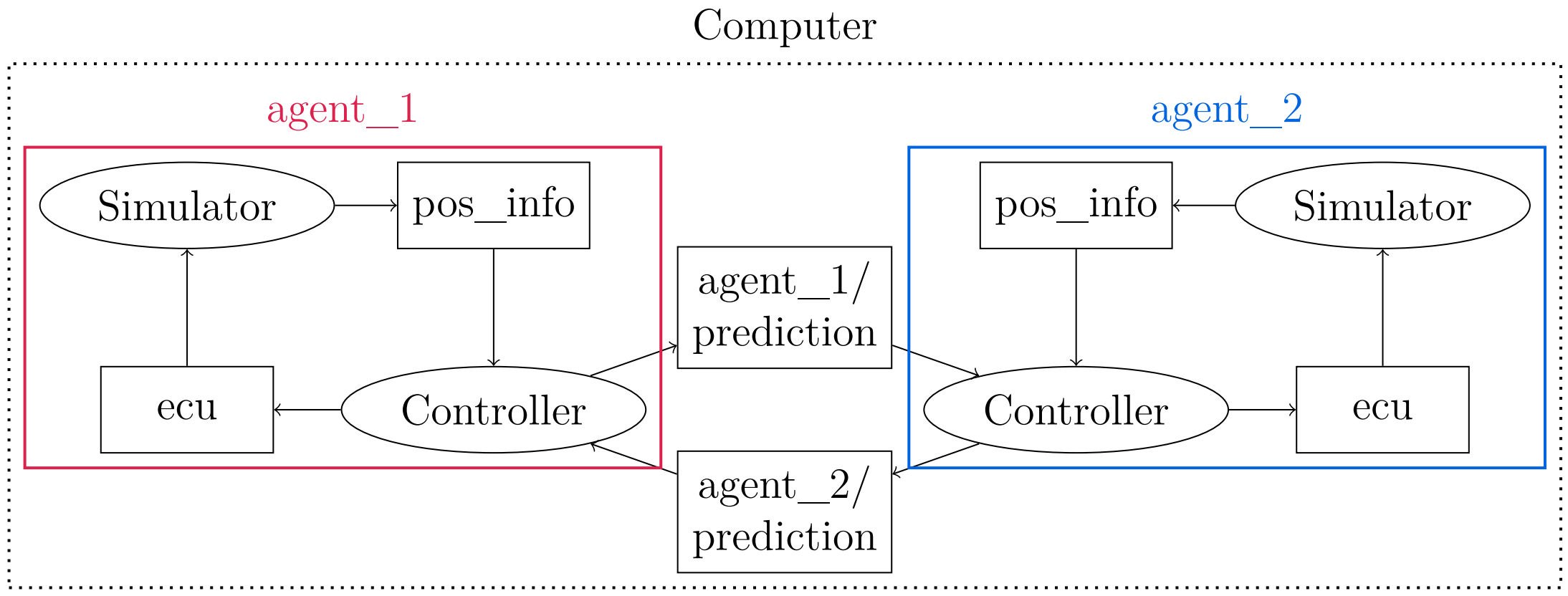

Figure 5.6: Simulation setup

### 5.2.1 Simulation Results

The multi-agent LMPC is tested in an optimal simulation environment on the L-shape race track. The first step in the proposed algorithm is the creation of the terminal safe set from multiple different initializations. For each agent three initializations are performed with a path following MPC followed by a single-agent LMPC only using the terminal safe set from a specific path following initialization. The path following initialization for the center line of the track with $e_{y,\mathrm{ref}} = 0$ is displayed in Figure 5.7.

In Figure 5.8, the choice of parameters does not allow the controller to take a more optimal trajectory, since the selected horizon is quite short. However, this has the advantage, that the learning progress is slower and the controller does not deviate a lot from the previously visited states. Eventually this allows for a more diverse exploration of states. Therefore, the horizon behaves as a learning rate. Nevertheless, the controller displays expected behavior, because it breaks in curves and accelerates in straight sections. The agent leaves the track multiple times, which is allowed because of the slack variables. Increasing the weight on this slack can cause even more undesirable behavior, where the agent applies high counter steering inputs, which results in a sinusoidal trajectory. The learning progress can be inferred from Figure 5.9. Starting from the path following initialization, the velocity of the vehicle increases with every iteration until it converges to a local optimal trajectory for later iterations. The velocity profile is not constant, because of the existing turns on the race track, where the agent breaks as expected.

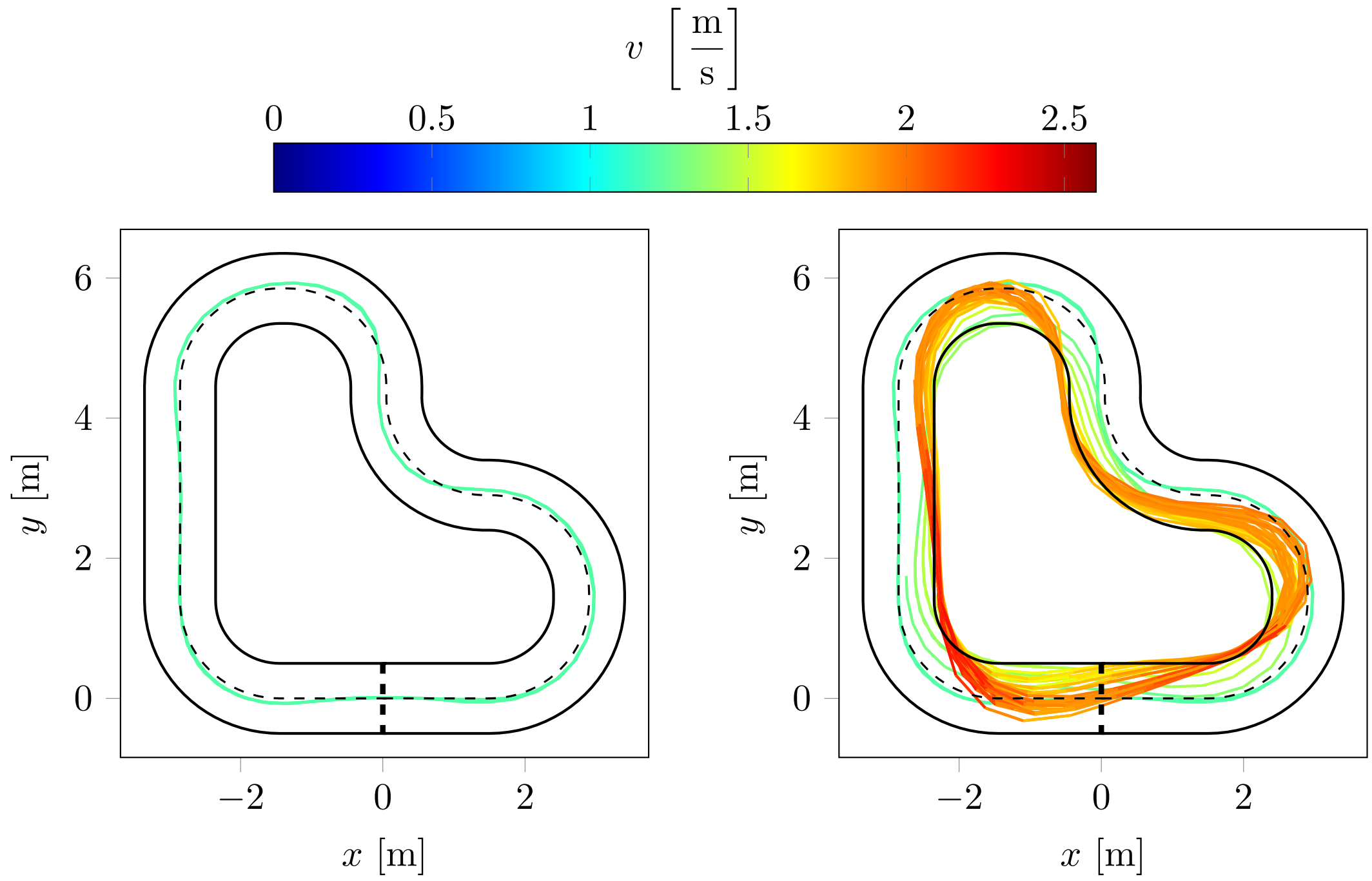

Figure 5.7: Path-following for $e_{y,ref} = 0.0$.

Figure 5.8: LMPC for center initialization.

The second initialization, starts from a path following MPC with $e_{y,\text{ref}} = 0.375$, followed by 30 iterations of LMPC, which is shown in the Figure 5.10 and Figure 5.11. The reference value is not chosen to be directly on the boundary in order to prevent most of the initialization to be outside of the track boundaries.

Similarly, the third initialization is set to be close to the outside of the track boundaries. Therefore, the reference value of $e_{y,ref} = -0.375$ is chosen for the path following MPC for the creation of an initial terminal safe set. Subsequently, this serves as the initialization for the LMPC. The results from the path following MPC and the following LMPC are shown in Figure 5.12 and Figure 5.13, respectively.

LMPC for nonlinear systems only converges to locally optimal solutions. Therefore, the three final trajectories for each initialization are displayed in Figure 5.14 for comparison. The trajectories exhibit quite similar behavior for the different initializations. Interestingly, at the bottom of the track the final trajectories for the outer and inner initialization are very close and nearly overlap. However, the two trajectories diverge in the first curve and the final trajectories for the outer and center initialization start to overlap for one more curve. Furthermore, the inner initialization results in the lowest velocities, which allows the vehicle to be steered around curves more tightly.

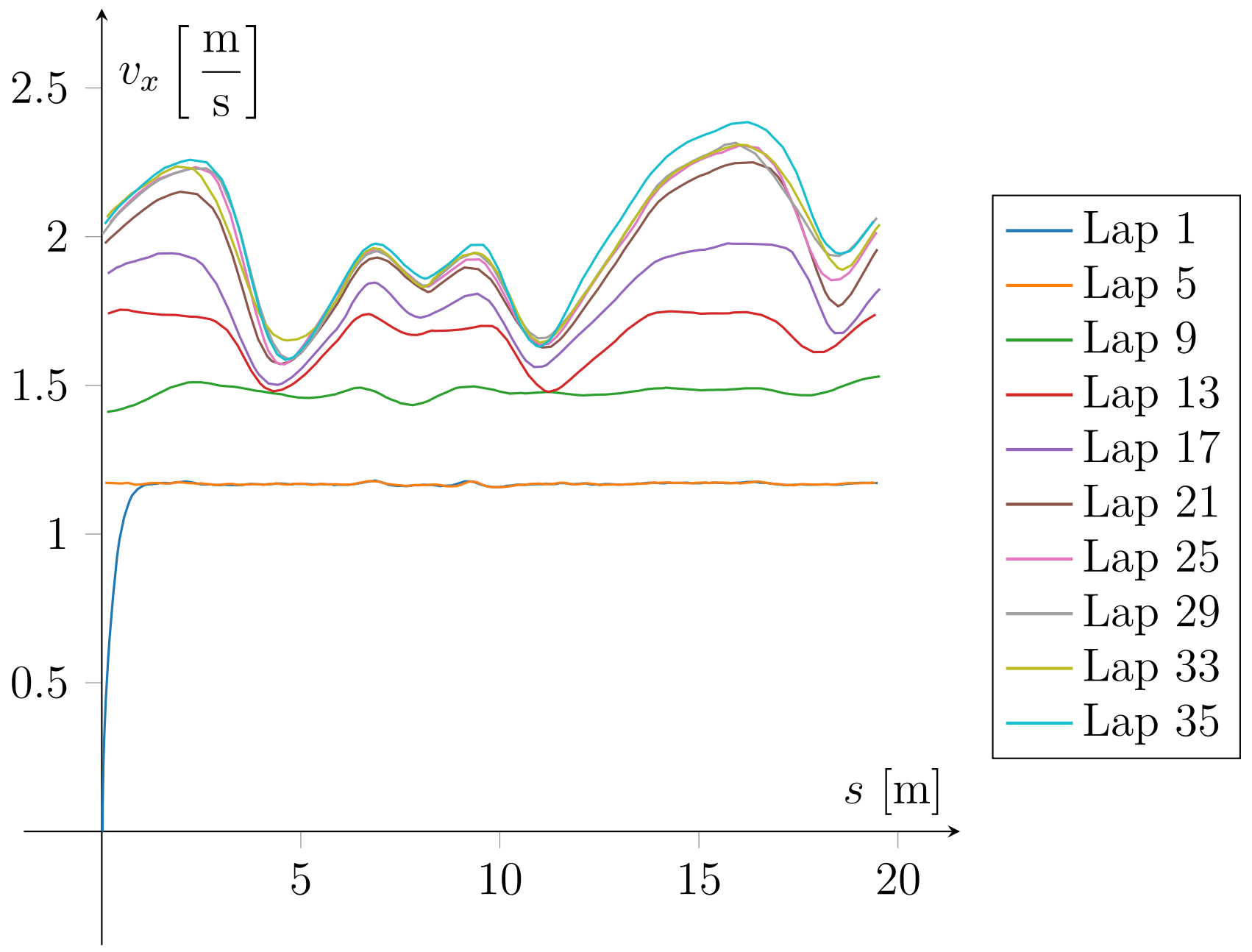

Figure 5.9: Velocity profile for the LMPC for the center initialization.

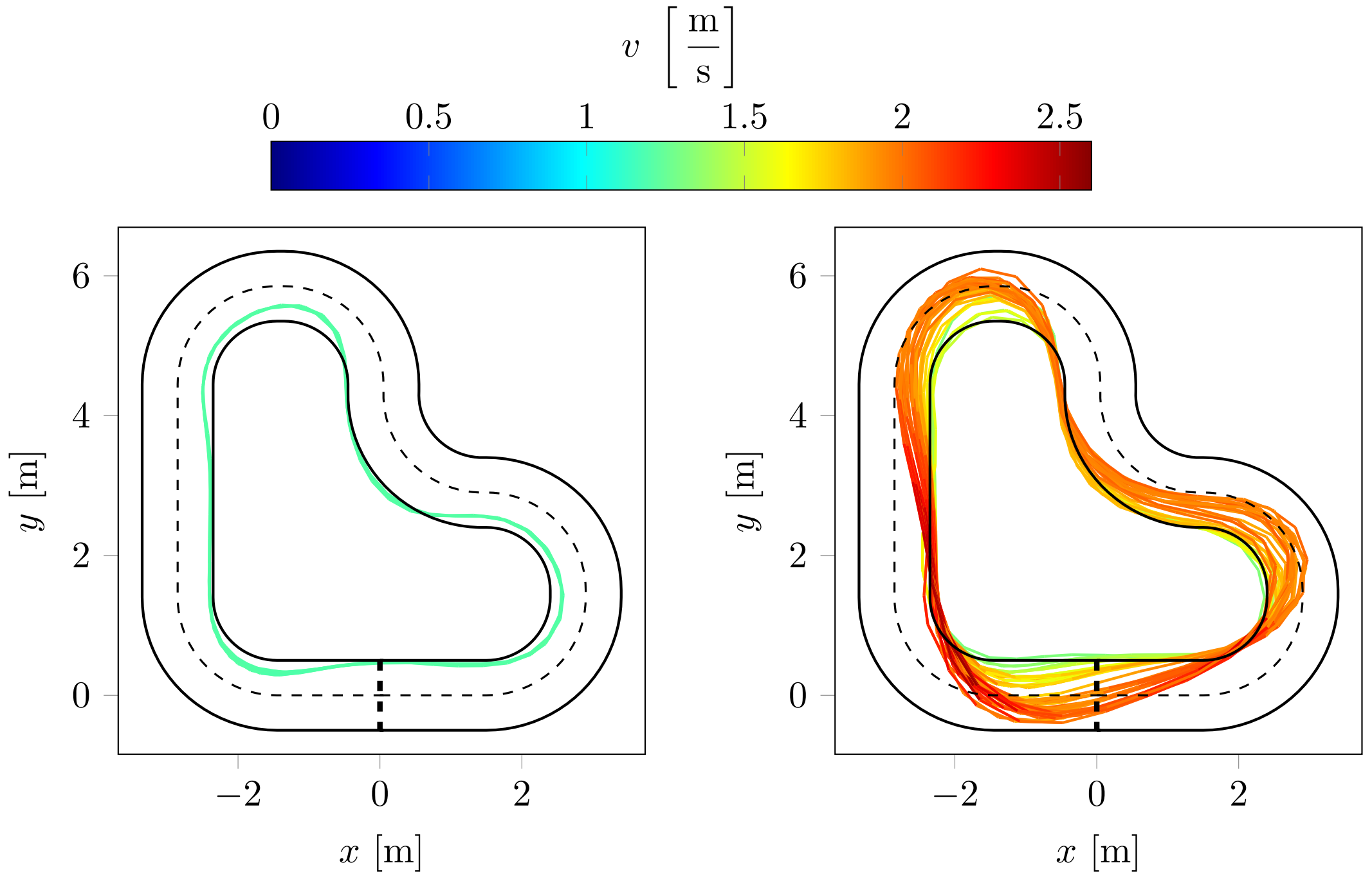

Figure 5.10: Path-following for $e_{y,ref} = 0.375$.

Figure 5.11: LMPC for inner initialization.

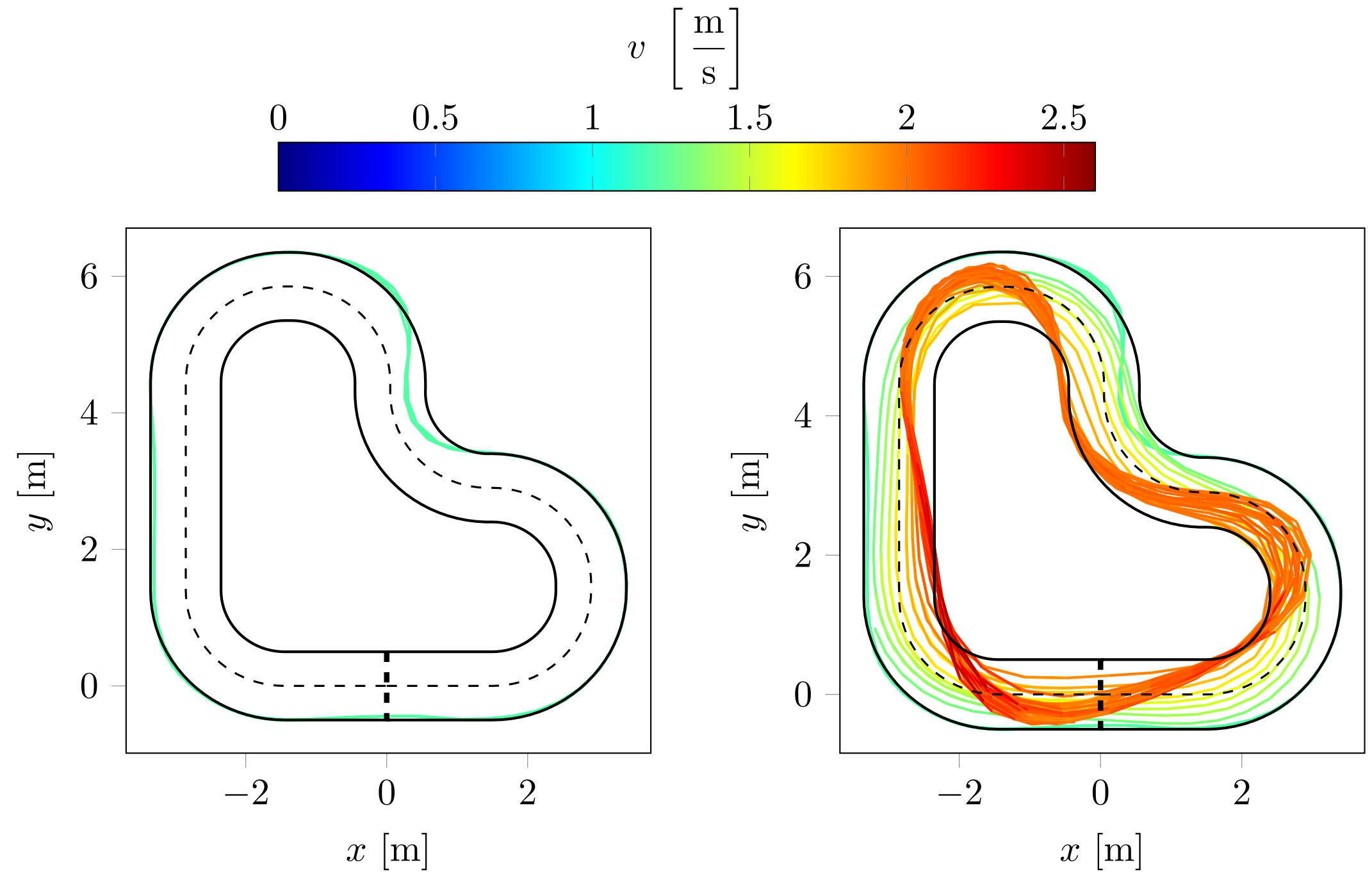

Figure 5.12: Path-following for $e_{y,ref} = -0.375$.

Figure 5.13: LMPC for outer initialization.

The convergence for LMPC for both agents on all of the initializations, is portrayed in Figure 5.15 and Figure 5.16 for agent 1 and agent 2, respectively. This shows that the number of iterations nearly monotonically decreases for all iterations. In general, the number of iterations is nearly the same for each executed LMPC for both agents. Nevertheless, there are exceptions, e.g. at iteration 8 for the outer initialization for agent 2, which reaches a local minimum. This can be caused by the use of slack variables or high model mismatch. The minimum time for terminating one iteration by agent 1 is 8.6 s, while agent 2 achieves a lap time of 8.8 s, given that $m_1 < m_2$.

Following all the above initializations, the terminal safe sets for each agent for every initialization are combined for the racing scenario with two agents running simultaneously. In comparison to the terminal safe set created by the LMPC based on the center initialization, the combination of multiple initializations yields a richer terminal safe set. The combination of all explored states during the initialization for agents 1 and 2 are displayed in Figure 5.17 and Figure 5.18, respectively.

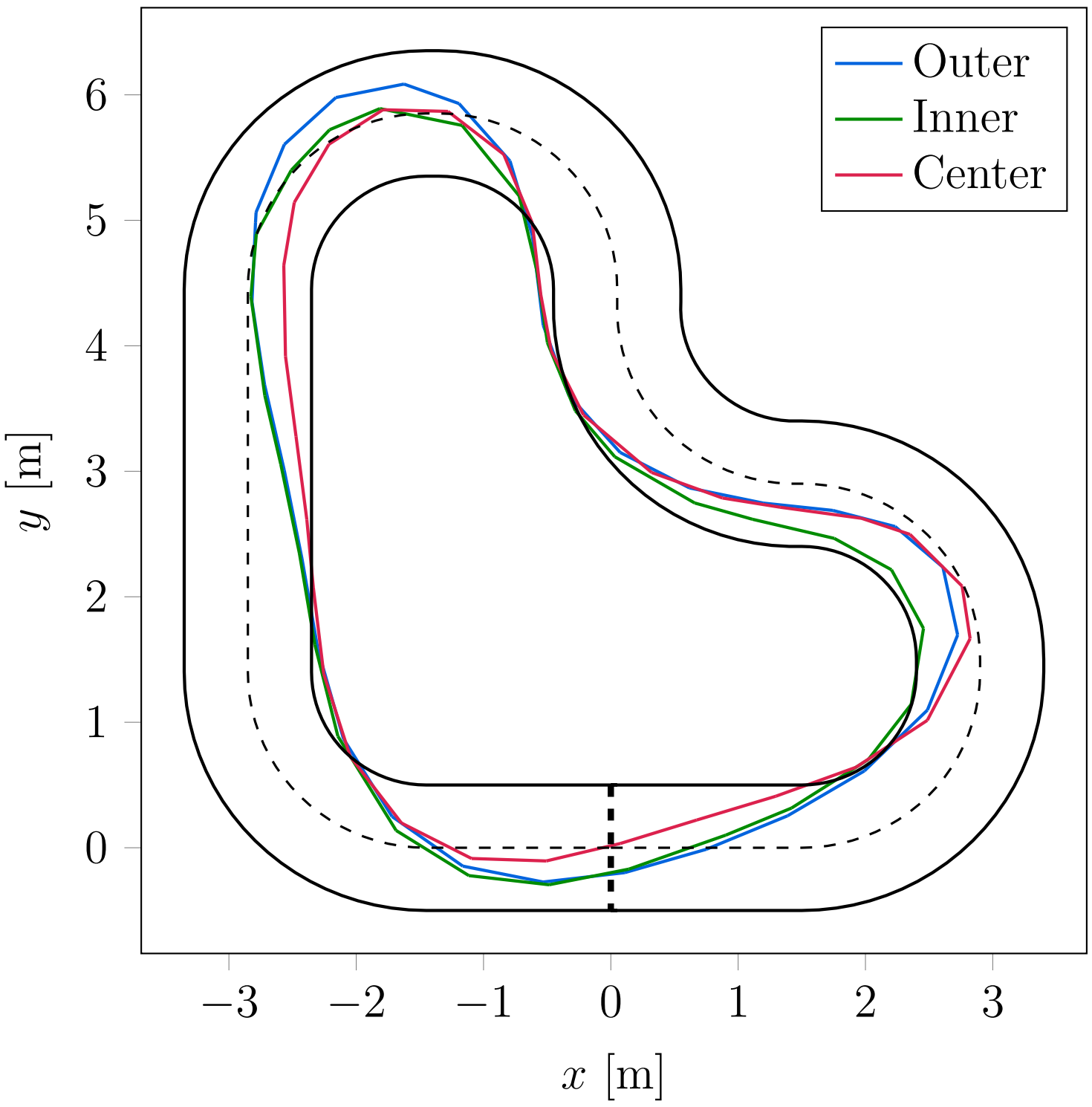

Figure 5.14: Last open-loop trajectory of the LMPC for the different initializations.

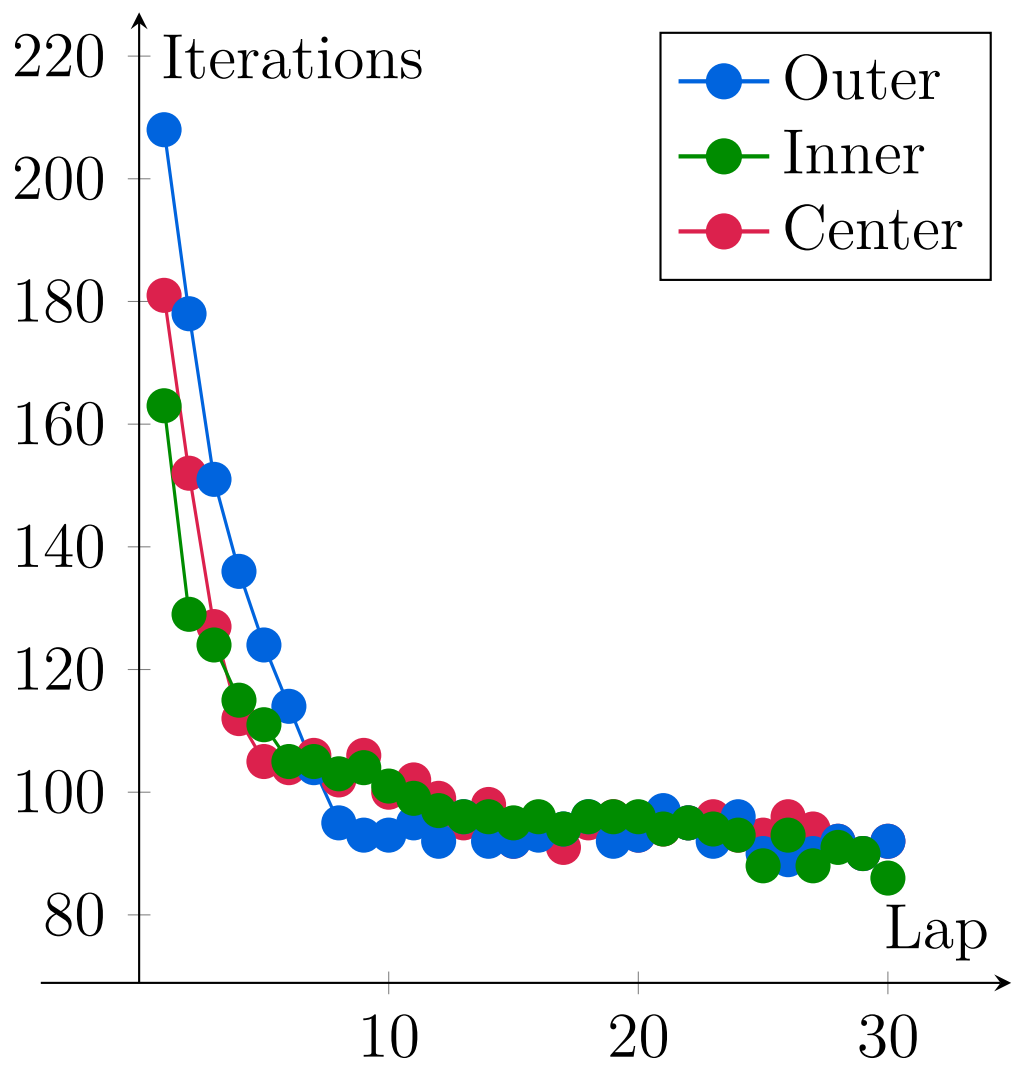

Figure 5.15: Number of time steps needed to complete each lap for agent 1.

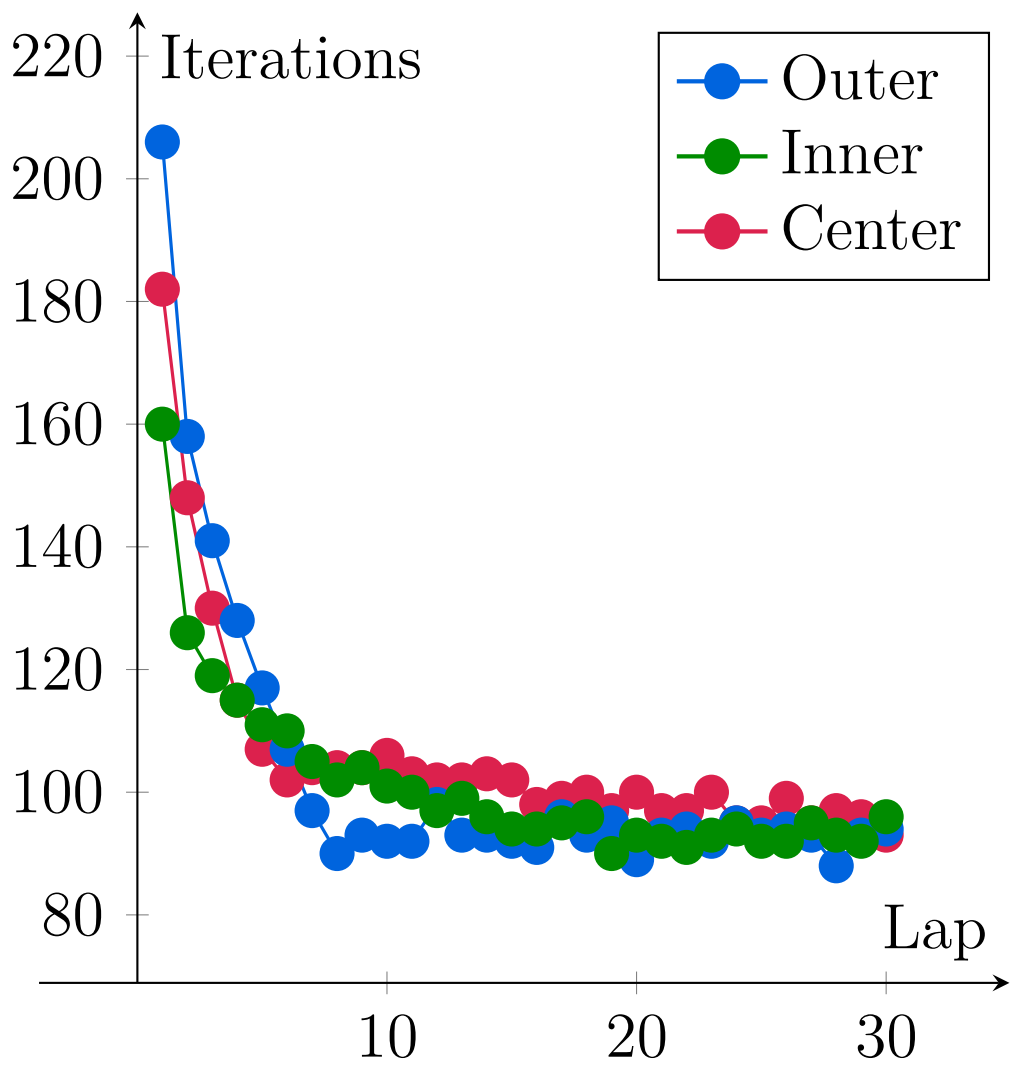

Figure 5.16: Number of time steps needed to complete each lap for agent 2.

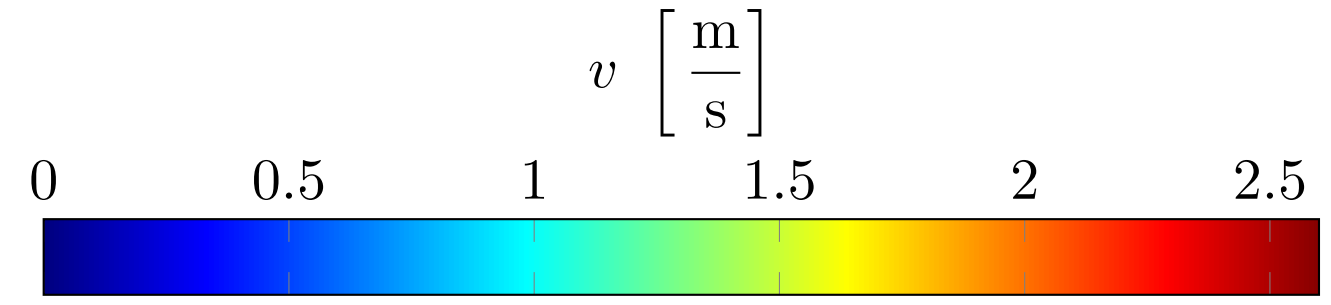

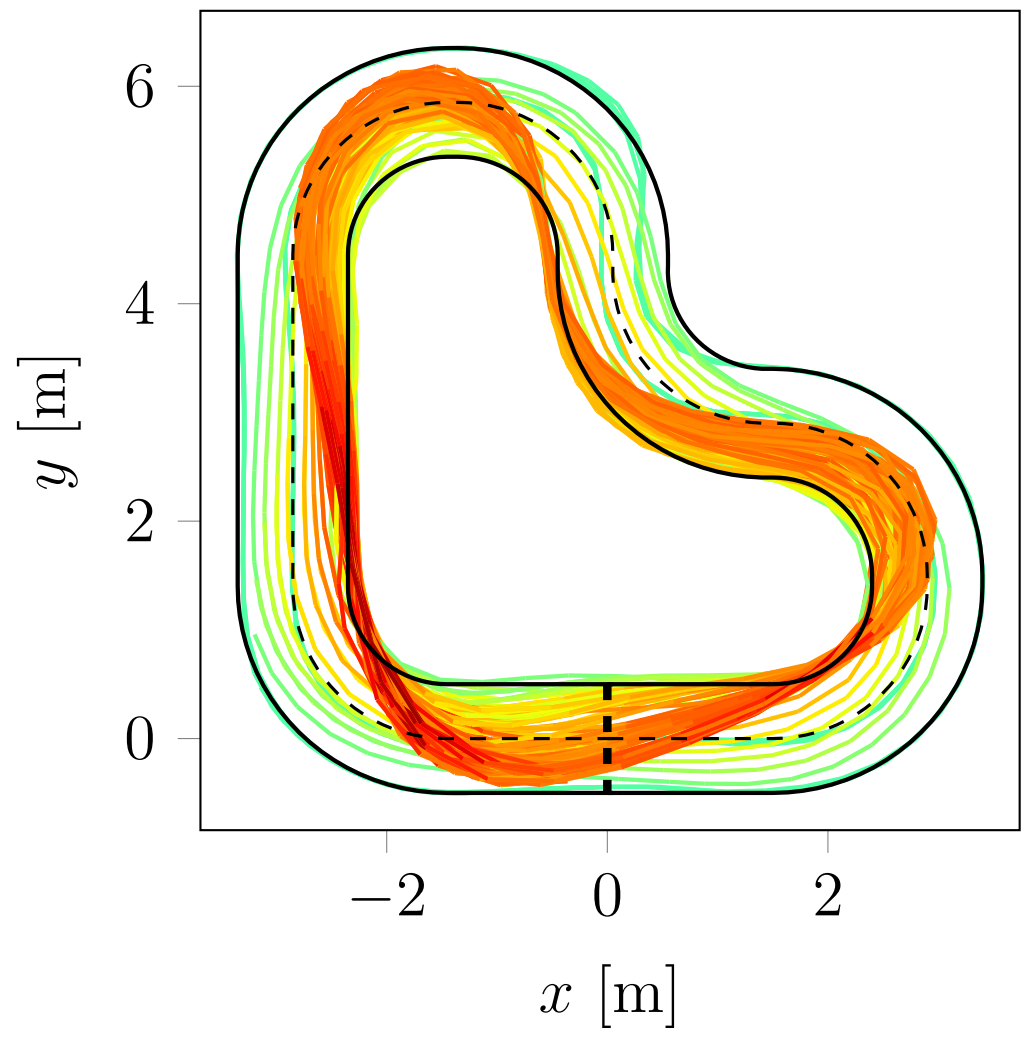

Figure 5.17: Agent 1's combined safe set after all initializations.

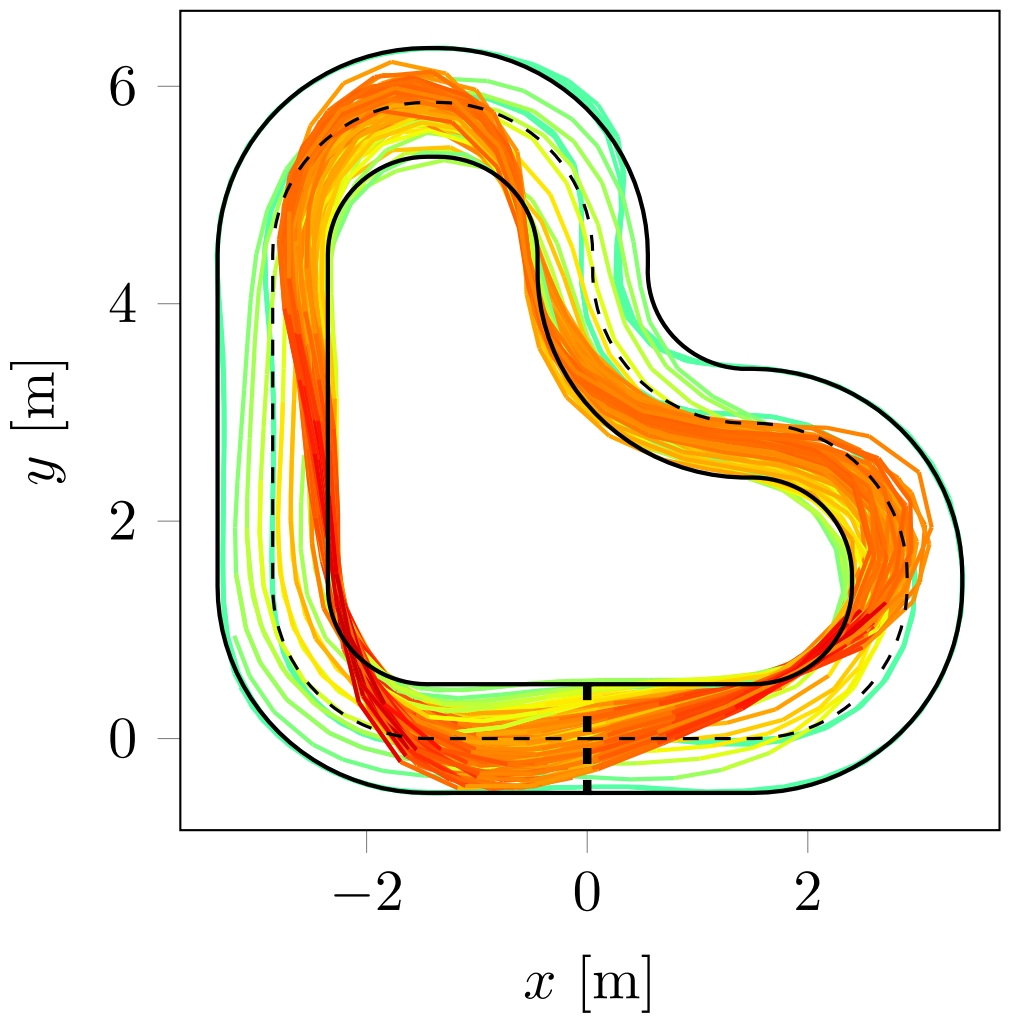

Figure 5.18: Agent 2's combined safe set after all initializations.

Executing the multi-agent LMPC for the two agents based on the previously collected states for the terminal safe set, results in the closed-loop trajectories for agent 1 and 2 shown in Figure 5.19 and Figure 5.20, respectively. Since both agents nearly reach a similar velocity and the track is quite long, there are very few interactions among the two agents. Therefore, a soft constraint on the velocity of the second is added with $v_{\max} = 1.5\,\mathrm{m/s}$. That way the faster car is able to overtake the slower vehicle multiple times over the course of the race and multiple interactions are recorded. Because of the state-varying safe set, the agent is immediately able to take advantage of the prerecorded states. The agent quickly accelerates up to high velocities, which were previously only achieved towards the end of the executed single-agent LMPCs. Agent 1 regularly diverges from the locally optimal trajectory determined by the LMPC to take suboptimal trajectories, which allow for overtaking maneuvers. Consequently, the multi-agent LMPC does not necessarily need to demonstrate a non-decreasing cost and lap time. The suboptimal lap time has an upper bound, which is depended on the states in the terminal safe set. Because the agent keeps adding the recorded iterations from the race to the terminal safe set as well, the agent is able to complete overtaking maneuvers at higher velocities. The recorded trajectories also show, that the application of the modified safe set for overtaking maneuvers behaves as intended. The controller successfully overtakes on the left and the right of the obstacle. Moreover, agent 1 prevents crashes with the adversary by going off the track.

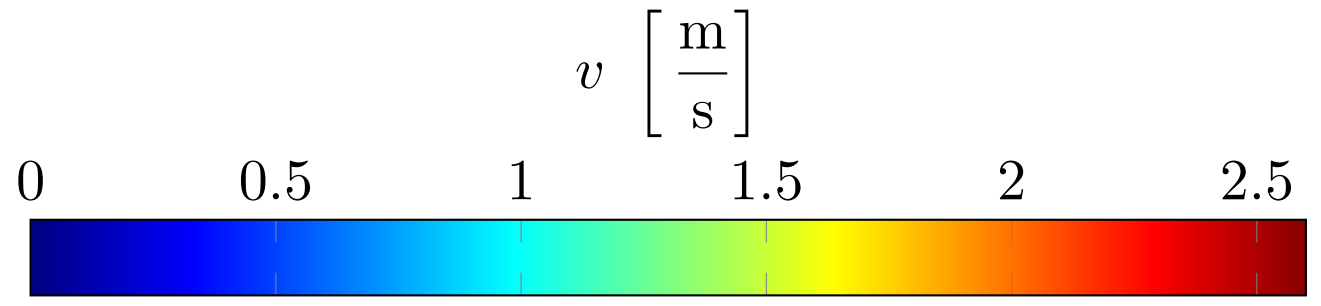

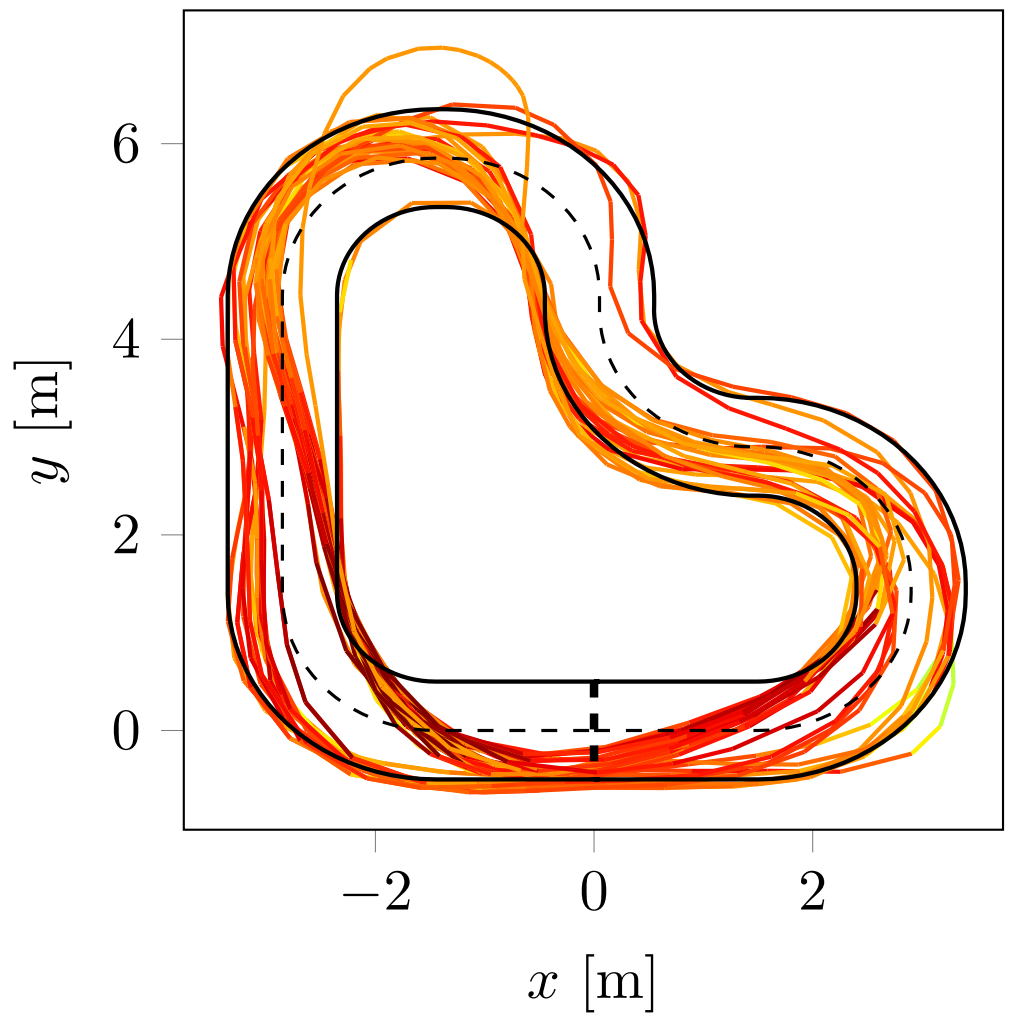

Figure 5.19: Closed-loop racing trajectory for agent 1.

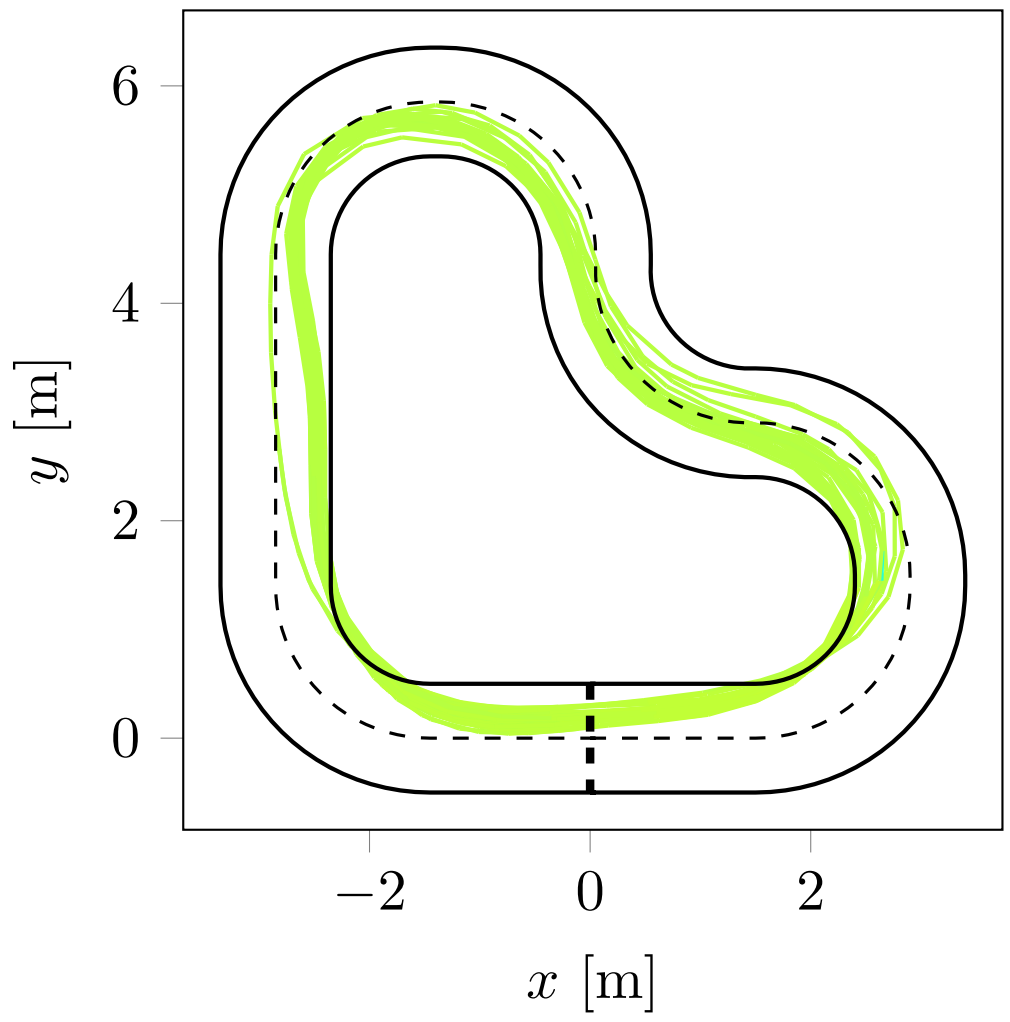

Figure 5.20: Closed-loop racing trajectory for agent 2.

One of the successful overtaking maneuvers is depicted in Figure 5.21. The overtaking maneuver occurs at $t_{\text{race}} = 97.84\,\text{s}$ in the race, when agent 1 is in iteration $j^1 = 9$ at time step $t^1 = 3$ and agent 2 has just started lap 8 and is at time step $t^2 = 2$. The slower agent 2, here in blue, tighly cuts the curve, while the faster agent 1 does not take the curve tightly, but then cuts in front the opponent and successfully terminates the maneuver.

## 5.3 Experimental Validation

The Berkeley Autonomous Race Car (BARC) is an open-source 1/10 scale vehicle platform developed at the Model Predictive Control Lab at the University of California, Berkeley, see Figure 5.22.

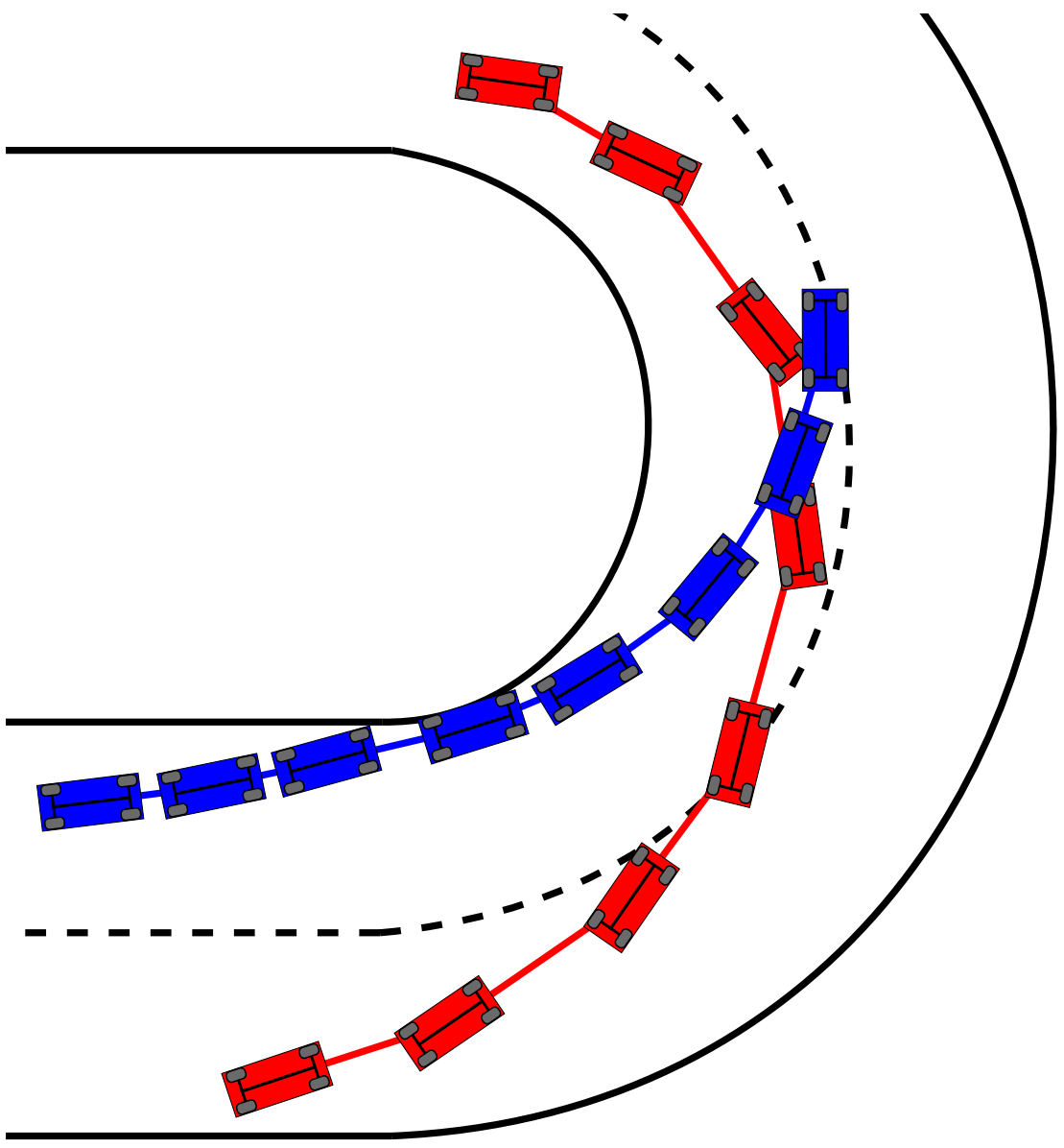

Figure 5.21: Overtaking maneuver

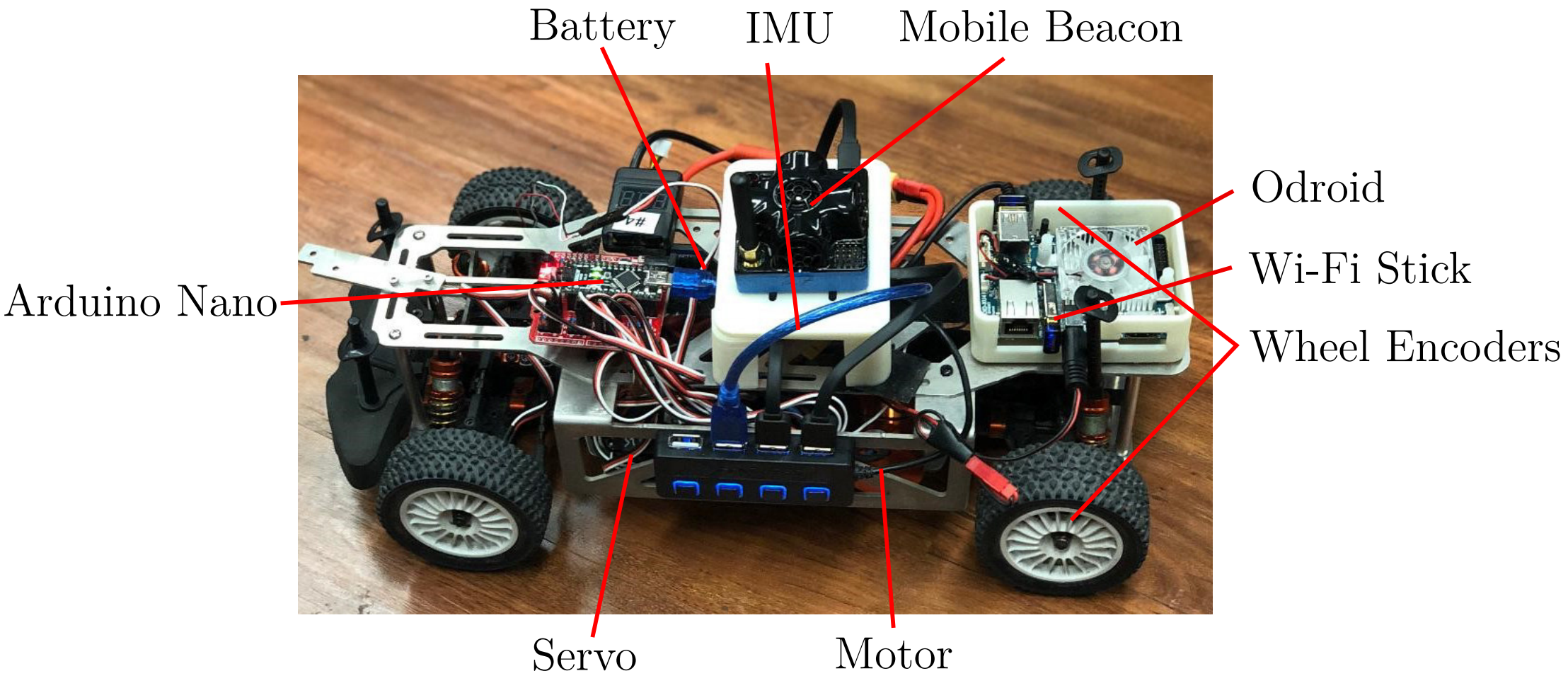

Figure 5.22: Berkeley Autonomus Race Car

The BARC provides a low-cost experimental system for controllers and self-driving car algorithms. The Basher RZ-4 1/10 Rally Racer, which is a toy race car, serves as the base of the BARC. Components for communication, computation and sensing add the functionalities necessary for testing control algorithms. Among the components is the Odroid-XU4, a single-board computer with two quad core CPUs with frequencies of 2.1 GHz and 1.3 GHz, respectively, 2 GB of RAM, running Ubuntu MATE, a Linux distribution, and is set up with a Wi-Fi USB stick that connects to a local network. Furthermore, the Arduino Nano microcontroller, a brushless DC motor and a steering servo are part of the BARC system. Additionally, the BARC is equipped with two rear wheel encoders, an IMU and a mobile beacon from a local GPS system. A detailed description of the sensors follows in subsection 5.3.1. The whole system is powered by a two-cell LiPo Battery.

During operation, a ROS node, which runs remotely on the Odroid, receives steering and motor commands over the wireless network from the computer running the controller node. This node then sends the commands to the Arduino Nano, which applies the commands to the motor and servo. In addition, there is one node running for each sensor, which publish the most recent sensor readings. The state estimation node subscribes to the sensor messages and estimates the current state. The implementation of the state estimator is discussed in section 5.3.2. Table 5.2 provides information and parameters of the two BARCs, which were used in the experiments. Note that $C_\text{f}$ and $C_\text{r}$ are the front and rear cornering stiffness, respectively, which are necessary for the state estimation and are identified in section 5.3.2 for both vehicles.

Table 5.2: Specification of the BARCs.

| | | BARC 1 | BARC 2 |
|---|---|---|---|
| Layout | | 4WD | RWD |
| Differential Gears | | ✓ | × |
| $m$ | [kg] | 2.00 | 1.75 |
| $l_\text{f}$ | [m] | 0.125 | 0.125 |
| $l_\text{r}$ | [m] | 0.125 | 0.125 |
| $I_\text{z}$ | [kgm$^2$] | 0.24 | 0.24 |
| $C_\text{f}$ | [N/rad] | 12.38 | 13.03 |
| $C_\text{r}$ | [N/rad] | 9.60 | 10.06 |

### 5.3.1 Sensors

The sensors enable the determination of the vehicle state from their measurements. The following describes each of the low-cost the sensors in more detail.

#### IMU

The Inertial Measurement Unit (IMU) measures linear ($x$, $y$ and $z$) and angular orientation and its corresponding angular accelerations (roll, pitch and yaw). The BARC is equipped with the myAHRS+ IMU, which publishes new measurements at 50 Hz to a ROS topic, while the data is transfered through a USB connection to the Odroid.

### Wheel Encoders

The two rear wheels are outfitted with a magnetically operating encoder each. Four magnets are equally distributed on the inside of each wheel and a hall effect sensor, which detects the change in the magnetic field, is mounted on the wheel hub. With every quarter rotation the magnetic field detected by the hall sensor changes and the time difference $\Delta t$ between changes is measured. Using the wheel radius $r_{\text{wheel}} = 3.6\,\text{cm}$, the velocities $\tilde{v}_{r,r}$ and $\tilde{v}_{r,l}$ for the right and left rear wheel can be determined by

$$\tilde{v}_{r,i} = \frac{\pi r_{\text{wheel}}}{2\Delta t_i}, i \in \{r, l\}. \tag{5.42}$$

Alternatively, it is possible to calculate the frequency at which the encoder readings are updated

$$f = \frac{1}{\Delta t_i} = \frac{2\tilde{v}_{r,i}}{\pi r_{\text{wheel}}}, i \in \{r, l\}. \tag{5.43}$$

Consequently, an update rate greater than 10 Hz requires the velocity of each rear wheel $\tilde{v}_{r,i} > 0.57\,\text{m/s}, i \in \{r, l\}$. The measurements for both rear wheels are averaged to create the measurement

$$\tilde{v}_x = \frac{\tilde{v}_{r,r} + \tilde{v}_{r,l}}{2}. \tag{5.44}$$

The encoder node publishes at a constant frequency of 50 Hz, therefore, for lower velocities the measurement will not necessarily change between two measurements. This also depends on the angular offset of the magnets on each wheel, which do not need to be synchronized.

### GPS

The vehicle's absolute position is determined by an indoor positioning system from Marvelmind Robotics. Four stationary beacons around the race track triangulate the position of the mobile beacon mounted on the BARCs. A similar approach is applied for the Global Positioning System (GPS), which is why the indoor positioning system is referred to as a GPS as well. The distances between two beacons for the triangulation are calculated by the time of flight of an ultrasonic wave signal. The manufacturer specifies a positional accuracy of ±2 cm. The operation of the beacons is controlled by the Marvelmind Dashboard software, which runs on a Windows computer. The Marvelmind Dashboard allows for creating the map for the beacons, changing the radio frequencies and monitoring the state and signal quality for all beacons. The calculation of the mobile beacon's position is executed by a modem, which is connected to the computer running the MARVELMIND DASHBOARD. The modem receives all distance measurements from the stationary beacons to the mobile beacons and has access to the absolute positions of the beacons from the MARVELMIND DASHBOARD. This information is sufficient to update the position of the mobile beacon, which is subsequently communicated to the mobile beacon and is received on the Odroid through a USB connection. The system can operate on different sampling frequencies, which can be set in the MARVELMIND DASHBOARD. A frequency of 12 Hz has proven to be the highest frequency, which still demonstrates a reasonably accurate position signal. However, using two mobile beacons simultaneously results in a reduction of the frequency by a factor of two, resulting in a frequency of 6 Hz. This is due to the fact, that every mobile beacon requires an updated position signal, but the overall sampling frequency of the system does not change. Therefore, the signal frequency is increased by an additional IMU integrated in the mobile beacon, which allows for an onboard fusion of the GPS and IMU signal to provide a position signal with a frequency of up 100 Hz.

### 5.3.2 Experimental Setup

The hardware setup for experiments with a single-agent is detailed in Figure 5.23. The execution of experiments with the BARC require space, which was provided by a seminar room at Etcheverry Hall at the UC Berkeley. The space was sufficiently big for setting up the oval track. A Wi-Fi router establishes a local Wi-Fi network, which the MSI laptop and the Odroids connect to. The exchanged Wi-Fi signals are displayed in blue in the figure. Furthermore, the stationary GPS beacons are placed around the track, such that the convex hull of the sensors contains the whole track. Effectively, the GPS coverage should also take into account for the case that any of the vehicles go off track. Once the absolute and relative position of the stationary beacons are determined with the MARVELMIND DASHOBARD on the Windows computer and the modem, the mobile beacon on the BARC can be localized. The ultrasonic signals from the stationary beacons are shown in red. Note, that the mobile beacon also emits ultrasonic signals, which are not displayed for the sake of clarity. The Windows computer used in the experiments for the MARVELMIND DASHBOARD is a Lenovo Thinkpad Yoga from 2014. Extending this setup for the second BARC only requires adding the second BARC and connecting it to the local network. A second controller node can then send steering commands to this agent as well.

The structure of the physical multi-agent ROS program differs from the setup in the simulation, such that the simulator is replaced by the sensor and estimator nodes. In addition, the nodes run on multiple different devices. This is han-dled by the MSI laptop as the ROS master machine. The MSI laptop is called "Computer" in Figure 5.24. The master manages the launch of all nodes on the devices. The master launches the two controller nodes for both agents on its own machine and launches an Arduino-node, an IMU-node, a GPS-node and an Estimator-node on each of the slave machines Odroid_1 and Odroid_2. Each Arduino-node receives the steering and acceleration as pulse-width-modulation signals, which are described in section 5.3.2, and applies them to the actuators. Furthermore, the Arduino-node receives the measurements from the wheel encoders and publishes the calculated estimate to the *vel_est* topic. The IMU-nodes and GPS-nodes take the current measurements and publish them to their respective ROS topics. All measurements are combined in the Estimator-node, which implements an extended Kalman filter. A detailed description of the implementation can be found in section 5.3.2. The Estimator-nodes then publish to their respective *pos_info* topics at 50 Hz, which contain the necessary state information for the LMPC. Running the estimator on the Odroid has the advantage, that the sensor measurements do not have to be send over the Wi-Fi connection. This reduces the amount of data being send between the Odroid and the laptop to the control inputs and the current state information. Consequently, there is less traffic on the network. Due to the possible interference of different operating Wi-Fi networks, reducing the amount of data sent minimizes a potential package loss.

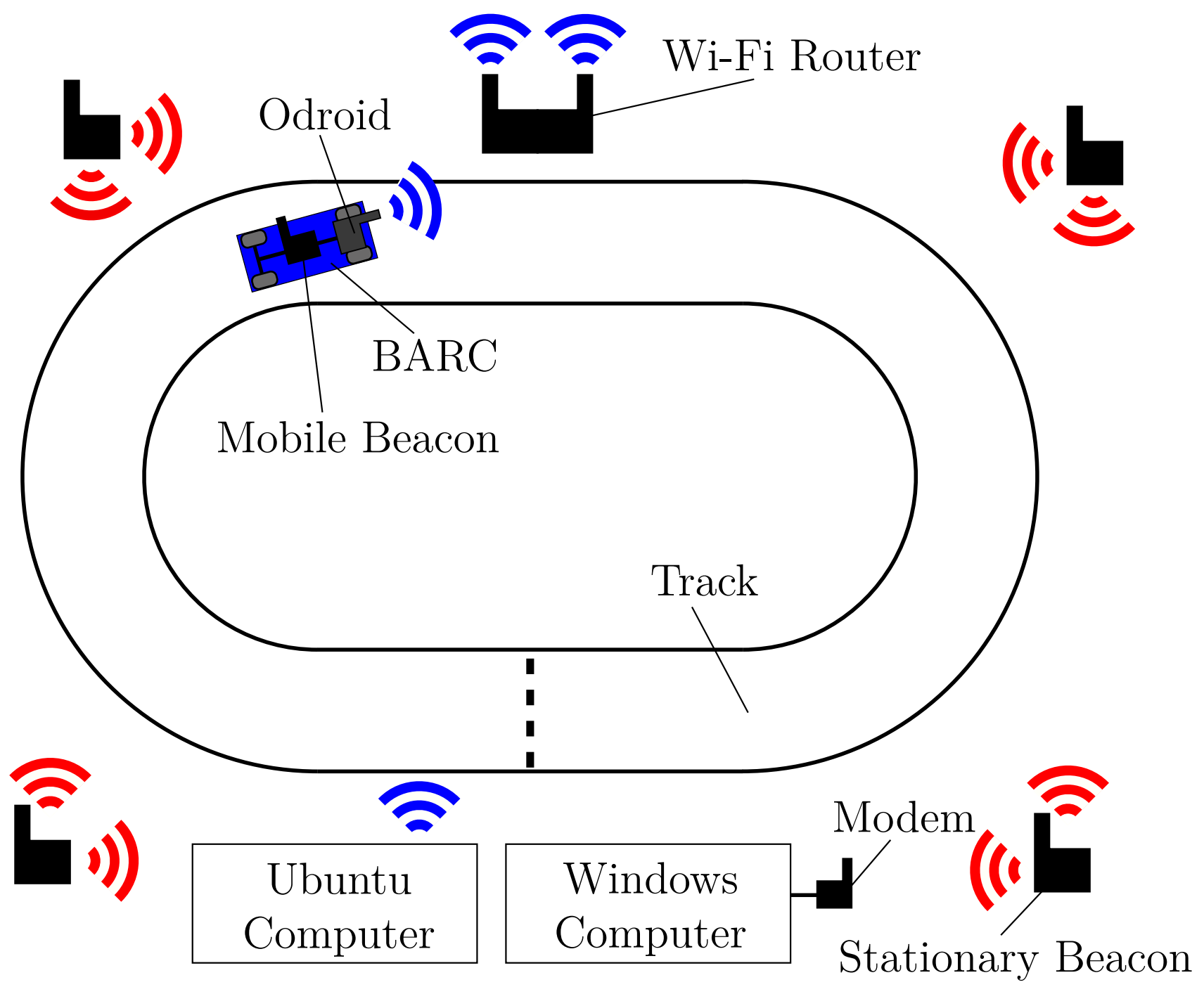

Figure 5.23: BARC experiment physical setup

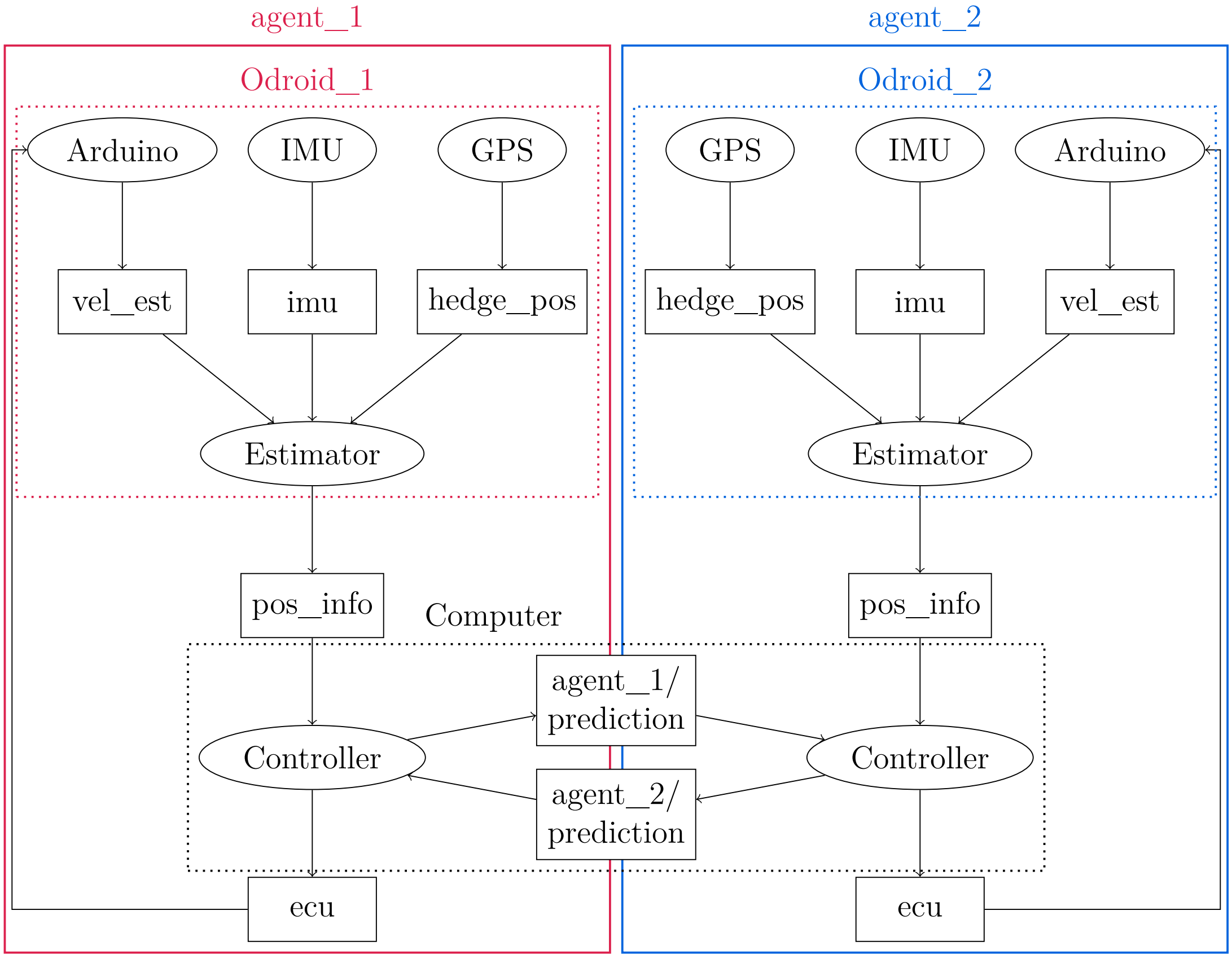

Figure 5.24: BARC experiment software setup

### Steering Map

Whereas the presented controller uses steering and acceleration inputs in SI units, the actuators use pulse-width-modulation (PWM) signals. PWM signals are in the range $[0, 180]$, which determine the turning angle of the servo. Therefore, a function $f_{\mathrm{PWM}}(\delta_{\mathrm{f}})$ mapping the steering angle $[-\pi, \pi] \to [0, 180]$ has to be determined.

The identification of the steering map is achieved by applying different steering PWM signals to the steering servo, while keeping a constant acceleration. This results in the vehicle reaching a steady state and following circular trajectories of different radii with a constant velocity. If this velocity is sufficiently low, the kinematic bicycle model is still a valid approximation of the vehicle dynamics. In this case the steering angle $\delta_{\mathrm{f}}$ is given by

$$\delta_{\mathrm{f}} = \arctan\left(\frac{\dot{\psi}(l_{\mathrm{f}} + l_{\mathrm{r}})}{v_x}\right), \tag{5.45}$$

where only $\dot{\psi}$ and $v_x$ are needed for the calculation. Both quantities are accurately obtained from the IMU and the encoders. The execution of this open-loop test yields different steering angles for the applied PWM values. This relationship is then approximated with an affine function by applying a linear regression. The mappings for both BARCs are determined as

$$f^1_{\mathrm{PWM}}(\delta_\mathrm{f}) = 89.1\delta_\mathrm{f} + 81.4 \tag{5.46}$$

$$f^2_{\mathrm{PWM}}(\delta_\mathrm{f}) = 103.1\delta_\mathrm{f} + 83.3. \tag{5.47}$$

### Cornering Stiffness Identification

The cornering stiffness identification is formulated as an optimization problem following an approach from [ZhangEtAl17]. A set of equations for the lateral tire forces, $F_{y,\mathrm{f}}$ and $F_{y,\mathrm{r}}$, is introduced, where the lateral tire forces directly depend on the front and rear cornering stiffnesses, $C_\mathrm{f}$ and $C_\mathrm{r}$, respectively. With the assumption of small steering angles, the lateral tire forces are approximately given by

$$F_{\mathrm{y,f}} = C_\mathrm{f}\left(\delta_\mathrm{f} - \beta - \frac{l_\mathrm{f} r}{v_x}\right) \tag{5.48}$$

$$F_{\mathrm{y,r}} = C_\mathrm{r}\left(-\beta + \frac{l_\mathrm{r} r}{v_x}\right). \tag{5.49}$$

Similar to section 5.3.2, the cornering stiffness identification is conducted while the vehicle is at steady state and therefore, constant accelerations and steering angles are applied. At steady state it is assumed that $\dot{v}_y = 0$ and $\ddot{\psi} = 0$. Then Equation 2.36 and Equation 2.37 with the small angle approximation yield

$$0 = \frac{1}{m v_x}\left(F_{y,\mathrm{f}} + F_{y\mathrm{r}}\right) - r \tag{5.50}$$

$$0 = \frac{1}{I_\mathrm{z}}\left(l_\mathrm{f} F_{\mathrm{y,f}} - l_\mathrm{r} F_{\mathrm{y,r}}\right). \tag{5.51}$$

The optimization problem is then formulated as

$$\min_{C_\mathrm{f}, C_\mathrm{r}, \beta} \sum_{k=0}^{K} \left[\left(\frac{1}{m\bar{v}_{x,k}}\left(\bar{F}_{\mathrm{f},k} + \bar{F}_{\mathrm{r},k}\right) - \bar{r}_k\right)^2 + \left(\frac{1}{I_\mathrm{z}}\left(l_\mathrm{f}\bar{F}_{\mathrm{f},k} - l_\mathrm{r}\bar{F}_{\mathrm{r},k}\right)\right)^2\right] \tag{5.52}$$

$$\text{subj. to } \beta_k \in [-1, 1], \ \forall k \in \{0, ..., K-1\}, \tag{5.53}$$

where $K$ is the number of open-loop experiments for different accelerations and steering angles. The lateral tire forces $\bar{F}_{\mathrm{f},k}$ and $\bar{F}_{\mathrm{r},k}$ are the averaged calculated forces over all recorded measurements for experiment $k$. Equivalently, $\bar{v}_{x,k}$ and $\bar{r}_k$ are the averaged measured longitudinal velocity and yaw rate, respectively, for experiment $k$. The constraint on the elements of $\boldsymbol{\beta}$ is added, such that they provide reasonable values. The values of the identified cornering stiffnesses for both BARCs are given in Table 5.2.

### State Estimation

The accuracy of the state estimation is of significant importance to the performance of the proposed controller, since it does not only affect the accuracy of the state estimate, but also directly influences the online linear regression model. In general the state estimation and sensor fusion are challenging tasks, because different sensors supply measurements at various frequencies and noise levels. The Kalman filter algorithm for linear systems by [Kalman60] is capable to filter noise and is able to accurately estimate unobserved variables, using a joint probability distribution for each time step. Therefore, the Kalman filter is applied for the given estimation problem of determining the dynamic vehicle state from the provided measurements of the sensors. Because of the nonlinearity of the vehicle dynamics, the standard Kalman filter cannot be applied and the extended Kalman filter (EKF) is used instead [AndersonMoore79]. In this application two different EKFs are running in parallel, whose operating ranges are based on the current longitudinal velocity of the vehicle. The first EKF implements a kinematic model for small longitudinal velocities and the second EKF implements a dynamic bicycle model for high longitudinal velocities. Once the vehicle reaches a longitudinal velcoity of $v_x > 0.5\,\mathrm{m/s}$, the estimation changes from the EKF using the kinematic point-mass model to the EKF with the dynamic bicycle model.

The algorithm of the standard Kalman filter and the EKF are similar. However, in contrast to the standard Kalman filter formulation, the EKF allows the use of a nonlinear model, which is linearized at every time step. The state estimation is divided into two steps: the prediction step and the update step. First, the prediction step determines a prediction of the next state $\hat{\boldsymbol{x}}_{k|k-1}$ using the previously estimated state $\hat{\boldsymbol{x}}_{k-1|k-1}$ and the applied control input $\boldsymbol{u}_{k-1}$. Subsequently, the update step corrects the prediction by taking into account the newly received measurement $\boldsymbol{z}_k$. Finally, the state estimate for step $k$, $\hat{\boldsymbol{x}}_{k|k}$, is the result of applying the Kalman gain $\boldsymbol{K}_k$. In more detail, the prediction and update step are given by:
Prediction step:

$$\hat{\boldsymbol{x}}_{k|k-1} = f\left(\hat{\boldsymbol{x}}_{k-1|k-1}, \boldsymbol{u}_{k-1}\right), \tag{5.54}$$

$$\boldsymbol{P}_{k|k-1} = \boldsymbol{F}_{k-1}\boldsymbol{P}_{k-1|k-1}\boldsymbol{F}_{k-1}^T + \boldsymbol{Q}_{k-1}. \tag{5.55}$$

Update step:

$$\tilde{\boldsymbol{y}}_k = \boldsymbol{z}_k - h\left(\hat{\boldsymbol{x}}_{k|k-1}\right), \tag{5.56}$$

$$\boldsymbol{S}_k = \boldsymbol{H}_k \boldsymbol{P}_{k|k-1} \boldsymbol{H}_k^T + \boldsymbol{R}_k, \tag{5.57}$$

$$\boldsymbol{K}_k = \boldsymbol{P}_{k|k-1} \boldsymbol{H}_k^T \boldsymbol{S}_k^{-1}, \tag{5.58}$$

$$\hat{\boldsymbol{x}}_{k|k} = \hat{\boldsymbol{x}}_{k|k-1} + \boldsymbol{K}_k \tilde{\boldsymbol{y}}_k, \tag{5.59}$$

$$\boldsymbol{P}_{k|k} = \left(\boldsymbol{I} - \boldsymbol{K}_k \boldsymbol{H}_k\right) \boldsymbol{P}_{k|k-1}. \tag{5.60}$$

In the above equations $\boldsymbol{Q}$ is the process noise matrix, which measures the uncertainty of the supplied model, and $\boldsymbol{R}$ is the measurement noise matrix, which quantifies the noise level of the different measurements. The estimate covariance matrix $\boldsymbol{P}_k$ quantifies the confidence of the state estimate. The state transition and observation matrices $\boldsymbol{F}_k$ and $\boldsymbol{H}_k$ are the linearized system model $f(\boldsymbol{x}, \boldsymbol{u})$ and measurement model $h(\boldsymbol{x})$, respectively. For the current state estimate $\hat{\boldsymbol{x}}_k$ and inputs $\boldsymbol{u}_k$ the linearizations are given as

$$\boldsymbol{F}_k = \left.\frac{\partial f}{\partial \boldsymbol{x}}\right|_{\hat{\boldsymbol{x}}_{k|k}, \boldsymbol{u}_k} \quad \boldsymbol{H}_k = \left.\frac{\partial h}{\partial \boldsymbol{x}}\right|_{\hat{\boldsymbol{x}}_{k|k-1}}. \tag{5.61}$$

### Kinematic Model

The first filter includes a simple kinematic point-mass model. The measurement vector is defined as

$$\boldsymbol{z}_{\text{kin}} = \left[\tilde{x}, \tilde{y}, \tilde{v}_x, \tilde{a}_x, \tilde{a}_y, \tilde{r}\right]^T. \tag{5.62}$$

The model equations are given as

$$\dot{x} = v_x \cos(\psi) - v_y \sin(\psi) \tag{5.63}$$

$$\dot{y} = v_x \sin(\psi) + v_y \cos(\psi) \tag{5.64}$$

$$\dot{v}_x = a_x + \dot{\psi} v_y \tag{5.65}$$

$$\dot{v}_y = a_y - \dot{\psi} v_x \tag{5.66}$$

$$\dot{a}_x = 0 \tag{5.67}$$

$$\dot{a}_y = 0 \tag{5.68}$$

$$\dot{\psi} = r \tag{5.69}$$

$$\tag{5.70}$$

and the measurement model is defined as

$$h_{\text{kin}}(\hat{\boldsymbol{x}}) = \left[\hat{x}, \hat{y}, \hat{v}_x, \hat{v}_y, \hat{a}_x, \hat{a}_y, \hat{r}\right]^T. \tag{5.71}$$

### Dynamic Bicycle Model

The second filter implements a dynamic bicycle model. The measurement vector is defined as

$$\boldsymbol{z}_{\mathrm{dyn}} = [\tilde{x}, \tilde{y}, \tilde{v}_x, \tilde{a}_x, \tilde{a}_y, \tilde{r}]^T . \tag{5.72}$$

According to [LiaoBorrelli18], the model equations are given as

$$\dot{x} = v_x \cos(\psi) - v_y \sin(\psi) \tag{5.73}$$

$$\dot{y} = v_x \sin(\psi) + v_y \cos(\psi) \tag{5.74}$$

$$\dot{v}_x = a_x + \dot{\psi} v_y \tag{5.75}$$

$$\dot{v}_y = v_y \frac{-C_\mathrm{f} - C_\mathrm{r}}{m v_x} - \dot{\psi}\left(v_x + \frac{l_\mathrm{f} C_\mathrm{f} - l_\mathrm{r} C_\mathrm{r}}{m v_x}\right) + \delta_\mathrm{f} \frac{C_\mathrm{f}}{m} \tag{5.76}$$

$$\dot{a}_x = 0 \tag{5.77}$$

$$\dot{a}_y = 0 \tag{5.78}$$

$$\dot{\psi} = r \tag{5.79}$$

$$\ddot{\psi} = v_y \frac{-l_\mathrm{f} C_\mathrm{f} + l_\mathrm{r} C_\mathrm{r}}{I_\mathrm{z} v_x} - \dot{\psi}\left(\frac{l_\mathrm{f}^2 C_\mathrm{f} - l_\mathrm{r}^2 C_\mathrm{r}}{I_\mathrm{z} v_x}\right) + \delta_\mathrm{f} \frac{l_\mathrm{f} C_\mathrm{f}}{I_\mathrm{z}} . \tag{5.80}$$

The measurement model is defined as

$$h_{\mathrm{dyn}}(\hat{\boldsymbol{x}}) = [\hat{x}, \hat{y}, \hat{v}_x, \hat{v}_y, \hat{a}_x, \hat{a}_y, \hat{r}]^T . \tag{5.81}$$

The singularities in Equation 5.76 and Equation 5.80 are the reason why this estimator is not suitable for low velocities. Therefore, the estimator build on the kinematic model is applied for velocities close to 0. However, for high velocities the impact of the lateral dynamics increase and cannot be sufficiently modeled by the kinematic model. Hence, the combination of the two estimators using the two presented models together can be employed for a more accurate estimation.

## 5.3.3 Experimental Results

The controller parameters for the experiments on the BARC are the same parameters as in the simulation experiments, which are listed in Table 5.1. The experiments for the multi-agent LMPC on the physical system begin with the data collection through the single-agent LMPCs. Therefore, the path-following MPCs for the three different initializations are executed followed by the single-agent LMPCs. Starting with the initialization close to the inside of the race track with $e_{y,\mathrm{ref}} = 0.45\,\mathrm{m}$ yields the closed-loop trajectories shown in Figure 5.25 and Figure 5.26. The trajectories exhibit some differences. Agent 2 reaches higher velocities, which is the reason of the the greater deviation from the inside of the track compared to agent 1. The cause for the greater velocities achieved by agent 2, is the difference in the masses of the BARCs. Moreover, for each agent there is one trajectory at low velocities, which deviates from the other taken closed-loop trajectories. This behavior is a result of the combination of sensor noise, connection issues and errors in the localization of the GPS. However, the system is robust enough to recover from these errors and continue the learning and performance improvement.

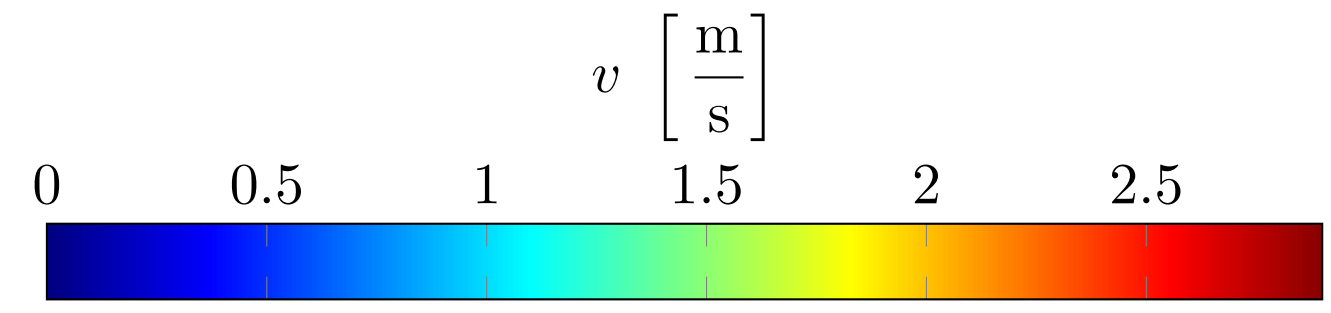

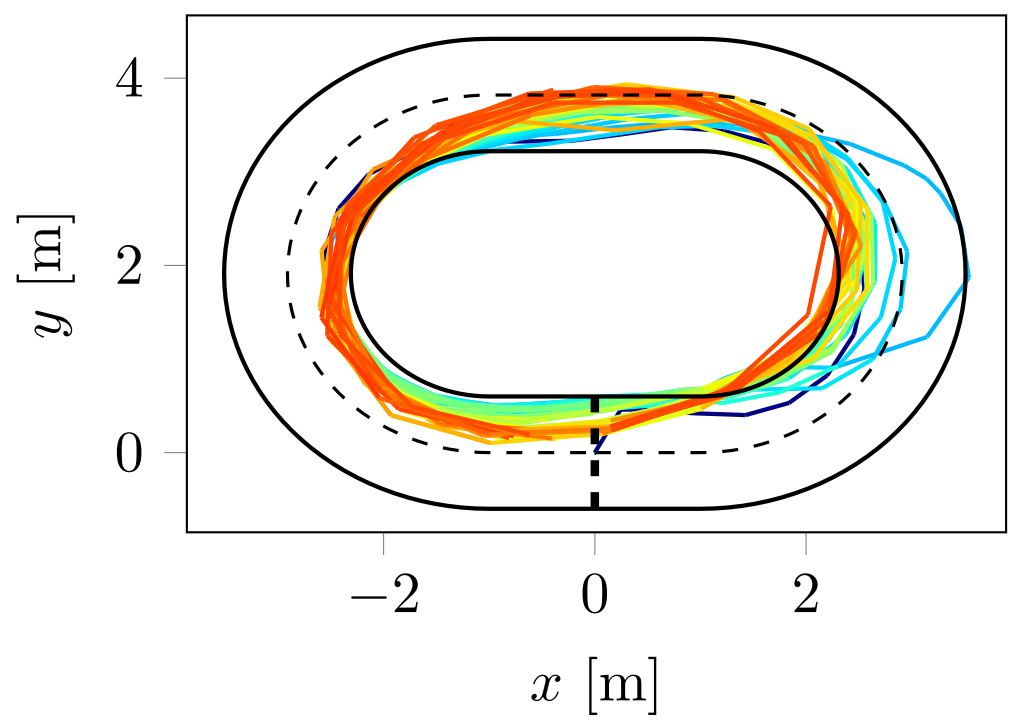

Figure 5.25: Agent 1's LMPC for inner initialization.

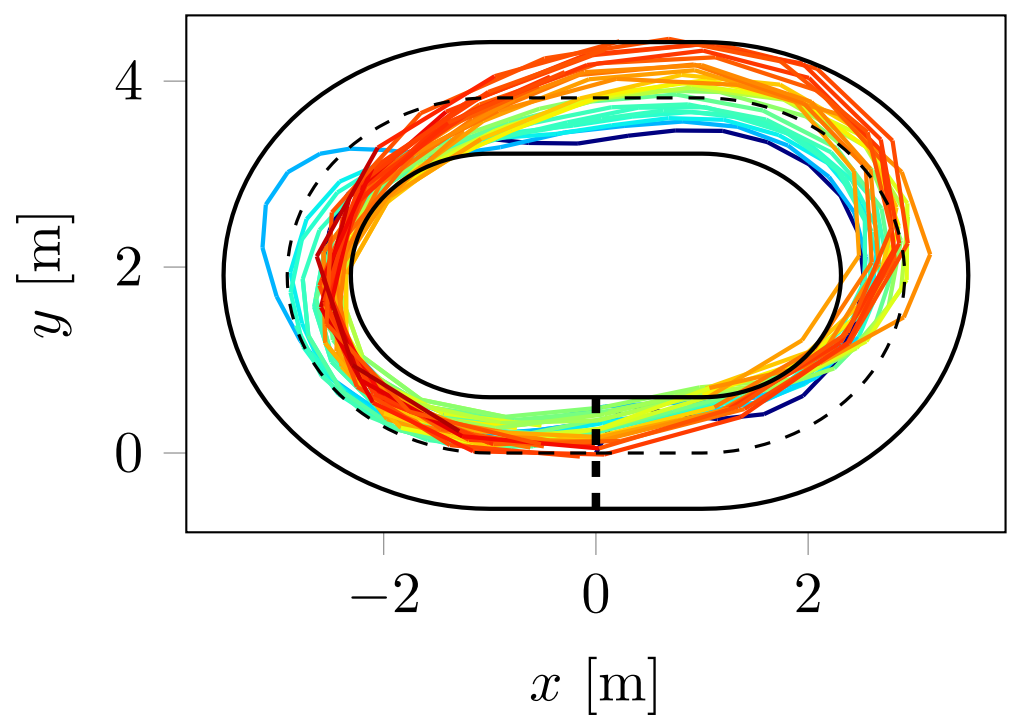

Figure 5.26: Agent 2's LMPC for inner initialization.

The center initialization LMPC based on a path-following MPC with $e_{y,\text{ref}} = 0.0\,\text{m}$ is shown in Figure 5.27 and Figure 5.28 for both agents. This initialization converges to comparable closed-loop trajectories for higher velocities. Nevertheless, the second agent's trajectory is more tilted compared to the one from agent 1. This difference in closed-loop trajectories can be the result of the different setup of the BARCs, since agent 1 has differential gears, while agent 2 does not.

The performance improvement for each iteration is displayed in Figure 5.29. For every iteration the velocity increases until it converges starting from lap 25. Furthermore, this velocity profile is less smooth than the velocity profile shown in Figure 5.9. However, this velocity profile is recorded from a real system and each sensor reading is noisy and consequently inaccuracies are expected.

The last single-agent LMPC is executed starting from a path-following MPC, which tracks a line on the outside of the track, where $e_{y,\text{ref}} = -0.45\,\text{m}$. It is shown in the Figure 5.30 and Figure 5.31 for the two agents. As for the previous experiments, the resulting closed-loop trajectories are very similar. Although, agent 2 exhibits a higher maximum velocity again.

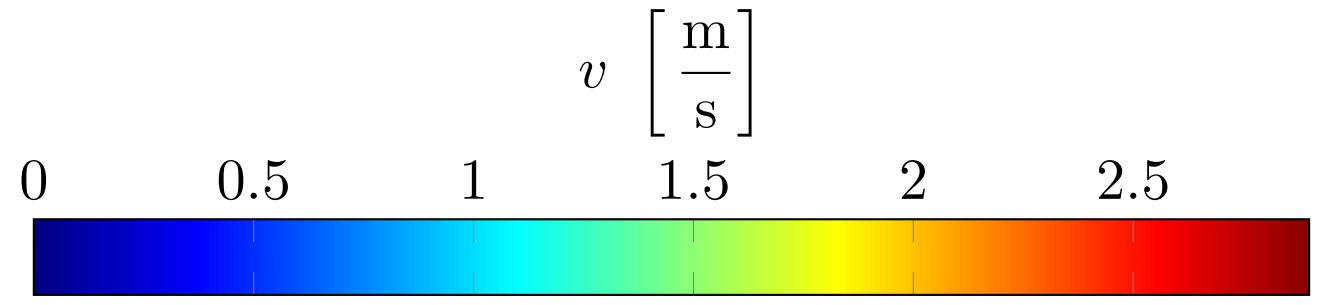

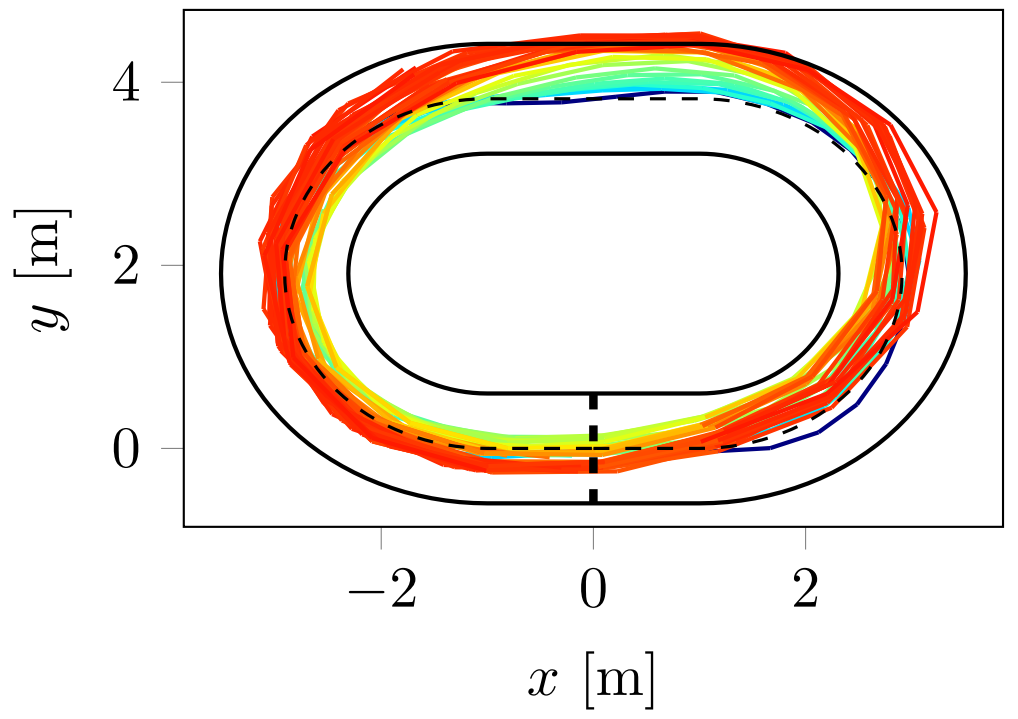

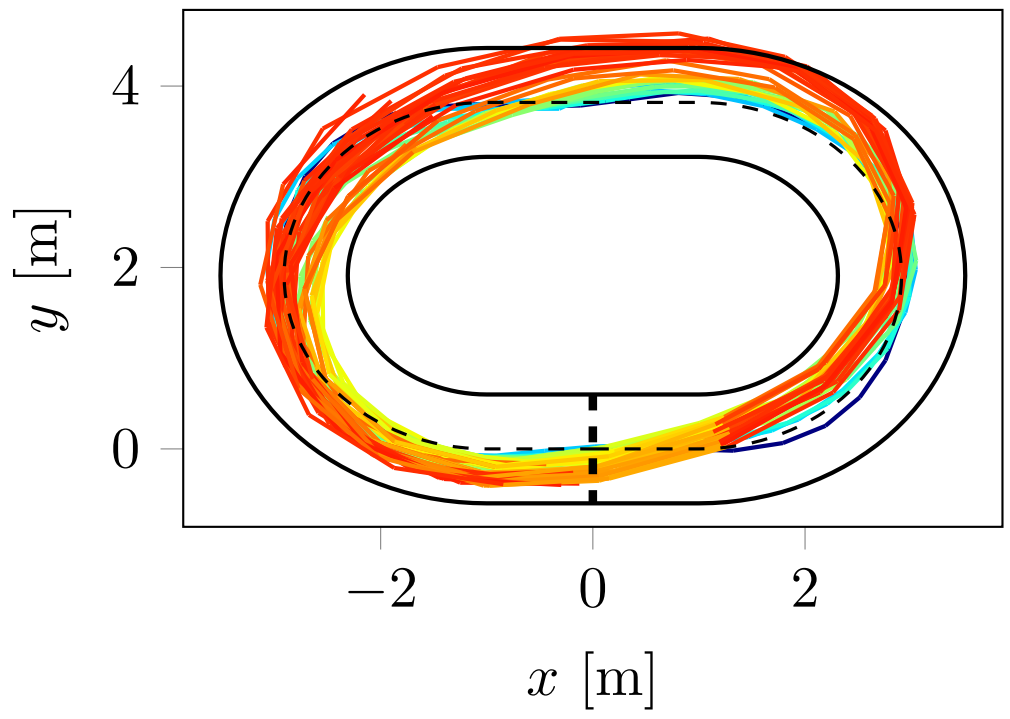

Figure 5.27: LMPC for center initialization of agent 1

Figure 5.28: LMPC for center initialization of agent 2

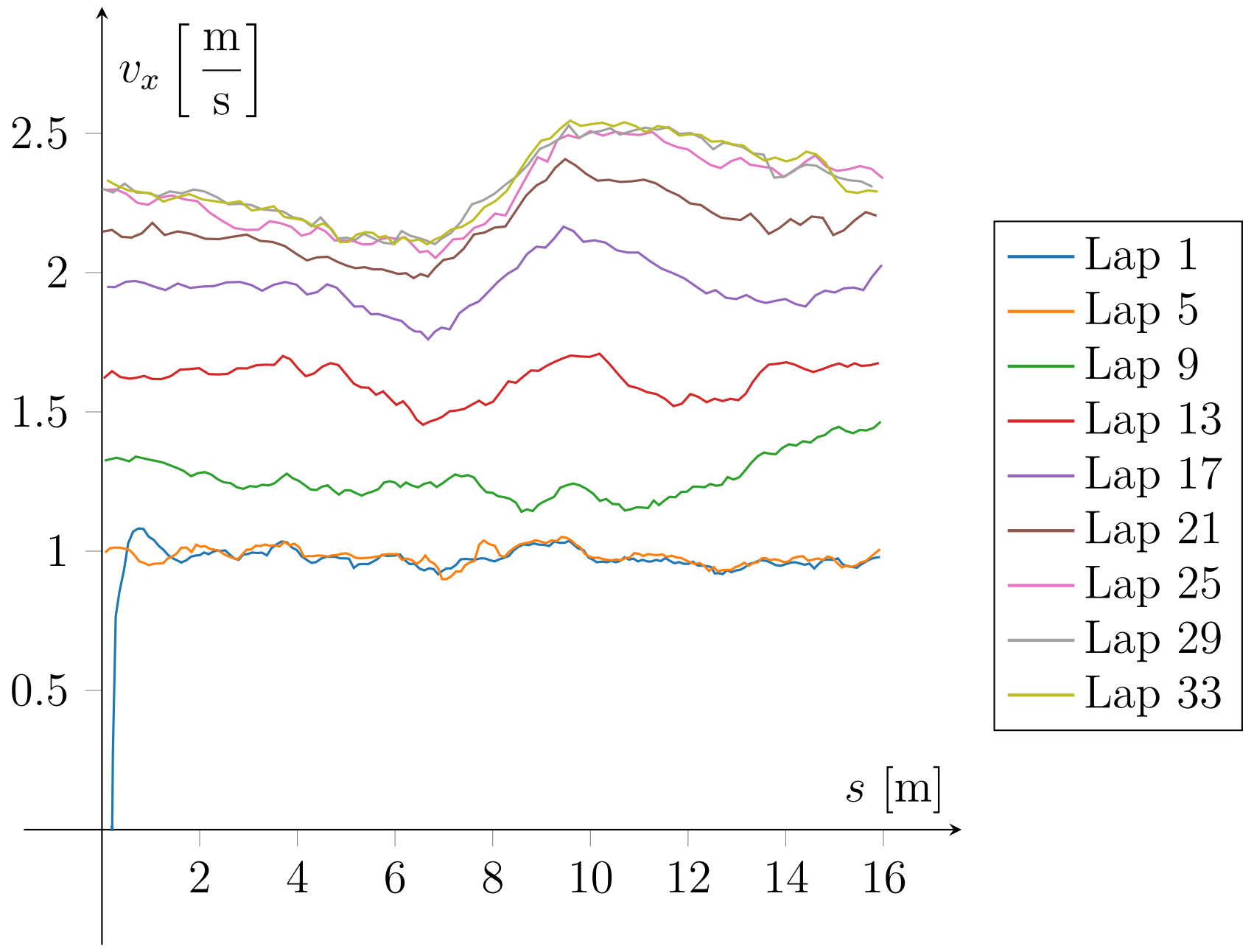

Figure 5.29: Velocity profile for the LMPC with center initialization for agent 1.

The convergence behavior of the single-agent LMPC can be deduced from Figure 5.32. In the beginning the time steps needed to complete an iteration decrease monotonically for all of the different initializations. The LMPC using the inner and outer initializations reach different locally optimal solutions, which results in a difference in the required time steps. Furthermore, the number of time steps required increase for the center initialization after 15 iterations and starts to converge to a different locally optimal solution. Therefore, for this nonlinear system the LMPC converges to distinct locally optimal solutions starting from the various initializations.

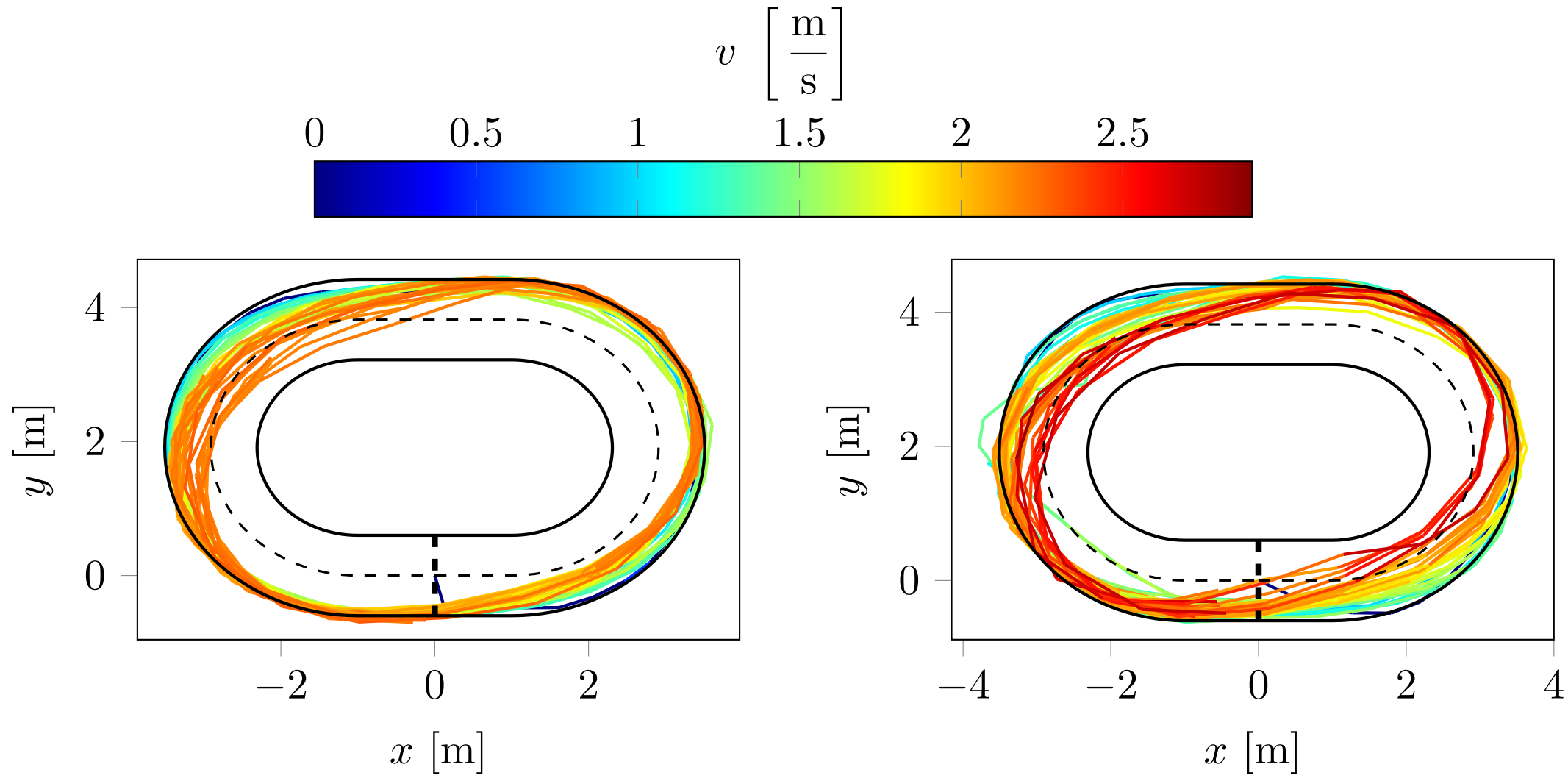

Figure 5.30: LMPC for outer initialization of agent 1

Figure 5.31: LMPC for outer initialization of agent 2

Combining all of the previous experiments for both agents result in the terminal safe sets displayed in Figure 5.33 and Figure 5.34. The figures show that this approach for initializing the terminal safe set results in a nearly completely explored state space for the positional states. Furthermore, for many of the reached positions high velocity states are available as well. This increases the performance for the multi-agent setting and allows for a better performance from the very start of the race.

Experiments for the two agents running on the oval track simultaneously have not been successful. Potential error sources have been investigated and examined, but a robust solution could not be determined over the course of this thesis. The behavior of the controller is already analyzed using the simulator, which leads to the conclusion, that the controller is functioning correctly. Therefore, the observed issues are either related to the sensors, the estimator or the communication among the different devices. During the experiments there are notable errors in the localization provided by the GPS system. These can result in irregular sudden offsets in the localization for two to three subsequent measurements. These inaccuracies are apparent for a frequency of 12 Hz and get worse for higher frequencies. For the single-agent case, the provided measurements are sufficient and the errors are mostly handled well by the state estimator. However, for two agents the frequency of the GPS measurements is halved, which results in a lower frequency than the frequency of the controller. This leads to a poor state estimation. Even though the included onboard fusion for the GPS with its IMU provide a frequency of up to 100 Hz, the received signal is noisy and is even more prone to inaccuracies in the raw GPS signal. Therefore, the fused signal is not used. Additionally, the correctness of the applied estimator is not guaranteed, since there does not exist any ground truth, which enables the appropriate selection of the different filter parameters. This potentially results in errors in the estimation and can prevent the proposed controller from running successfully. The last error source are connection issues, which can produce package loss and delays. Occasionally, there is a noticeable loss of packages from the sensors to the Odroid or from the Odroid to the MSI laptop and vice versa. This can be related to the amount of traffic on the network or the chosen Wi-Fi channel. All of the above contribute to an imprecise state estimation. Finally, there is an inadmissible offset between the true and estimated state. Furthermore, since the linear regression model is learned online from the estimated states, the great model mismatch will also impact the performance of the controller. This complicates the successful and robust execution of the multi-agent LMPC significantly.

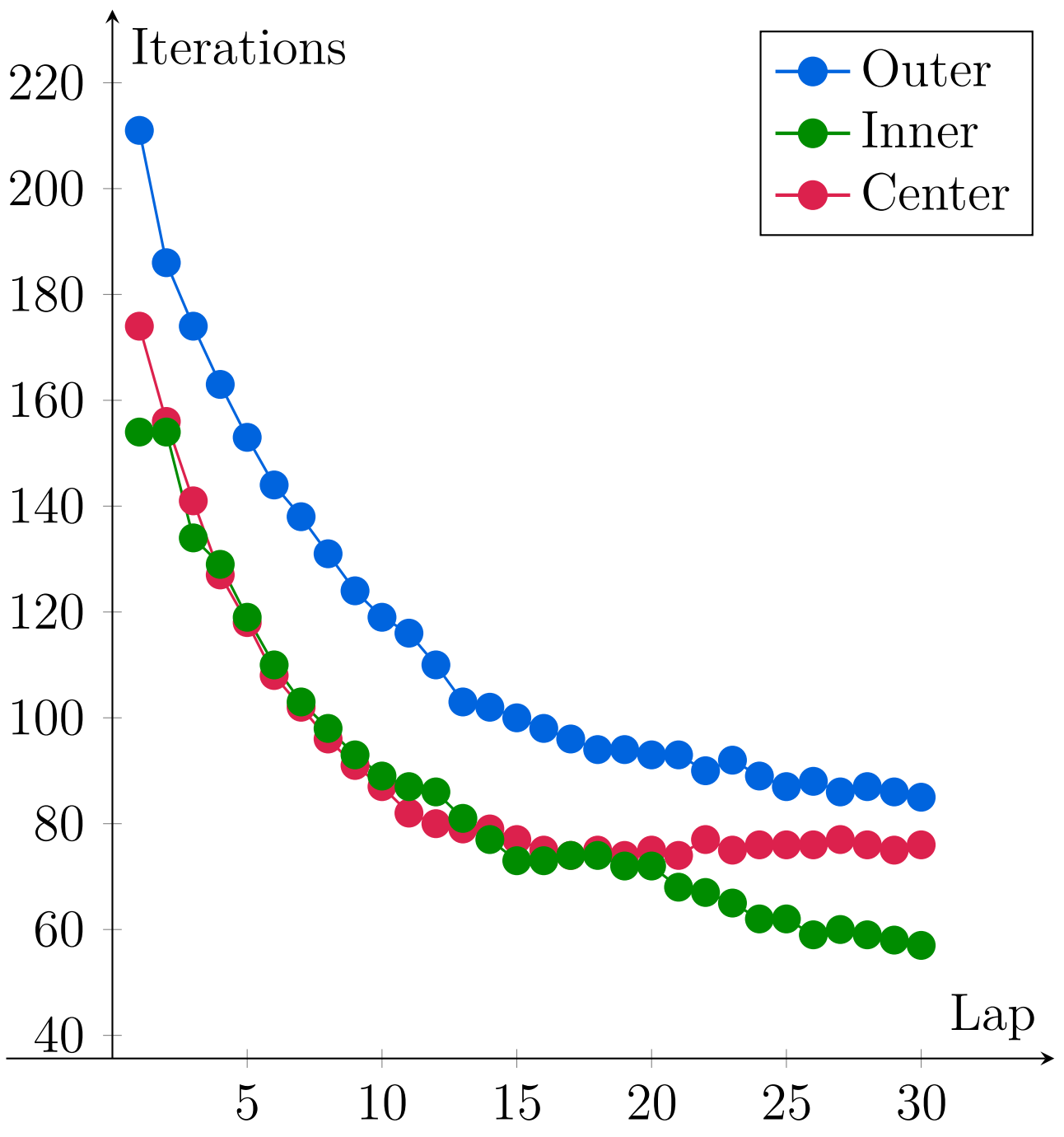

Figure 5.32: Number of time steps needed to complete each lap for agent 1.

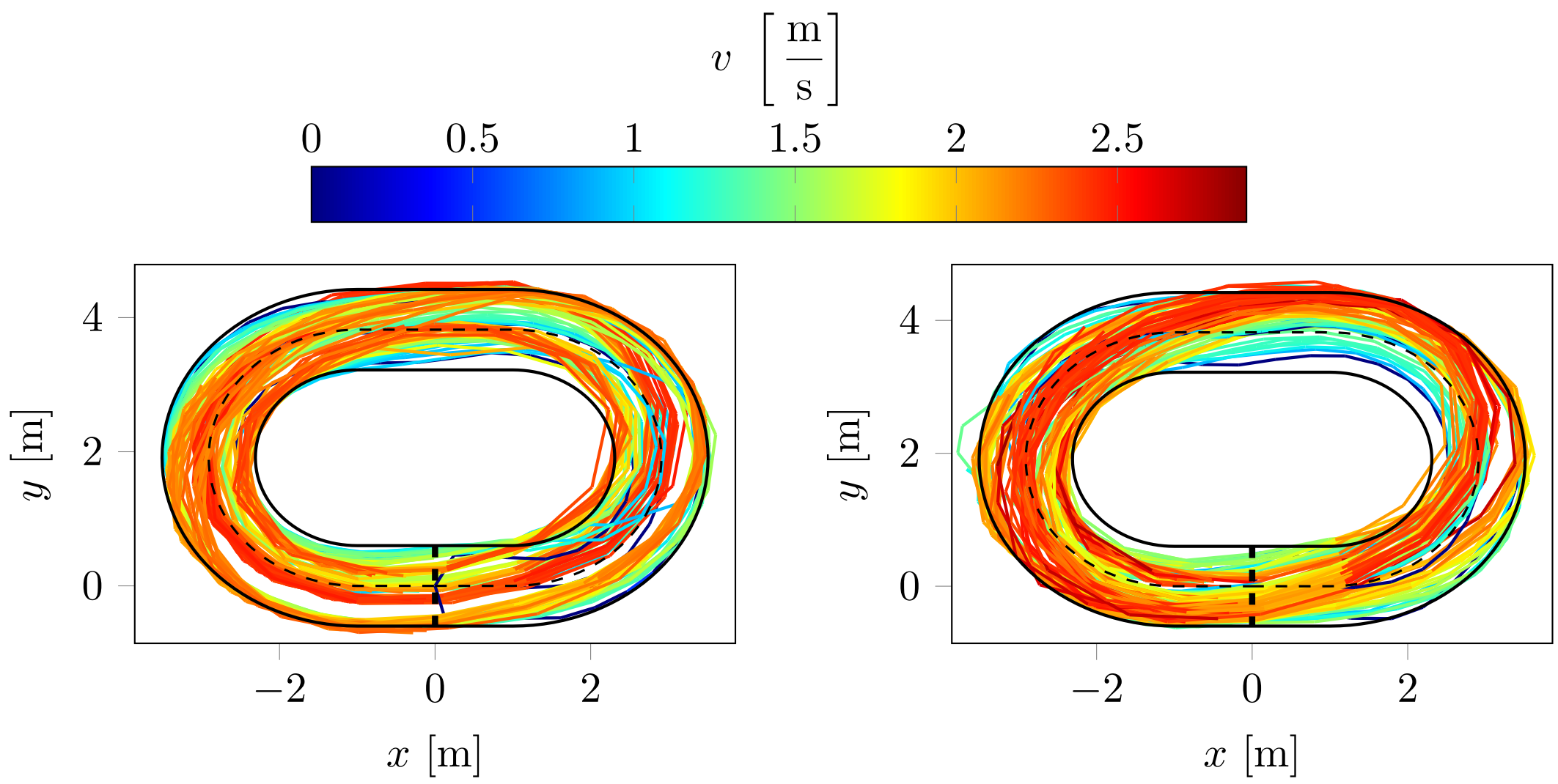

Figure 5.33: Agent 1's combined safe set after all initializations.

Figure 5.34: Agent 2's combined safe set after all initializations.

# Chapter 6

# Conclusion and Outlook

The goal of this project was to extend the single-agent LMPC to the multi-agent scenario for the car racing driving problem. This thesis introduced a learning model predictive controller for the competitive multi-agent racing problem. During a simulated race, multiple overtaking maneuvers were successfully executed. Generally, the agent must deviate from its optimal trajectory during an overtaking maneuver. The overtaking maneuver has been enabled by the introduced state-varying safe set and the safe set initialization.

The conducted simulations validate the proposed improvements on the single-agent LMPC and their extension to the multi-agent LMPC. The race between the two agents has been formulated as a two-player game and the corresponding optimization problem is solved by optimizing each agent's prediction while assuming the adversary's prediction is fixed. The proposed strategy for initializing the terminal safe set by running multiple LMPCs from different initializations resulted in a richer terminal safe set, which is crucial for completing overtaking maneuvers. The state-varying safe set ensures convexity and enables using collected states from previous recordings. Furthermore, the terminal cost function has been adapted such that the controller demonstrates a more desirable behavior on the race track. Finally, a linear cost on the signed distance is added, which motivates competitive behavior among agents.

The experiments for two 1/10 scale remote controlled race cars were not successful, mainly due to the estimator's performance and the provided measurements. Implementing the estimator directly on the Odroid and using a dynamic bicycle model in the EKF did not eliminate inaccuracies in the estimation. A confirmation of the correctness of the state estimation is not possible because there is no ground truth available. Furthermore, the GPS system's noise increases when using multiple mobile beacons simultaneously. For future experiments, an outlier detection for determining errors in the localization is suggested. Additionally, the estimator's accuracy should be tested. This could be achieved using a motion capture system like the Vicon camera-based system.

This thesis assumed that a perfect prediction of the adversarial agent is available at every time step. The perfect prediction has been provided by exchanging the current prediction with the opponent. However, the communication of strategies is contradictory in any competitive scenario. Future work will focus on developing an accurate prediction of the opponent's prediction. This is a challenging problem because interactions on the race track are limited and change depending on the distance between the agents.

Overall, the proposed LMPC for the multi-agent racing scenario only needs few prerecorded iterations to successfully race adversarial agents at the limit of handling.

# Appendix

## A.1 Contents Archive

There is a folder **MSC__016__Brunke/** in the archive. The main folder contains the entries

- **MSC__016__Brunke.pdf**: the pdf-file of the thesis MSC-016.
- **Data/**: a folder with all the relevant data, programs, scripts and simulation environments.
- **Latex/**: a folder with the *.tex documents of the thesis MSC-016 written in Latex and all figures (also in *.svg data format if available).
- **Presentation/**: a folder with the relevant data for the presentation including the presentation itself, figures and videos.

**Erklärung**

Ich, Lukas Brunke (Student des Maschinenbaus an der Technischen Universität Hamburg-Harburg, Matrikelnummer 21267608), versichere, dass ich die vorliegende Masterarbeit selbstständig verfasst und keine anderen als die angegebenen Hilfsmittel verwendet habe. Die Arbeit wurde in dieser oder ähnlicher Form noch keiner Prüfungskommission vorgelegt.

| | |
|---|---|
| Unterschrift | Datum |